\documentclass{article}

\PassOptionsToPackage{numbers, compress}{natbib}
\usepackage[main,final]{neurips_2025}

\usepackage[utf8]{inputenc} % allow utf-8 input
\usepackage[T1]{fontenc}    % use 8-bit T1 fonts
\usepackage{hyperref}       % hyperlinks
\usepackage{url}            % simple URL typesetting
\usepackage{booktabs}       % professional-quality tables
\usepackage{amsfonts}       % blackboard math symbols
\usepackage{nicefrac}       % compact symbols for 1/2, etc.
\usepackage{microtype}      % microtypography
\usepackage{xcolor}         % colors
\usepackage{amsmath}
\usepackage{multirow}
\usepackage{graphicx}
\usepackage{pifont}
\usepackage[normalem]{ulem}
\usepackage{cleveref}
\usepackage{subcaption}
\usepackage{array}
\usepackage{enumitem}

\newcommand{\cmark}{\ding{51}}%
\newcommand{\xmark}{\ding{55}}%
\newcommand{\myparagraph}[1]{\smallskip\noindent\textbf{#1}}

\title{MATCH: \underline{M}ulti-faceted \underline{A}daptive \underline{T}opo-\underline{C}onsistency for Semi-Supervised \underline{H}istopathology Segmentation}

\author{
Meilong Xu$^{1,\dagger}$ \quad
Xiaoling Hu$^{2, \dagger}$ \quad
Shahira Abousamra$^{3}$ \quad
Chen Li$^{1}$ \quad
Chao Chen$^{1}$ \\[3pt]
$^{1}$Stony Brook University, NY, USA \\
$^{2}$
    Massachusetts General Hospital and Harvard Medical School, MA, USA \\
$^{3}$Department of Biomedical Data Science, Stanford University, CA, USA \\
}

\begin{document}

\maketitle
\renewcommand{\thefootnote}{\fnsymbol{footnote}}
\footnotetext{$\dagger$ Equal contribution. Email: Meilong Xu (meixu@cs.stonybrook.edu).}
\renewcommand{\thefootnote}{\arabic{footnote}}

\begin{abstract}
In semi-supervised segmentation, capturing meaningful semantic structures from unlabeled data is essential. This is particularly challenging in histopathology image analysis, where objects are densely distributed. To address this issue, we propose a semi-supervised segmentation framework designed to robustly identify and preserve relevant topological features. Our method leverages multiple perturbed predictions obtained through stochastic dropouts and temporal training snapshots, enforcing topological consistency across these varied outputs. This consistency mechanism helps distinguish biologically meaningful structures from transient and noisy artifacts. A key challenge in this process is to accurately match the corresponding topological features across the predictions in the absence of ground truth. To overcome this, we introduce a novel matching strategy that integrates spatial overlap with global structural alignment, minimizing discrepancies among predictions. Extensive experiments demonstrate that our approach effectively reduces topological errors, resulting in more robust and accurate segmentations essential for reliable downstream analysis. Code is available at \href{https://github.com/Melon-Xu/MATCH}{https://github.com/Melon-Xu/MATCH}.
\end{abstract}

\section{Introduction}
\label{sec:intro}
Accurate segmentation of glands and nuclei in histopathology images is critical for digital pathology, significantly influencing diagnosis, prognosis, and treatment planning by enabling precise quantification of morphological and structural tissue features~\cite{fleming2012colorectal, montironi2005gleason, karasaki2023evolutionary}. Numerous fully-supervised segmentation methods~\cite{ronneberger2015u, zhou2018unet++, graham2019mild, graham2019hover, isensee2021nnu, he2023toposeg, ma2024segment, horst2024cellvit} have demonstrated substantial success. However, densely distributed cellular structures in histopathology images often induce topological errors, including false merges or splits, severely impacting clinical reliability. Additionally, fully supervised methods demand extensive annotated datasets, which are costly, time-consuming, and not scalable~\cite{sirinukunwattana2017gland, kumar2019multi}. This limitation motivates exploring semi-supervised learning (SSL) strategies capable of leveraging abundant unlabeled data alongside limited annotations.

\begin{figure}[t]
    \centering
    \includegraphics[width=1\columnwidth]{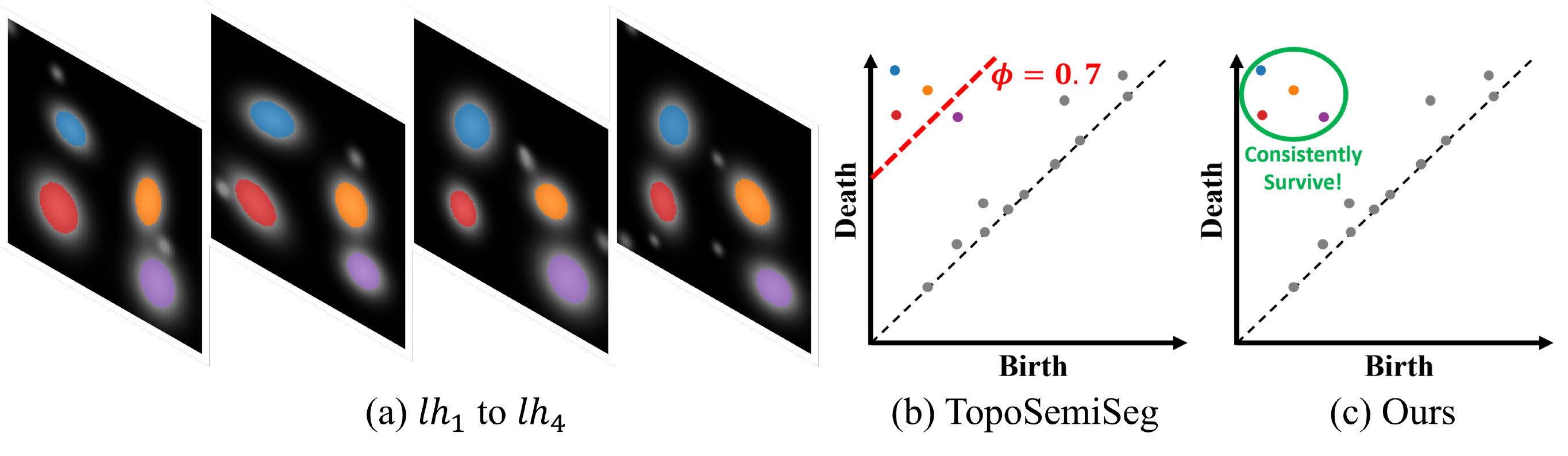}
    \caption{Intuition of the proposed framework.  
  (a) Colored likelihood maps are coming from the MC dropout. Connected components consistently matched in at least three predictions retain identical colors across instances, indicating topological stability; components shown in grey fail to reach this consensus and are therefore treated as topologically transient. 
  (b) Limitation of TopoSemiSeg~\cite{xu2024semi}, which relies on a fixed persistence threshold ($\phi=0.7$, \textcolor{red}{red} dashed line) and therefore overlooks less-persistent yet meaningful structures (e.g.\ the \textcolor{violet}{violet} point). 
  (c) Our method adaptively identifies relevant topological structures without the need for human-selected thresholds.}
  \label{fig:motivation_fixed_thres}
  % %\vspace{-.27in}
\end{figure}

Recent SSL approaches have significantly enhanced segmentation accuracy in contexts of limited supervision~\cite{yu2019uncertainty, sohn2020fixmatch, li2020transformation, luo2021semi, luo2022semi, wu2022cross, you2022simcvd, zhang2022boostmis, you2023rethinking, zhang2022discriminative,basak2023pseudo,zhou2023xnet,you2024mine, konwer2025enhancing, zhou2024xnet, yang2024anomaly, yang2024multimodal, nguyen2025semi}. 
Nevertheless, these methods typically do not explicitly target topological errors, resulting in seemingly small segmentation errors with consequential significant topological inaccuracies that affect segmentation robustness.
To explicitly address topological errors, persistent homology~\cite{edelsbrunner2010computational} offers a rigorous mathematical framework that captures and characterizes topological features, such as connected components and loops in data across multiple scales. 
The output, persistence diagram, summarizes these structures as dots in a 2D diagram. For each dot, the coordinate difference $(y-x)$ captures the \emph{persistence} of the topological structure across scales. 
Building upon this mathematical foundation, TopoSemiSeg~\cite{xu2024semi} introduces topology-aware constraints into SSL frameworks, utilizing persistent homology to enforce topological consistency between teacher and student model predictions. 
Despite its effectiveness, TopoSemiSeg mainly relies on a predefined, hand-picked persistence threshold to identify meaningful topological structures. Such fixed thresholds are not data-driven, potentially biased, and can exclude relevant structures or retain irrelevant ones, as shown in~\Cref{fig:motivation_fixed_thres}. 

To address this issue, we investigate how to identify reliable topological structures from model predictions in a robust and adaptive manner, and enforce model consistency over these structures.
We first revisit the fundamental principles of semi-supervised learning -- robustness against perturbations. For an image without a training label, to identify reliable information, semi-supervised approaches typically add perturbations at the input level (i.e., augmentation) and at the model level (i.e., Monte Carlo Dropout). Pixel-level predictions that persist across these perturbations are considered reliable and used to self-supervise the model. 

Our main idea is to tightly couple this SSL robustness-against-perturbation principle with topological reasoning. Moving beyond pixel-level, we identify topological structures that persist across different perturbations. These structures are considered reliable and used to self-supervise the model. This idea avoids a hand-picked threshold to determine reliable topological structures, and adaptively identifies truly relevant structures to enhance the model's topological reasoning power in an SSL setting.

Building on this idea, we propose a novel SSL segmentation framework employing \textbf{dual-level topological consistency}. Our method identifies significant topological features by examining predictions generated with different model perturbations. We formulate the structure correspondence task as a contrastive learning problem, distinguishing stable features, i.e., those consistently detected across multiple predictions, from transient or noisy structures. To identify the stable topological structures, we introduce an advanced matching algorithm that integrates spatial overlap, topological persistence, and spatial proximity criteria to associate topological structures across diverse predictions reliably.

As for perturbations, we propose to employ Monte Carlo (MC) dropout perturbations~\cite{gal2016dropout}. Meanwhile, we stress the importance of a temporal view of SSL. Previous works, such as~\cite{laine2016temporal, li2023confidence, li2023calibrating, shin2024revisiting}, demonstrate that evaluating the predictions in different training snapshots can reveal informative signals for robust prediction. Inspired by this, we propose to also compare topological structures across model snapshots at different training epochs.
% \cc{Revert to temporal?}
By explicitly optimizing for dual-level topological consistency, our framework enhances structural coherence within the student model without relying on extensive pixel-wise annotations. Our key contributions can be summarized as follows:
\begin{itemize}[topsep=0pt,itemsep=1pt,partopsep=0pt, parsep=0pt,leftmargin=.1in]
    \item We propose a novel integration of topological reasoning into the semi-supervised segmentation framework to robustly identify and preserve meaningful topological structures.
    \item We introduce dual-level topological consistency, measuring structural stability from intra-perturbed predictions (MC dropout) and temporal training snapshots, to effectively utilize unlabeled data.
    \item We develop a novel matching algorithm that integrates spatial overlap, topological persistence, and spatial proximity to accurately match topological structures across predictions.
\end{itemize}
Through extensive experiments on three widely used histopathology image datasets, our method significantly improves the topological accuracy while achieving comparable pixel-wise performance with limited annotations. 

\section{Related Works}
\label{sec:related_works}
\myparagraph{Segmentation with Limited Supervision.}
Semi-supervised learning enhances medical image segmentation by effectively utilizing limited labeled data together with abundant unlabeled data.
Consistency regularization approaches, such as the Mean Teacher model~\cite{tarvainen2017mean}, ensure stable segmentation despite input variations~\cite{tarvainen2017mean, li2020transformation, wang2022semi, ouali2020semi, xu2024semi}. Pseudo-labeling progressively improves accuracy by leveraging confident model predictions on unlabeled data~\cite{yao2022enhancing, seibold2022reference, zhang2022discriminative}. Adversarial training aligns segmentation outputs from labeled and unlabeled datasets using discriminator networks~\cite{hung2018adversarial, lei2022semi}. Additionally, uncertainty estimation methods such as MC dropout and Bayesian neural networks enhance reliability by effectively handling uncertainty during pseudo-label generation~\cite{gal2016dropout, yu2019uncertainty, nair2020exploring, luo2022semi, xu2023ambiguity}, while entropy minimization 
is used to reduce prediction uncertainty~\cite{grandvalet2004semi, berthelot2019mixmatch, xie2024entropy}.
Contrastive learning strengthens segmentation robustness by training models to differentiate similarities and distinctions among data pairs, thereby boosting overall segmentation quality~\cite{you2022simcvd, basak2023pseudo, you2023rethinking, you2024mine}.

\begin{figure}[t]
    \centering
    \includegraphics[width=1\columnwidth]{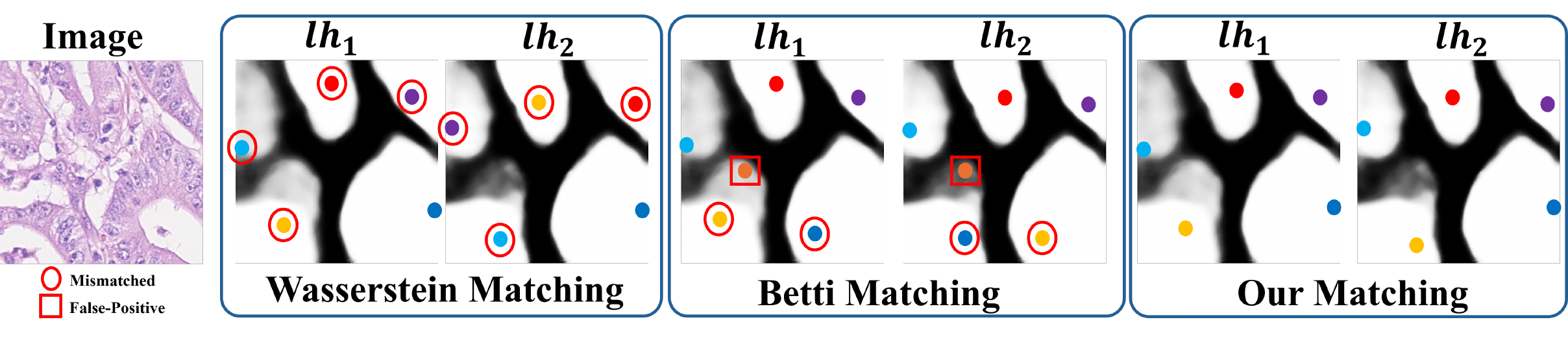}
    \caption{Comparison of our matching with Betti Matching~\cite{stucki2023topologically} and Wasserstein Matching~\cite{hu2019topology}. We match two likelihood maps obtained from the same input histopathology patch. The birth critical points of the matched pairs are highlighted in the same color. Note that Wasserstein Matching gets most matches wrong, and Betti Matching also gets two matches wrong while pairing biologically unrelated features when lacking the guidance of the ground truth.}
    \label{fig:matching_comparison}
\end{figure}

\myparagraph{Topology-Aware Image Segmentation.}
Topology-aware methods have been proposed to enforce correct topology, like connectivity or correct counts in segmentation tasks~\cite{hu2019topology, hu2020topology, shit2021cldice, clough2020topological, yang20213d, yang2021topological, hu2022structure, gupta2022learning, wang2022ta, stucki2023topologically, wang2020topogan, xu2024topocellgen, zhang2023topology, lux2024topograph}. These methods typically use differentiable loss functions derived from topological data analysis tools, including persistent homology~\cite{hu2019topology, clough2020topological, stucki2023topologically}, discrete Morse theory~\cite{hu2020topology, hu2023learning, gupta2023topology}, topological interactions~\cite{gupta2022learning, berger2024topologically}, homotopy warping~\cite{hu2022structure}, centerline-based comparisons~\cite{shit2021cldice, wang2022ta}.
These methods generally rely heavily on precisely annotated labels.
Xu \textit{et al.}~\cite{xu2024semi} propose TopoSemiSeg to combine SSL with topological constraints.
Classical persistent homology-based segmentation methods rely on Wasserstein matching~\cite{hu2019topology,xu2024semi}, which compares persistence diagrams based solely on feature lifespans. However, this approach may produce ambiguous or incorrect correspondences, as illustrated in~\Cref{fig:matching_comparison}. To alleviate spatial inconsistencies, several methods were proposed~\cite{stucki2023topologically, wen2024topology}. Betti Matching~\cite{stucki2023topologically} embeds predictions and ground truth into a shared super-level filtration, ensuring alignment only among overlapping topological features. However, as shown~\Cref{fig:matching_comparison}, it cannot ensure fully correct matching when the ground truth is missing and is too sensitive to preserve some transient structures. Our proposed MATCH-Pair could achieve almost completely accurate matching without the ground truth.

\section{Methodology}
\label{sec:method}
The motivation of our proposed framework is to identify meaningful topological structures directly from perturbed predictions without the ground truth. 
The main challenge is to accurately match corresponding topological structures across multi-facet predictions that often contain substantial noise and variability. To overcome this challenge, we introduce MATCH-Pair, a pairwise matching algorithm, and MATCH-Global, an extended global matching algorithm, to robustly identify stable structures across multiple predictions. Building upon these matching algorithms, we propose dual-level topological consistency constraints: intra-topological consistency, enforcing consistency across multiple stochastic predictions, and temporal-topological consistency, ensuring stability across consecutive training snapshots. These consistency constraints directly optimize the student model, enabling it to learn robust segmentation representations from limited labeled data.

\begin{figure*}[ht]
    \centering 
    \includegraphics[width=1\linewidth]{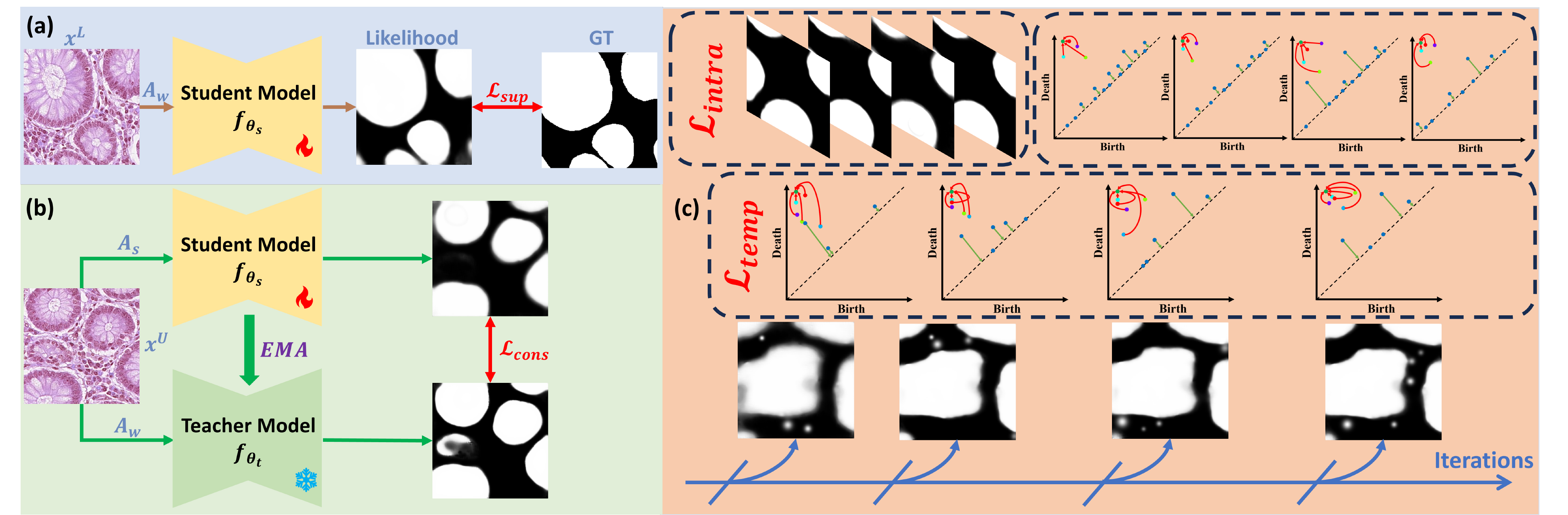}
    \caption{Overview of the proposed MATCH framework with dual-level topological consistency. Note that the $\mathcal{L}_{\text{intra}}$ and $\mathcal{L}_{\text{temp}}$ are used to directly optimize the parameters of the student model.    
    }
\label{fig:overall_pipeline}
\end{figure*}

Our method overview is shown in Figure \ref{fig:overall_pipeline}. The proposed MATCH framework leverages labeled data via supervised loss and unlabeled data through pixel-wise and dual-level topological consistency.

In this section, we will start by introducing the preliminaries of classic SSL. Next, we will use $3$ subsections to introduce MATCH-Pair, MATCH-Global, and the dual-level topological consistency.

\myparagraph{Preliminaries: SSL training.} We address the semi-supervised image segmentation problem by leveraging a teacher-student framework, a widely adopted paradigm in semi-supervised learning \cite{tarvainen2017mean}. Let $\mathcal{D}_L = \{(x_i^L, y_i^L)\}_{i=1}^{N_L}$ denote the labeled dataset, where $x_i^L$ represents the input image and $y_i^L \in \{0, 1\}^{H \times W}$ is the corresponding pixel-wise annotation. Let $\mathcal{D}_U = \{x_j^U\}_{j=1}^{N_U}$ denote the unlabeled dataset. In our setting, the number of labeled samples is significantly smaller than the number of unlabeled samples, i.e., $N_L \ll N_U$. Our objective is to train a segmentation model $f_\theta$, parameterized by $\theta$, that accurately predicts segmentation masks using labeled and unlabeled data.

In this framework, the student model $f_{\theta_s}$ is trained using both supervised and unsupervised losses, while the teacher model $f_{\theta_t}$ provides stable targets for the student by being updated as an exponential moving average (EMA) of the student’s parameters: $\theta_t^{(\tau + 1)} = \alpha \theta_t^{(\tau)} + (1 - \alpha) \theta_s^{(\tau + 1)}$, where $\alpha$ controls the update rate.
For the supervised loss on labeled data, we employ a combination of Dice loss and cross-entropy loss to capture both overlap and pixel-wise discrepancies, $\mathcal{L}_{\text{sup}} = \mathcal{L}_{\text{Dice}}(f_{\theta_s}(x^L), y^L) + \mathcal{L}_{\text{CE}}(f_{\theta_s}(x^L), y^L)$.

To leverage the unlabeled data, we enforce consistency between the student and teacher predictions. Specifically, the student receives a strongly augmented version of an unlabeled image $x^U$, while the teacher processes a weakly augmented version. The pixel-wise consistency loss is defined as the cross-entropy between the student and teacher outputs, $\mathcal{L}_{\text{cons}} = \mathcal{L}_{\text{CE}}(f_{\theta_s}(\mathcal{A}_s(x^U)), f_{\theta_t}(\mathcal{A}_w(x^U)))$,
where $\mathcal{A}_s$ and $\mathcal{A}_w$ denote strong and weak augmentations, respectively.

%\vspace{-0.1in}
\subsection{MATCH-Pair: Spatial-Aware Pairwise Matching}
\label{subsec:match-pair}
\begin{figure*}[th]
    \centering 
    \includegraphics[width=1\linewidth]{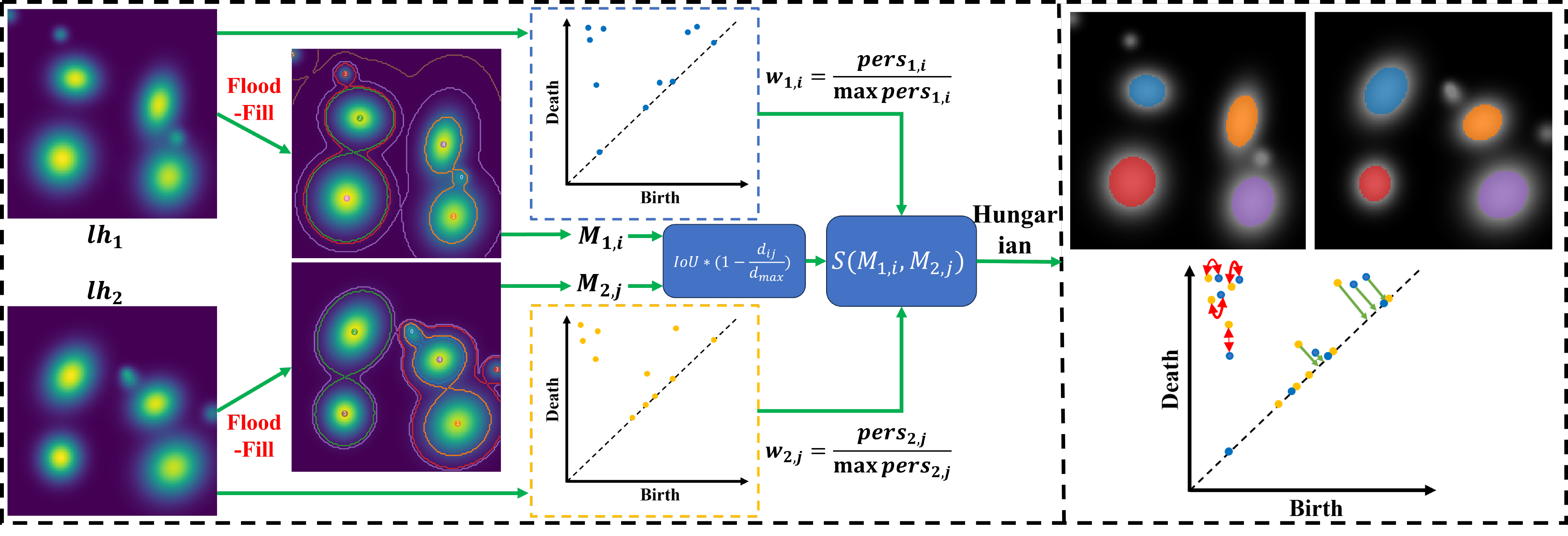}
    \caption{Pipeline of the MATCH-Pair algorithm between two persistence diagrams.}
\label{fig:individual_matching}
\end{figure*}

Accurate identification of corresponding topological structures between the likelihood maps is crucial for robust histopathology image segmentation. 
We employ persistent homology with a \textbf{super-level set filtration} to extract 0-D topological features from likelihood maps, producing persistence diagrams that characterize each component by its persistence and critical points. To find correspondence between different persistence diagrams, traditional methods based on Wasserstein distance emphasize topological persistence without considering spatial relationships, often leading to incorrect associations between spatially distant yet similarly persistent features. In contrast, approaches based solely on spatial overlap tend to match transient structures of minimal significance incorrectly. To address these limitations, we propose MATCH-Pair, a Hungarian overlap-matching algorithm that integrates spatial overlap, topological persistence, and spatial proximity. The overall pipeline is depicted in~\Cref{fig:individual_matching}.

Given two likelihood maps \(lh_1, lh_2 \in [0,1]^{H \times W}\), which are the softmax-activated outputs of the final UNet layer,
we compute the persistence diagrams: $\mathrm{Dgm}(lh_k)=\{(b_i,d_i),\quad k\in\{1,2\}$ with the persistence $\mathrm{pers}_i = |d_i - b_i|$. Each persistence pair $(b_i, d_i)$ yields a connected spatial region $M_i$, defined by flood-fill algorithm~\cite{smith1979tint}. This algorithm generates a binary mask \(M_i\) starting from the birth pixel \(b_i\). The region is expanded iteratively to neighboring pixels, provided that their likelihood exceeds the threshold \(1 - d_i\). 

To compute the relative significance of each structure, the persistence values are normalized to derive a weighting factor: $w_{k,i}=\frac{\mathrm{pers}_{k,i}}{\displaystyle\max_{j}\mathrm{pers}_{k,j}},\quad k\in\{1,2\}$. Here, $i$ and $j$ index the topological features from the $1_{st}$ and $2_{nd}$ persistence diagrams respectively, where $i\in1,...,n_1$ and $j\in1,...,n_2$ with $n_1$ and $n_2$ being the number of features in each diagram. $k$ distinguishes between the two likelihood maps being compared. The notation $w_{k,i}$ refers to the normalized persistence weight of the $i$-th topological feature in the $k$-th likelihood map.

To evaluate the similarity between spatial masks \(M_{1,i}\) and \(M_{2,j}\), a combined metric that integrates spatial overlap, normalized persistence weights, and spatial proximity is defined as:
\begin{equation*}
S_{ij} = \,w_{1,i}\,w_{2,j}\frac{|M_{1,i}\cap M_{2,j}|}{|M_{1,i}\cup M_{2,j}|}\left(1 - \frac{d_{ij}}{d_{\max}}\right)
\end{equation*}
where \(d_{ij}\) is the Euclidean distance between birth critical points of the corresponding masks, and \(d_{\max}\) denotes the maximum distance among all mask pairs. This similarity metric ensures the prioritization of spatially close, persistent, and well-overlapping structures.

A global one-to-one assignment between features from the two maps is obtained via the Hungarian algorithm~\cite{kuhn1955hungarian}, minimizing the cost (defined as the complement of similarity):
\begin{equation*}
\min_{\pi_{ij}}\sum_{i,j}(1 - S_{ij})\,\pi_{ij},\quad \pi_{ij}\in\{0,1\}
\end{equation*}
Pairs achieving scores above a predefined threshold \(\tau_{\mathrm{primary}}\) constitute valid matches.

\subsection{MATCH-Global: Multi-faceted Global Matching}
\label{subsec:global_matching}
While MATCH-Pair addresses an optimal correspondence between two persistence diagrams, many practical scenarios often involve multiple stochastic predictions (facets). Finding the corresponding topological structures among multiple facets is a challenge. Thus, we extend MATCH-Pair to MATCH-Global, a global multi-facet matching approach to link homologous $0$-dimensional components consistently across all facets, assigning global indices to anatomical or topological structures.

Given a series of likelihood maps \(\mathcal{L}=\{lh_t\}_{t=1}^{T}\), \(lh_t\in[0,1]^{H\times W}\), each generates a persistence diagram: $\mathrm{Dgm}_t
=\{(b_{t,i},d_{t,i})\}_{i=1}^{n_t}$. Each pair $(b_{t,i},d_{t,i})$ corresponds to a spatial mask $M_{t,i}$, the normalized persistence weight $w_{t,i}=|d_{t,i}-b_{t,i}|\,/\!\max_{t',j}|d_{t',j}-b_{t',j}|$, and birth-critical point $c_{t,i}$.

Matching is performed sequentially across facets. For each adjacent pair of facets \((t,t+1)\) we form the weighted overlap matrix:
\begin{equation*}
S_{ij}^{(t)}
= \operatorname{w_{t,i}\,w_{t+1,j}IoU}\!\bigl(M_{t,i},M_{t+1,j}\bigr)\Bigl(1-\tfrac{\lVert c_{t,i}-c_{t+1,j}\rVert_2}{d_{\max}^{(t)}}\Bigr)
\end{equation*}
with 
\(d_{\max}^{(t)}=\max_{i,j}\lVert c_{t,i}-c_{t+1,j}\rVert_2\)
introduces a soft spatial penalty.
Optimal assignments are solved via the proposed MATCH-Pair algorithm.

These matches form an undirected graph $G=(\mathcal{V},\mathcal{E})$ with vertices $\mathcal{V}=\{(t,i)\mid 1\le t\le T,\;1\le i\le n_t\}$, representing the structures and edges $\mathcal{E}=\bigcup_{t=1}^{T-1}\mathcal{E}^{(t)}$ indicating matches.
Connected components $\{\mathcal{C}_k\}_{k=1}^{K}$ of $G$ are identified by breadth-first search, providing globally consistent identities:
\(
\mathcal{C}_k
=\{(t,i)\mid\text{mask }M_{t,i}\text{ belongs to identity }k\}.
\)

Thus, the global multi-facet matching framework integrates pairwise correspondences into globally coherent tracks, robustly accommodating missing detections, splits, and merges, thereby ensuring topological consistency across multiple facets.

\subsection{Dual-Level Topological Consistency}
\label{subsec:topo_losses}
After identifying consistent topological structures across multiple facets, we propose dual-level topological consistency losses to enhance segmentation reliability and coherence. Specifically, we introduce two complementary loss terms: \textit{intra-topological consistency}, which ensures consistency among stochastic predictions from MC dropout realizations~\cite{gal2016dropout}, and \textit{temporal-topological consistency}, which maintains consistency across consecutive training iterations.

In both scenarios, topological features are extracted from multiple prediction facets. We then apply our proposed MATCH-Global algorithm (see~\Cref{subsec:global_matching}) to classify these topological structures into two distinct categories: \textbf{matched} ($\mathcal{C}^{\text{match}}_{\text{intra}}$, $\mathcal{C}^{\text{match}}_{\text{temp}}$), representing features consistently identified across multiple predictions, and \textbf{unmatched} ($\mathcal{C}^{\text{unmatch}}_{\text{intra}}$, $\mathcal{C}^{\text{unmatch}}_{\text{temp}}$), denoting features that are inconsistent or unstable across predictions.
Specifically, matched structures are encouraged toward optimal probability distributions at their birth and death critical points, whereas unmatched structures, indicative of prediction uncertainty or instability, are driven toward shorter topological lifespans. Formally, we define the associated losses as:
\begin{equation*}
\mathcal{L}_{\text{match}}(t,i) = \bigl(P^{(t)}_{b_{t,i}}\bigr)^2 + \bigl(1 - P^{(t)}_{d_{t,i}}\bigr)^2,\quad
\mathcal{L}_{\text{diag}}(t,i)  = \bigl(P^{(t)}_{b_{t,i}} - P^{(t)}_{d_{t,i}}\bigr)^2.
\end{equation*}
where $P^{(t)}_{b_{t,i}}$ and $P^{(t)}_{d_{t,i}}$ represent the predicted probability values at the birth ($b_{t,i}$) and death ($d_{t,i}$) critical points, respectively, of the $i$-th topological feature extracted from the $t$-th prediction.

The intra-topological consistency loss aggregates these penalties over multiple stochastic predictions through MC dropout within each iteration:
\begin{equation*}
    \mathcal{L}_{\text{intra}} = \frac{1}{B_{\text{intra}}}\sum_{b=1}^{B_{\text{intra}}} \left[
    \frac{1}{|\mathcal{C}^{\text{match}, (b)}_{\text{intra}}|}\sum_{(t,i)\in\mathcal{C}^{\text{match}, (b)}_{\text{intra}}}\mathcal{L}_{\text{match}}(t,i) + \frac{1}{|\mathcal{C}_{\text{intra}}^{\text{unmatch},(b)}|}\sum_{(t,i)\in\mathcal{C}_{\text{intra}}^{\text{unmatch},(b)}}\mathcal{L}_{\text{diag}}(t,i)
    \right]
\end{equation*}
where $B_{\text{intra}}$ indicates the number of MC dropout predictions within each iteration.
Similarly, the temporal-topological consistency enforces the constraints across predictions from consecutive training snapshots:
\begin{equation*}
    \mathcal{L}_{\text{temp}} = \frac{1}{B_{\text{temp}}}\sum_{b=1}^{B_{\text{temp}}} \left[
    \frac{1}{|\mathcal{C}^{\text{match}, (b)}_{\text{temp}}|}\sum_{(t,i)\in\mathcal{C}^{\text{match}, (b)}_{\text{temp}}}\mathcal{L}_{\text{match}}(t,i) + \frac{1}{|\mathcal{C}_{\text{temp}}^{\text{unmatch},(b)}|}\sum_{(t,i)\in\mathcal{C}_{\text{temp}}^{\text{unmatch},(b)}}\mathcal{L}_{\text{diag}}(t,i)
    \right]
\end{equation*}
where $B_{\text{temp}}$ presents the number of temporal training snapshots.
Finally, our dual-level topological consistency losses are integrated into the overall training objective alongside the supervised and pixel-wise consistency terms:
\begin{equation*}
    \mathcal{L}_{\text{total}} = \mathcal{L}_{\text{sup}} + \lambda_{\text{cons}}\mathcal{L}_{\text{cons}} + \lambda_{\text{intra}}\mathcal{L}_{\text{intra}} + \lambda_{\text{temp}}\mathcal{L}_{\text{temp}}
\end{equation*}
where hyperparameters $\lambda_{\text{cons}}$, $\lambda_{\text{intra}}$, and $\lambda_{\text{temp}}$ balance their respective contributions, ensuring the model jointly meets pixel-level accuracy and robust topological coherence.

\section{Experiments}
We conduct comprehensive evaluations on three publicly available histopathology image datasets on both pixel-wise and topology-wise metrics. We benchmark our method against classic and recent state-of-the-art semi-supervised segmentation methods, including MT~\cite{tarvainen2017mean}, EM~\cite{vu2019advent}, UA-MT~\cite{yu2019uncertainty}, URPC~\cite{luo2022semi}, XNet~\cite{zhou2023xnet}, PMT~\cite{gao2024pmt}, and TopoSemiSeg~\cite{xu2024semi}. 

\myparagraph{Implementation Details.} The implementation details will be provided in the Supplementary. 

\begin{table}[t]
\centering
\footnotesize
\caption{
Quantitative results on three histopathology image datasets. We compare our method with several state-of-the-art semi-supervised medical image segmentation methods on two settings of $10\%$ and $20\%$ labeled data. The statistically significant best results are highlighted in \textbf{bold}, while the second-best are marked with \uline{underline}.}
\resizebox{\linewidth}{!}{%
\begin{tabular}{lclccccc}
\toprule
\multirow{2}{*}{Dataset} & \multirow{2}{*}{Label Ratio (\%)} & \multirow{2}{*}{Method} 
& \multicolumn{1}{c}{Pixel-wise} &  & \multicolumn{3}{c}{Topology-wise} \\ 
\cmidrule(lr){4-4}\cmidrule(lr){6-8}
 &  &  & Dice\_Obj $\uparrow$ &  & BE $\downarrow$ & BME $\downarrow$ & DIU $\downarrow$ \\ 
\midrule
% =============================================================  CRAG
\multirow{17}{*}{CRAG} 
 & \multirow{8}{*}{10}  & MT~\cite{tarvainen2017mean}        & 0.821 $\pm$ 0.006 &  & 2.238 $\pm$ 0.153 & 62.250 $\pm$ 3.127 & 74.630 $\pm$ 2.967 \\
 &                       & EM~\cite{vu2019advent}             & 0.789 $\pm$ 0.007 &  & 2.178 $\pm$ 0.147 & 80.100 $\pm$ 3.809 & 78.210 $\pm$ 3.298 \\
 &                       & UA-MT~\cite{yu2019uncertainty}     & 0.837 $\pm$ 0.005 &  & 1.703 $\pm$ 0.112 & 66.450 $\pm$ 3.218 & 65.420 $\pm$ 2.847 \\
 &                       & URPC~\cite{yu2019uncertainty}      & 0.829 $\pm$ 0.005 &  & 1.732 $\pm$ 0.118 & 74.600 $\pm$ 3.407 & 68.300 $\pm$ 3.004 \\
 &                       & XNet~\cite{zhou2023xnet}           & 0.872 $\pm$ 0.004 &  & 0.578 $\pm$ 0.053 & 15.050 $\pm$ 1.118 & 55.880 $\pm$ 2.516 \\
 &                       & PMT~\cite{gao2024pmt}              & 0.876 $\pm$ 0.004 &  & 0.520 $\pm$ 0.051 & 14.200 $\pm$ 1.013 & 57.100 $\pm$ 2.638 \\
 &                       & TopoSemiSeg~\cite{xu2024semi}      & \uline{0.884 $\pm$ 0.002} &  & \uline{0.227 $\pm$ 0.014} & \uline{10.475 $\pm$ 0.458} & \uline{49.690 $\pm$ 1.947} \\
 &                       & \textbf{Ours}                      & \textbf{0.888 $\pm$ 0.002} &  & \textbf{0.197 $\pm$ 0.012} & \textbf{9.175 $\pm$ 0.580}  & \textbf{45.950 $\pm$ 1.880} \\
\cmidrule(lr){2-8}
 & \multirow{8}{*}{20}  & MT~\cite{tarvainen2017mean}        & 0.858 $\pm$ 0.008 &  & 2.603 $\pm$ 0.161 & 99.025 $\pm$ 3.912 & 95.215 $\pm$ 3.487 \\
 &                       & EM~\cite{vu2019advent}             & 0.869 $\pm$ 0.006 &  & 1.933 $\pm$ 0.136 & 75.225 $\pm$ 3.772 & 63.823 $\pm$ 3.139 \\
 &                       & UA-MT~\cite{yu2019uncertainty}     & 0.859 $\pm$ 0.006 &  & 1.822 $\pm$ 0.129 & 70.850 $\pm$ 3.586 & 61.138 $\pm$ 2.918 \\
 &                       & URPC~\cite{luo2022semi}            & 0.849 $\pm$ 0.007 &  & 2.489 $\pm$ 0.152 & 99.500 $\pm$ 4.085 & 87.681 $\pm$ 3.276 \\
 &                       & XNet~\cite{zhou2023xnet}           & 0.883 $\pm$ 0.005 &  & 0.422 $\pm$ 0.055 & 10.900 $\pm$ 1.127 & 50.537 $\pm$ 2.547 \\
 &                       & PMT~\cite{gao2024pmt}              & 0.889 $\pm$ 0.004 &  & 0.460 $\pm$ 0.062 & 11.800 $\pm$ 1.203 & 48.300 $\pm$ 2.321 \\
 &                       & TopoSemiSeg~\cite{xu2024semi}      & \uline{0.898 $\pm$ 0.004} &  & \uline{0.226 $\pm$ 0.019} & \uline{8.575 $\pm$ 0.736} & \uline{43.712 $\pm$ 1.842} \\
 &                       & \textbf{Ours}                      & \textbf{0.909 $\pm$ 0.005} &  & \textbf{0.188 $\pm$ 0.018} & \textbf{7.425 $\pm$ 0.570} & \textbf{40.250 $\pm$ 1.720} \\
\cmidrule(lr){2-8}
 & 100 (Full)           & Fully-Supervised                   & 0.928 $\pm$ 0.002 &  & 0.149 $\pm$ 0.015 & 5.650 $\pm$ 0.223 & 29.425 $\pm$ 1.782 \\
\midrule
% =============================================================  GlaS
\multirow{17}{*}{GlaS}
 & \multirow{8}{*}{10}  & MT~\cite{tarvainen2017mean}        & 0.790 $\pm$ 0.005 &  & 2.392 $\pm$ 0.162 & 31.125 $\pm$ 3.274 & 76.130 $\pm$ 2.965 \\
 &                       & EM~\cite{vu2019advent}             & 0.819 $\pm$ 0.006 &  & 1.431 $\pm$ 0.143 & 19.188 $\pm$ 3.846 & 61.245 $\pm$ 3.302 \\
 &                       & UA-MT~\cite{yu2019uncertainty}     & 0.845 $\pm$ 0.004 &  & 2.086 $\pm$ 0.117 & 26.650 $\pm$ 3.245 & 68.025 $\pm$ 2.873 \\
 &                       & URPC~\cite{yu2019uncertainty}      & 0.849 $\pm$ 0.004 &  & 1.155 $\pm$ 0.123 & 19.588 $\pm$ 3.408 & 54.832 $\pm$ 3.017 \\
 &                       & XNet~\cite{zhou2023xnet}           & 0.874 $\pm$ 0.003 &  & 0.843 $\pm$ 0.051 & 14.238 $\pm$ 1.154 & 40.912 $\pm$ 2.422 \\
 &                       & PMT~\cite{gao2024pmt}              & 0.872 $\pm$ 0.004 &  & 0.798 $\pm$ 0.052 & 13.920 $\pm$ 1.097 & 39.850 $\pm$ 2.487 \\
 &                       & TopoSemiSeg~\cite{xu2024semi}      & \uline{0.878 $\pm$ 0.003} &  & \uline{0.551 $\pm$ 0.014} & \uline{8.300 $\pm$ 0.478}  & \uline{35.845 $\pm$ 1.965} \\
 &                       & \textbf{Ours}                      & \textbf{0.884 $\pm$ 0.003} &  & \textbf{0.501 $\pm$ 0.023} & \textbf{7.850 $\pm$ 0.391} & \textbf{30.525 $\pm$ 1.641} \\
\cmidrule(lr){2-8}
 & \multirow{8}{*}{20}  & MT~\cite{tarvainen2017mean}        & 0.863 $\pm$ 0.005 &  & 2.126 $\pm$ 0.171 & 29.963 $\pm$ 3.987 & 64.275 $\pm$ 3.496 \\
 &                       & EM~\cite{vu2019advent}             & 0.865 $\pm$ 0.006 &  & 1.255 $\pm$ 0.138 & 17.275 $\pm$ 3.783 & 58.673 $\pm$ 3.255 \\
 &                       & UA-MT~\cite{yu2019uncertainty}     & 0.866 $\pm$ 0.005 &  & 1.123 $\pm$ 0.132 & 18.038 $\pm$ 3.599 & 53.014 $\pm$ 3.069 \\
 &                       & URPC~\cite{luo2022semi}            & 0.878 $\pm$ 0.004 &  & 0.759 $\pm$ 0.067 & 14.350 $\pm$ 1.212 & 42.587 $\pm$ 2.601 \\
 &                       & XNet~\cite{zhou2023xnet}           & 0.884 $\pm$ 0.004 &  & 0.735 $\pm$ 0.065 & 10.188 $\pm$ 1.154 & 35.298 $\pm$ 2.328 \\
 &                       & PMT~\cite{gao2024pmt}              & 0.887 $\pm$ 0.003 &  & 0.698 $\pm$ 0.062 & 9.980 $\pm$ 1.118 & 34.805 $\pm$ 2.271 \\
 &                       & TopoSemiSeg~\cite{xu2024semi}      & \textbf{0.895 $\pm$ 0.003} &  & \uline{0.510 $\pm$ 0.053} & \uline{9.825 $\pm$ 0.813} & \uline{30.462 $\pm$ 1.978} \\
 &                       & \textbf{Ours}                      & \textbf{0.894 $\pm$ 0.004} &  & \textbf{0.392 $\pm$ 0.056} & \textbf{7.925 $\pm$ 0.725} & \textbf{26.175 $\pm$ 1.633} \\
\cmidrule(lr){2-8}
 & 100 (Full)           & Fully-Supervised                   & 0.917 $\pm$ 0.006 &  & 0.273 $\pm$ 0.026 & 6.875 $\pm$ 0.276 & 19.620 $\pm$ 0.712 \\
\midrule
% =============================================================  MoNuSeg
\multirow{17}{*}{MoNuSeg}
 & \multirow{8}{*}{10}  & MT~\cite{tarvainen2017mean}        & 0.748 $\pm$ 0.006 &  & 10.210 $\pm$ 0.486 & 292.857 $\pm$ 6.542 & 1526.079 $\pm$ 35.842 \\
 &                       & EM~\cite{vu2019advent}             & 0.757 $\pm$ 0.006 &  & 10.339 $\pm$ 0.503 & 257.071 $\pm$ 5.445 & 1319.815 $\pm$ 31.784 \\
 &                       & UA-MT~\cite{yu2019uncertainty}     & 0.741 $\pm$ 0.007 &  & 10.227 $\pm$ 0.497 & 255.428 $\pm$ 5.983 & 1316.272 $\pm$ 30.216 \\
 &                       & URPC~\cite{yu2019uncertainty}      & 0.774 $\pm$ 0.004 &  & 6.829 $\pm$ 0.319 & 214.428 $\pm$ 5.327 & 1098.372 $\pm$ 24.392 \\
 &                       & XNet~\cite{zhou2023xnet}           & 0.762 $\pm$ 0.005 &  & 7.152 $\pm$ 0.338 & 220.405 $\pm$ 4.611 & 1122.799 $\pm$ 25.116 \\
 &                       & PMT~\cite{gao2024pmt}              & 0.764 $\pm$ 0.004 &  & 7.515 $\pm$ 0.352 & 227.650 $\pm$ 4.805 & 1210.400 $\pm$ 26.954 \\
 &                       & TopoSemiSeg~\cite{xu2024semi}      & \textbf{0.783 $\pm$ 0.003} &  & \uline{6.661 $\pm$ 0.376} & \uline{196.357 $\pm$ 3.067} & \uline{1068.401 $\pm$ 17.500} \\
 &                       & \textbf{Ours}                      & \textbf{0.785 $\pm$ 0.003} &  & \textbf{5.594 $\pm$ 0.361} & \textbf{192.863 $\pm$ 1.137} & \textbf{1011.857 $\pm$ 12.648} \\
\cmidrule(lr){2-8}
 & \multirow{8}{*}{20}  & MT~\cite{tarvainen2017mean}        & 0.767 $\pm$ 0.005 &  & 12.522 $\pm$ 0.547 & 246.786 $\pm$ 8.018 & 1350.751 $\pm$ 32.407 \\
 &                       & EM~\cite{vu2019advent}             & 0.777 $\pm$ 0.006 &  & 7.160 $\pm$ 0.335 & 198.571 $\pm$ 6.731 & 1142.661 $\pm$ 27.581 \\
 &                       & UA-MT~\cite{yu2019uncertainty}     & 0.772 $\pm$ 0.007 &  & 9.406 $\pm$ 0.444 & 246.857 $\pm$ 7.944 & 1336.684 $\pm$ 31.268 \\
 &                       & URPC~\cite{luo2022semi}            & 0.779 $\pm$ 0.004 &  & 5.325 $\pm$ 0.254 & 193.429 $\pm$ 6.105 & 1025.431 $\pm$ 23.799 \\
 &                       & XNet~\cite{zhou2023xnet}           & 0.776 $\pm$ 0.003 &  & 6.750 $\pm$ 0.316 & 198.525 $\pm$ 5.421 & 1117.406 $\pm$ 26.014 \\
 &                       & PMT~\cite{gao2024pmt}              & 0.778 $\pm$ 0.006 &  & 6.500 $\pm$ 0.308 & 195.125 $\pm$ 6.289 & 1080.476 $\pm$ 25.145 \\
 &                       & TopoSemiSeg~\cite{xu2024semi}      & \textbf{0.793 $\pm$ 0.004} &  & \uline{5.150 $\pm$ 0.145} & \uline{188.642 $\pm$ 3.215} & \uline{1105.946 $\pm$ 18.486} \\
 &                       & \textbf{Ours}                      & \textbf{0.790 $\pm$ 0.006} &  & \textbf{4.930 $\pm$ 0.156} & \textbf{179.225 $\pm$ 2.383} & \textbf{982.286 $\pm$ 14.953} \\
\cmidrule(lr){2-8}
 & 100 (Full)           & Fully-Supervised                   & 0.817 $\pm$ 0.010 &  & 2.491 $\pm$ 0.460 & 142.429 $\pm$ 4.674 & 729.017 $\pm$ 17.662 \\
\bottomrule
%\vspace{-.1in}
\end{tabular}}
\label{tab:quant}
%\vspace{-.2in}
\end{table}

\myparagraph{Datasets}.
We evaluate our proposed method on \textbf{Colorectal Adenocarcinoma Gland (CRAG)}~\cite{graham2019mild}, \textbf{Gland Segmentation in Colon Histology Images Challenge (GlaS)}~\cite{sirinukunwattana2017gland}, and \textbf{Multi-Organ Nuclei Segmentation (MoNuSeg)}~\cite{kumar2019multi}. More details are provided in the Supplementary.

\myparagraph{Evaluation Metrics}.
To better evaluate our proposed method, we use pixel-wise metrics including \textbf{Object-level Dice Score (Dice\_obj)}~\cite{xie2019deep}; topology-wise metrics including \textbf{Betti Error}~\cite{hu2019topology}, \textbf{Betti Matching Error}~\cite{stucki2023topologically}, and \textbf{Discrepancy between Intersection and Union (DIU)}~\cite{lux2024topograph}. More details are provided in the Supplementary.

%\vspace{-.05in}
\subsection{Results}
\begin{figure*}[t]
  \centering
  % ------------------------------------------------------------------
  % Compact spacing between images
  \setlength{\tabcolsep}{1pt}
  \renewcommand{\arraystretch}{0}
  % ------------------------------------------------------------------
  \begin{tabular}{cccccc}
    % ---------------- 1st row (input + 4 likelihoods + majority vote)
    \includegraphics[width=.155\textwidth]{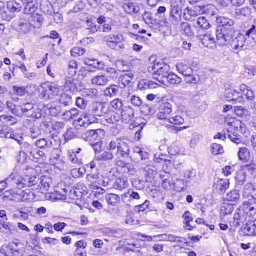}          &
    \includegraphics[width=.155\textwidth]{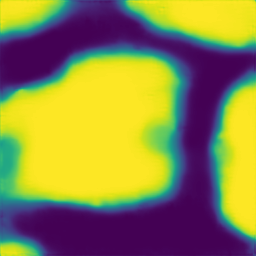}   &
    \includegraphics[width=.155\textwidth]{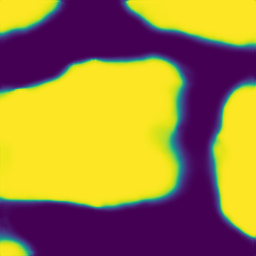}   &
    \includegraphics[width=.155\textwidth]{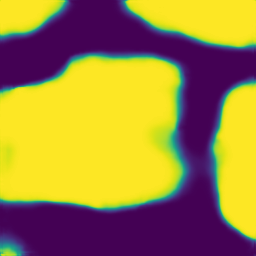}   &
    \includegraphics[width=.155\textwidth]{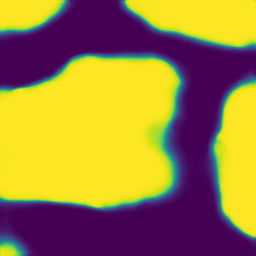}   &
    \includegraphics[width=.155\textwidth]{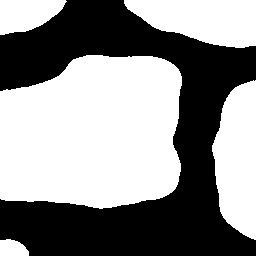}  \\[2pt] % ← space between 1st & 2nd lines
    % ---------------- 2nd row (GT + 4 error maps + uncertainty)
    \includegraphics[width=.155\textwidth]{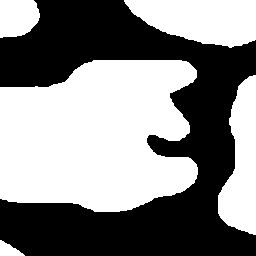}             &
    \includegraphics[width=.155\textwidth]{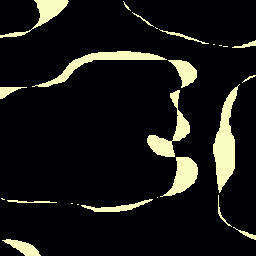}        &
    \includegraphics[width=.155\textwidth]{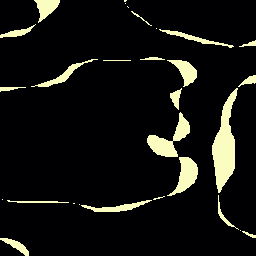}        &
    \includegraphics[width=.155\textwidth]{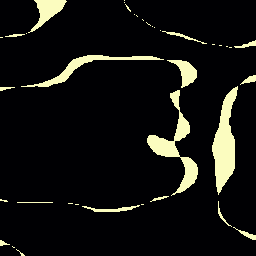}        &
    \includegraphics[width=.155\textwidth]{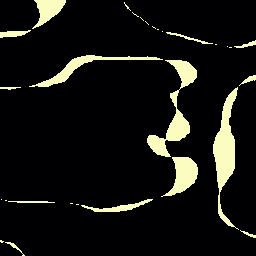}        &
    \includegraphics[width=.155\textwidth]{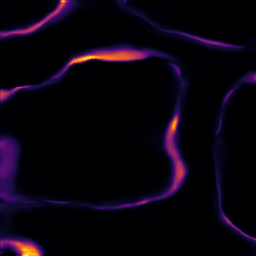}    \\[2pt]
    \multicolumn{1}{c}{\scriptsize (a)  Original Img / GT}        &
    \multicolumn{1}{c}{\scriptsize (b) $lh_1 / err_1$ }  &
    \multicolumn{1}{c}{\scriptsize (c) $lh_2 / err_2$ }  &
    \multicolumn{1}{c}{\scriptsize (d) $lh_3 / err_3$ }  &
    \multicolumn{1}{c}{\scriptsize (e) $lh_4 / err_4$ }  &
    \multicolumn{1}{c}{\scriptsize (f) Result / Uncertainty}   \\
  \end{tabular}
  %\vspace{2pt} % ← space between the table and the caption
  \caption{Qualitative illustration of MC dropout predictions (after the model convergence).
           \textbf{Top\, row:} original patch, the four likelihood
           maps, and the final segmentation. \textbf{Bottom\, row:} ground-truth mask, corresponding
           error maps, and the pixel-wise variance (uncertainty) map.}
  \label{fig:mc_dropout_qualitative}
  %\vspace{-.1in}
\end{figure*}

%\vspace{-.05in}
\myparagraph{Uncertainty Throughout the Topological Consistency}.
As illustrated in~\Cref{fig:mc_dropout_qualitative}, our proposed MATCH not only produces a robust segmentation result (top, (f)) but also furnishes an informative pixel-wise uncertainty map without any uncertainty-specific training objective or doing post hoc calibration. Visually, the variance map (bottom, (f)) concentrates along the gland boundaries where the four likelihood maps disagree, and these regions coincide almost perfectly with the binary error maps (bottom, (b) - (e)). Quantitatively, the Pearson correlation coefficients (PCC)~\cite{pearson1895vii} between the uncertainty and the error maps are \textcolor{red}{$0.768$}, \textcolor{red}{$0.728$}, \textcolor{red}{$0.757$}, and \textcolor{red}{$0.753$} for the four facets, respectively. This confirms that the uncertainty is tightly coupled with prediction errors. Hence, reliable uncertainty estimation and the attendant suppression of spurious structures emerge naturally as a by-product of the proposed consistency mechanism, with no additional supervision or model modification required.

\myparagraph{Quantitative Results}.
As shown in~\Cref{tab:quant}, across the three histopathology image datasets, our proposed method consistently achieves superior performance compared to state-of-the-art semi-supervised segmentation methods, under both 10\% and 20\% labeled data settings. Specifically, our method yields higher topology-wise accuracy with comparable pixel-wise performance. These results collectively illustrate that our framework effectively leverages limited annotations to achieve robust segmentation accuracy and enhanced topological fidelity.

\begin{figure*}[t]
  \centering
  % -------------------------------------------------------------
  % Compact spacing between images
  \setlength{\tabcolsep}{1pt}
  \renewcommand{\arraystretch}{0}
  % -------------------------------------------------------------
  % -- images plus column-caption row ---------------------------
  \begin{tabular}{*{10}{>{\centering\arraybackslash}m{.095\textwidth}}}
    % ------------ first row ------------
    \includegraphics[width=\linewidth]{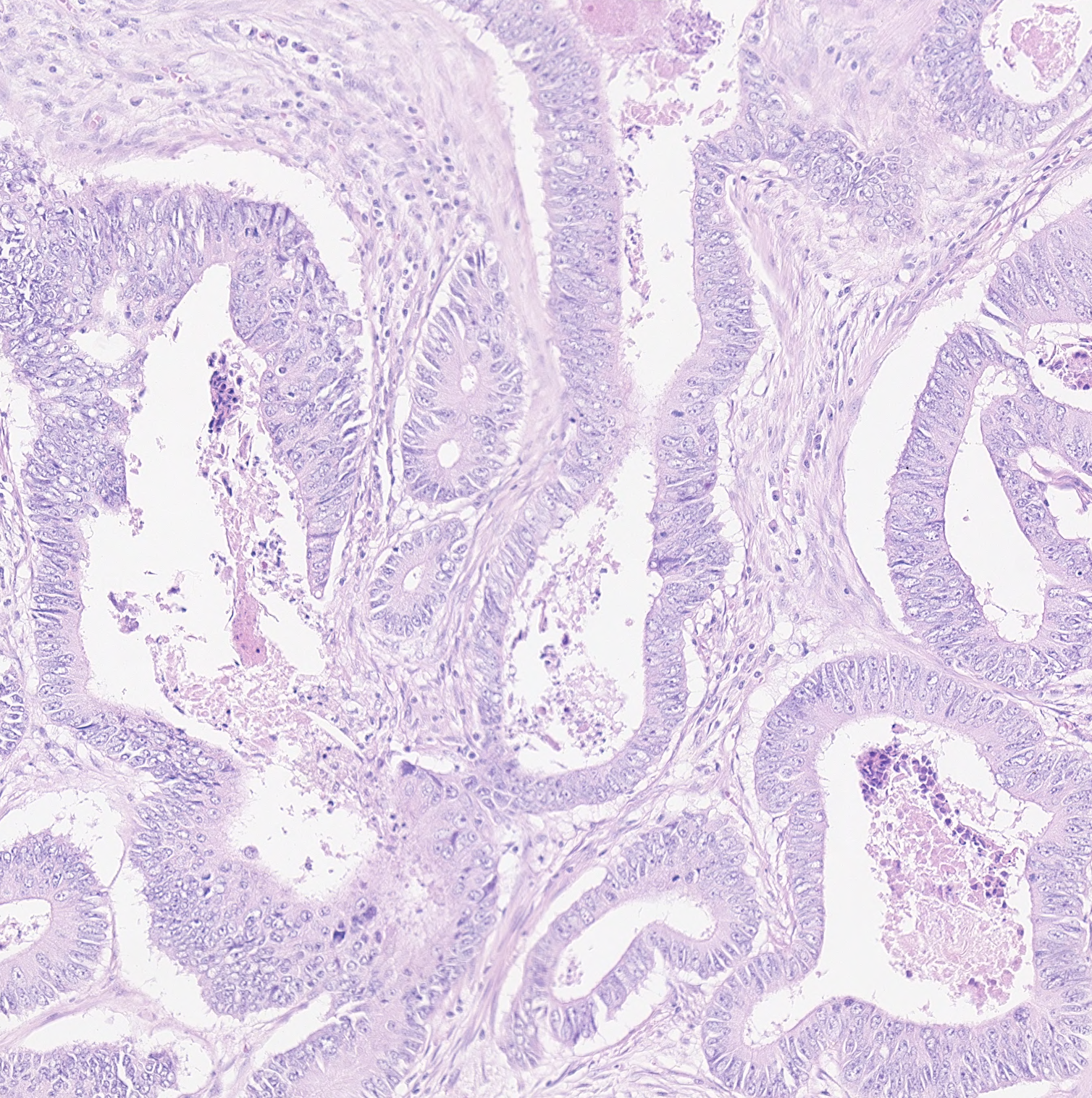}  &
    \includegraphics[width=\linewidth]{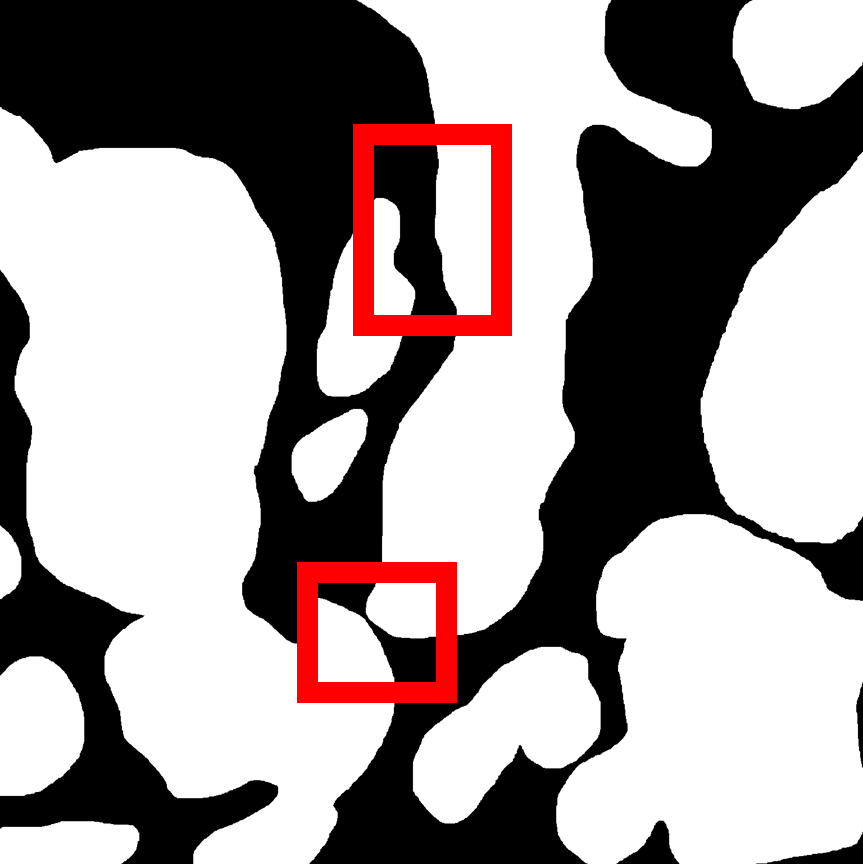}   &
    \includegraphics[width=\linewidth]{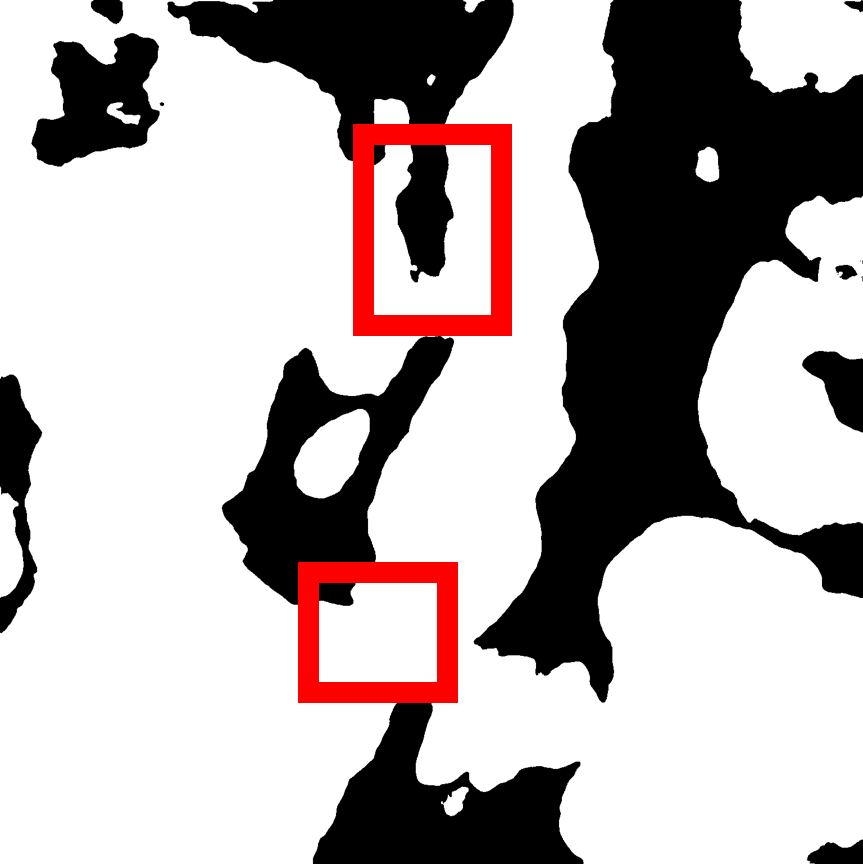}   &
    \includegraphics[width=\linewidth]{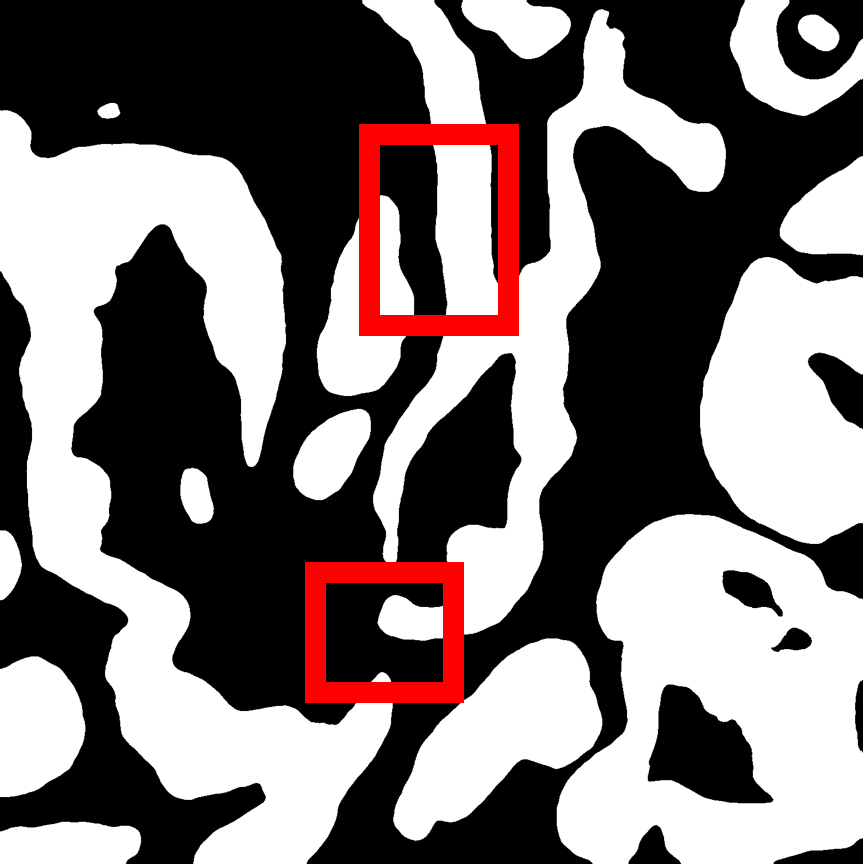}   &
    \includegraphics[width=\linewidth]{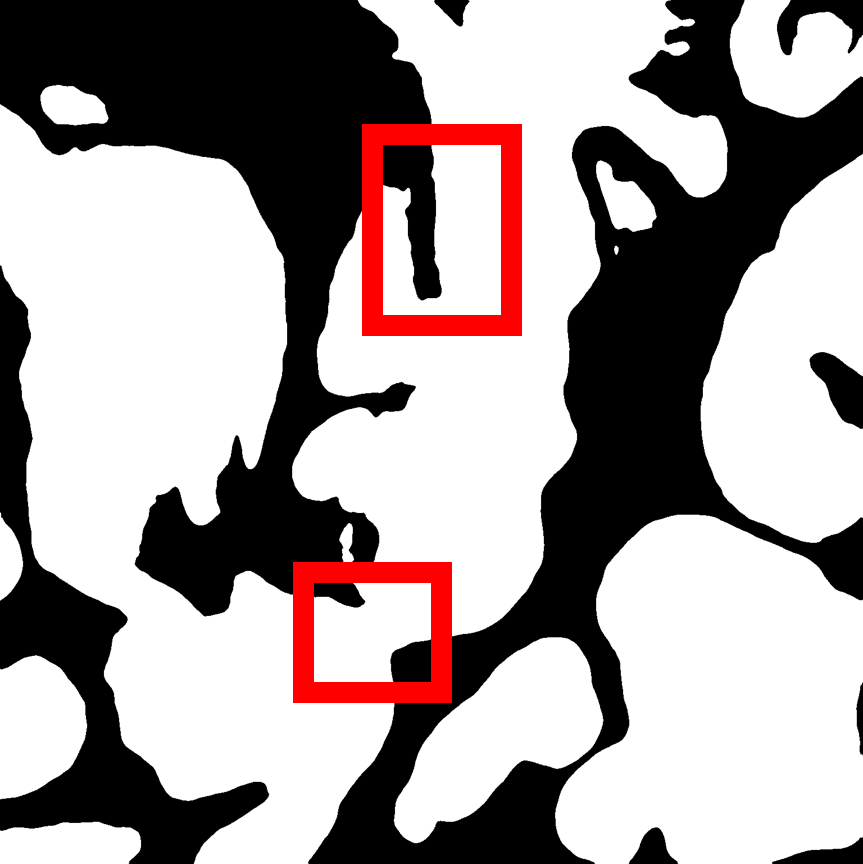} &
    \includegraphics[width=\linewidth]{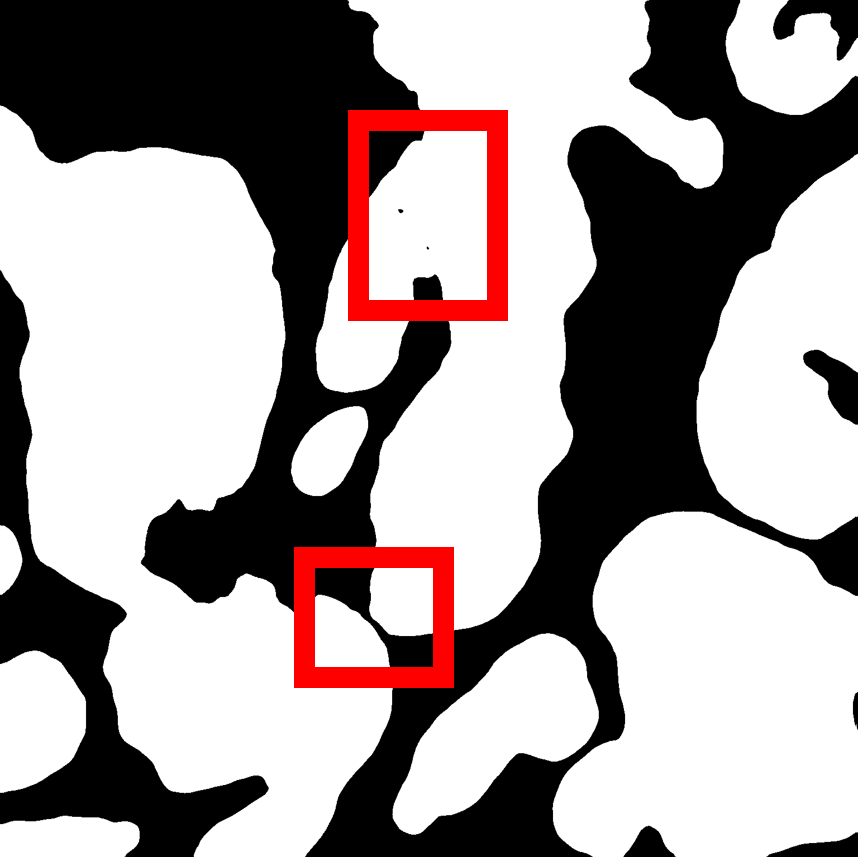}  &
    \includegraphics[width=\linewidth]{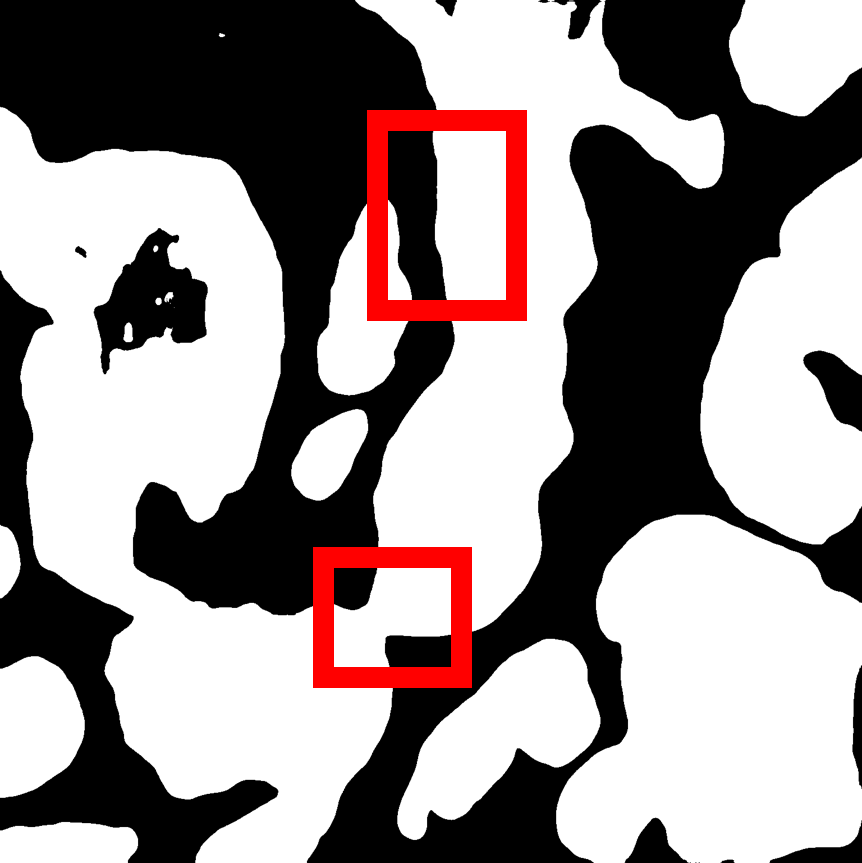} &
    \includegraphics[width=\linewidth]{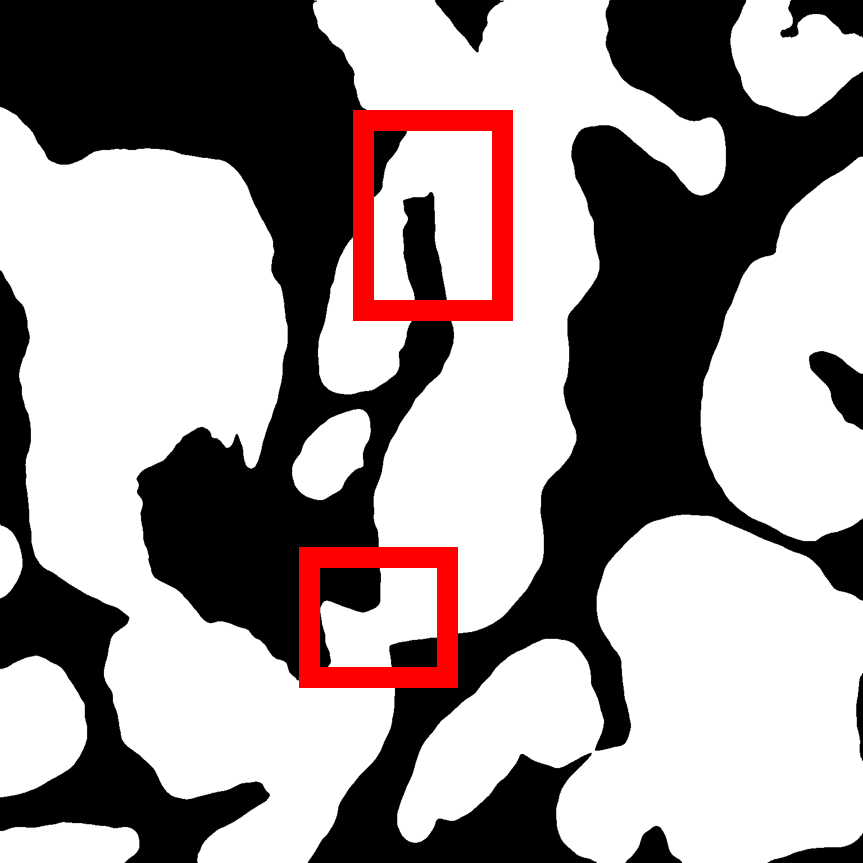} &
    \includegraphics[width=\linewidth]{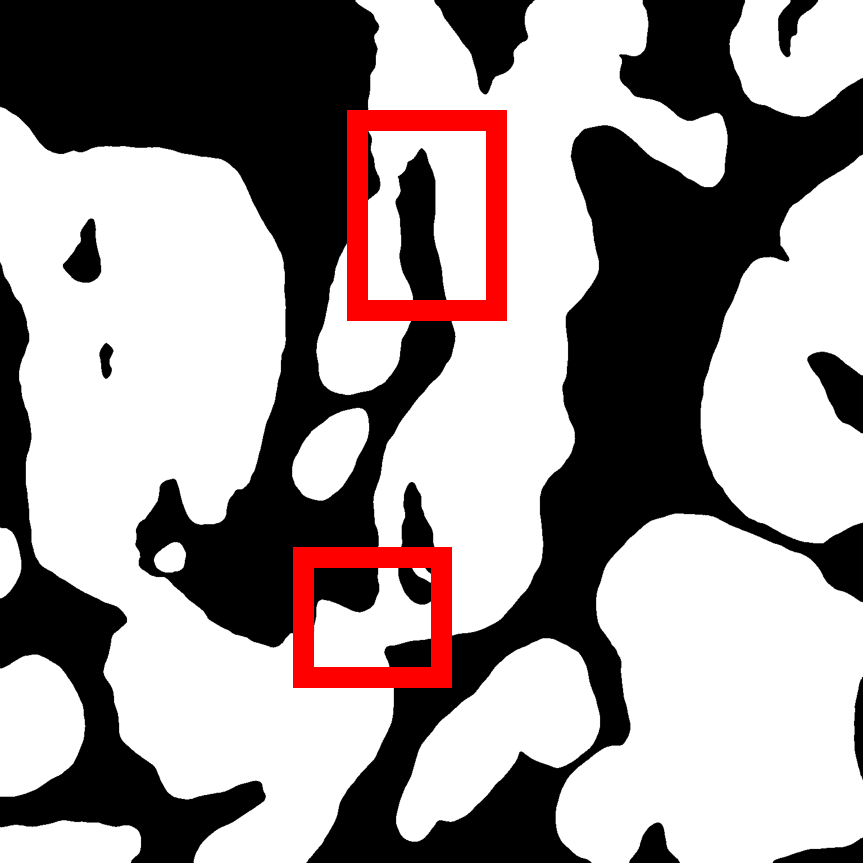}  &
    \includegraphics[width=\linewidth]{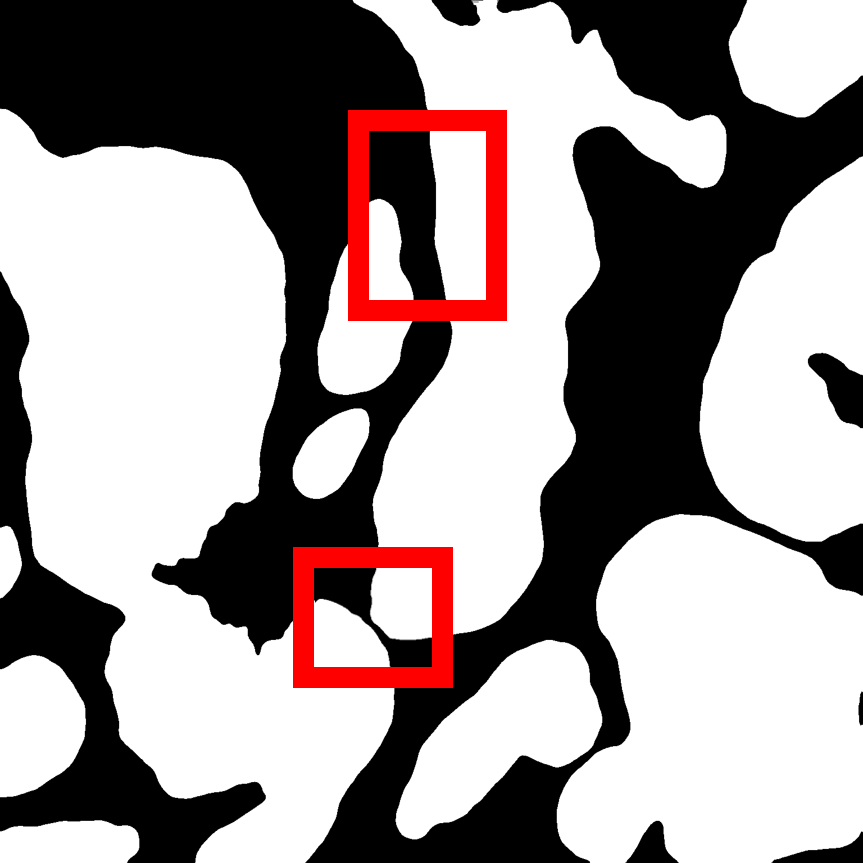} \\[10pt]
    % ------------ second row ----------
    \includegraphics[width=\linewidth]{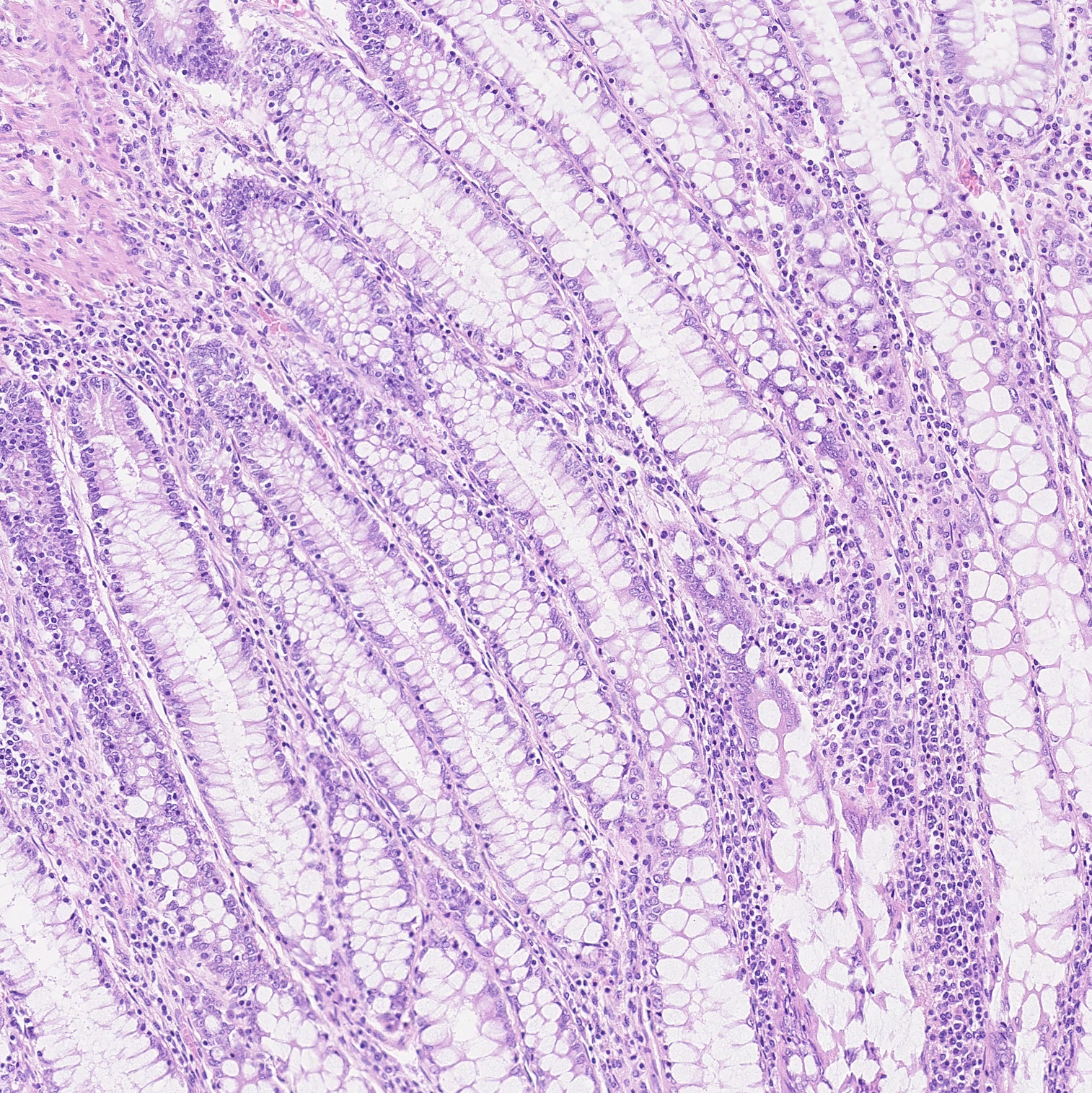}  &
    \includegraphics[width=\linewidth]{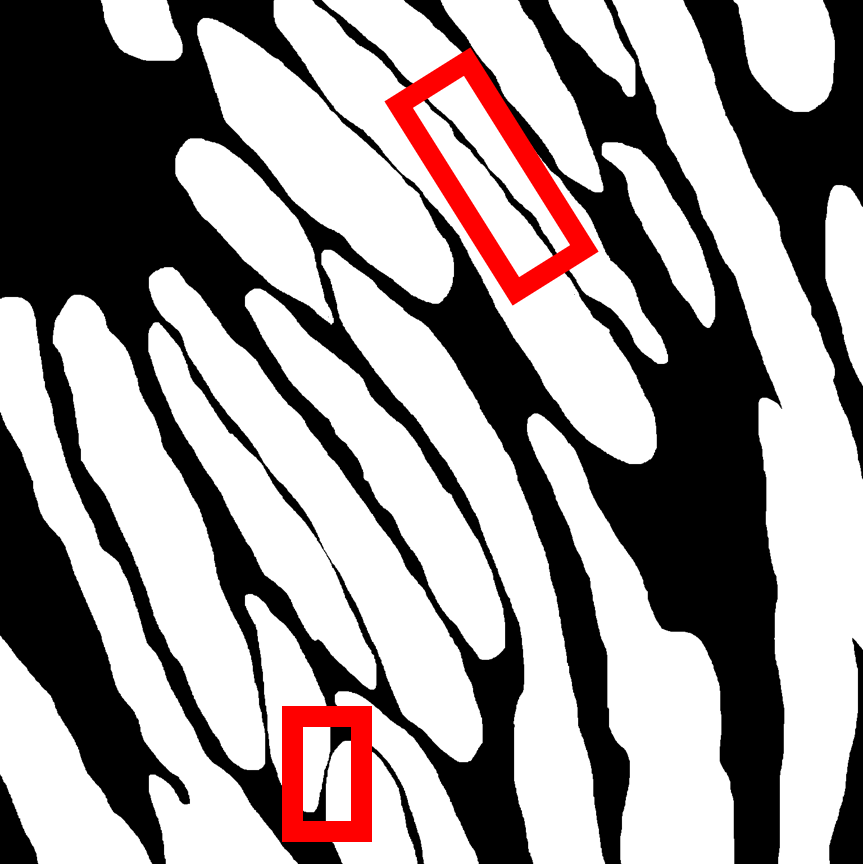}   &
    \includegraphics[width=\linewidth]{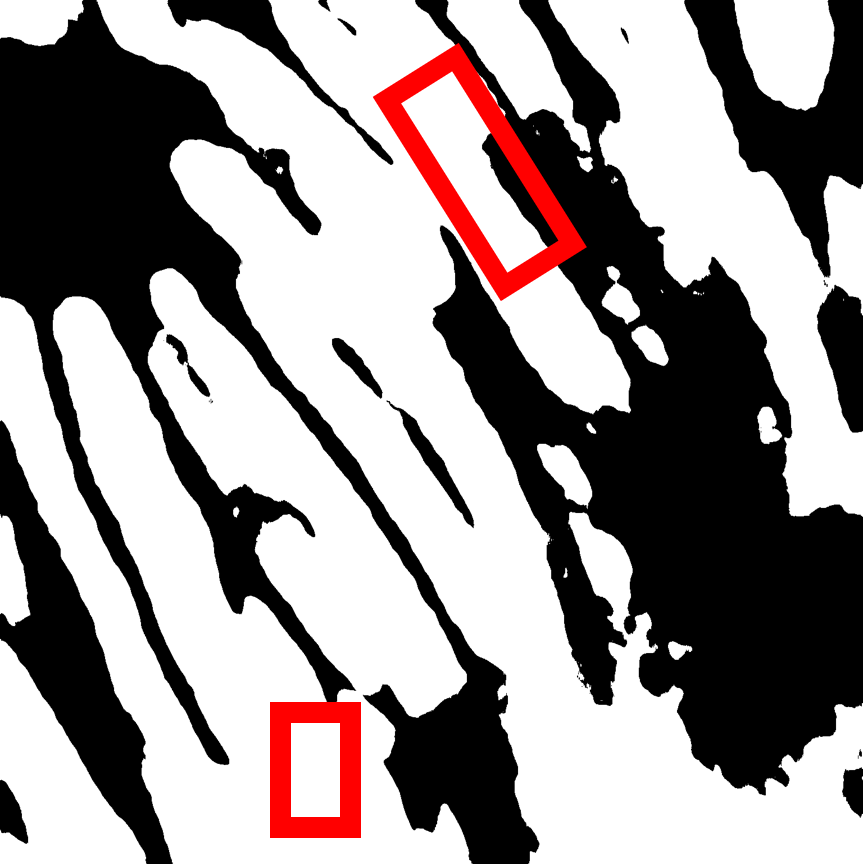}   &
    \includegraphics[width=\linewidth]{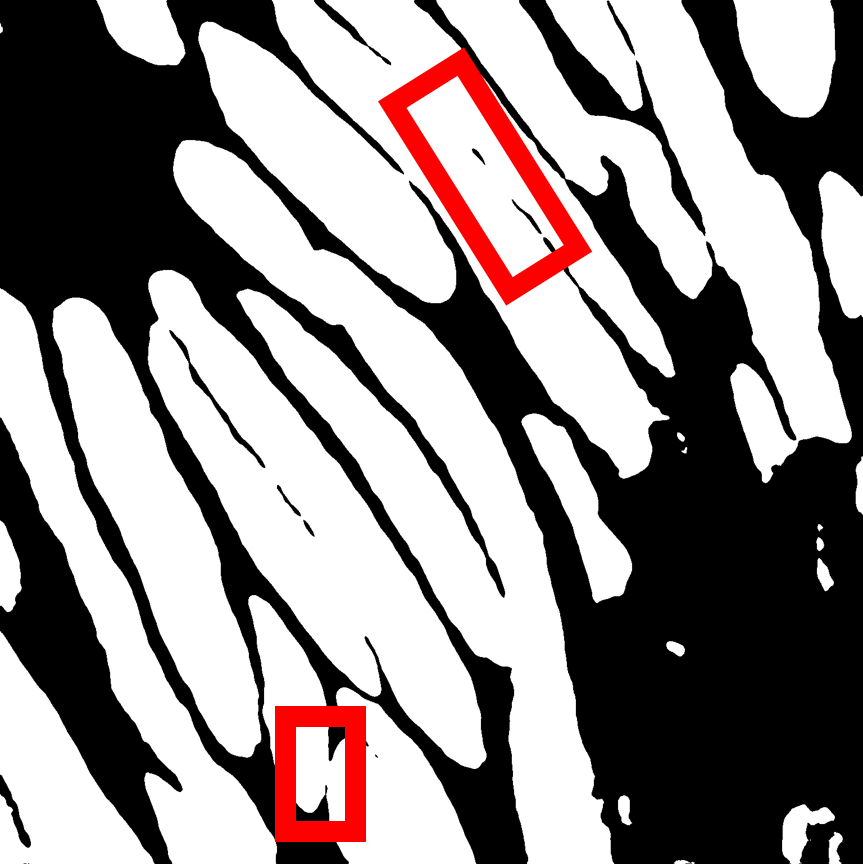}   &
    \includegraphics[width=\linewidth]{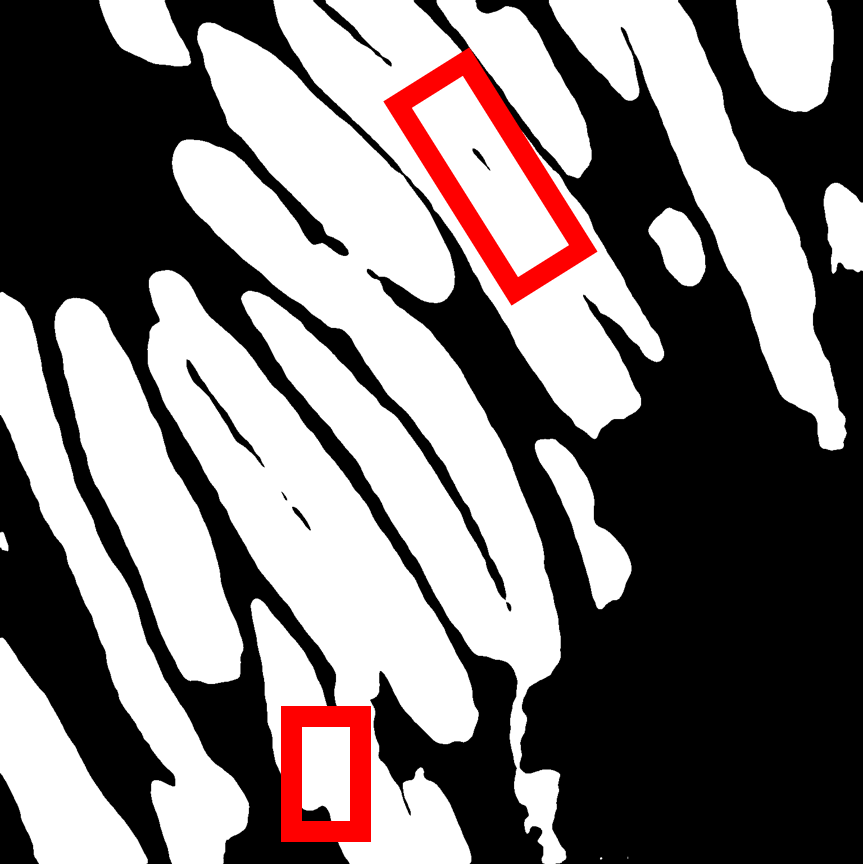} &
    \includegraphics[width=\linewidth]{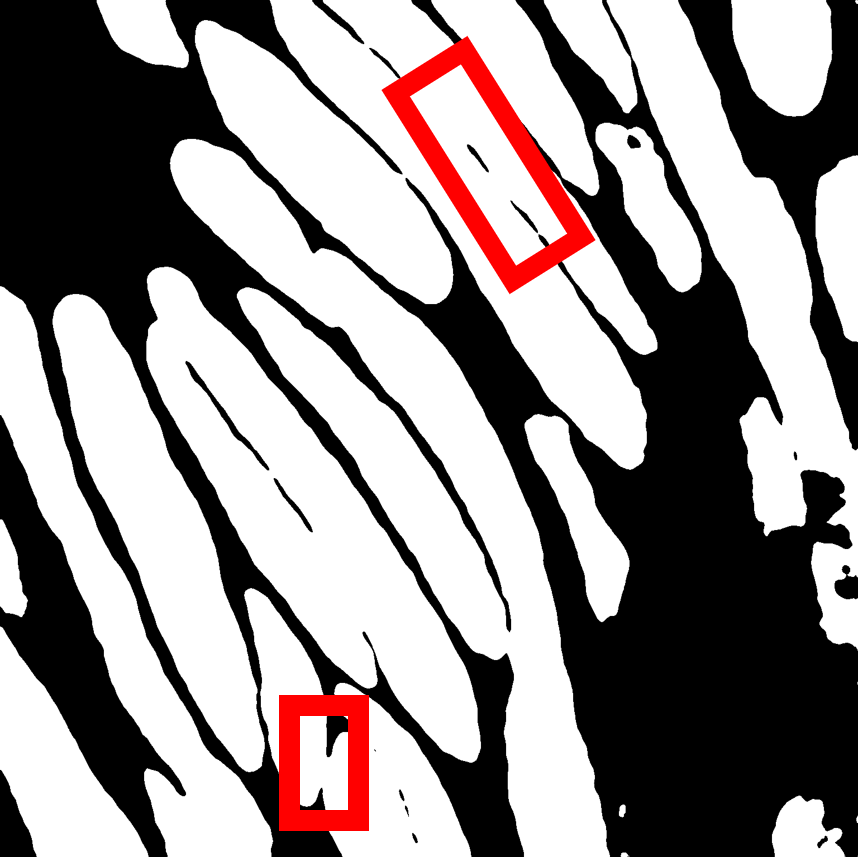}  &
    \includegraphics[width=\linewidth]{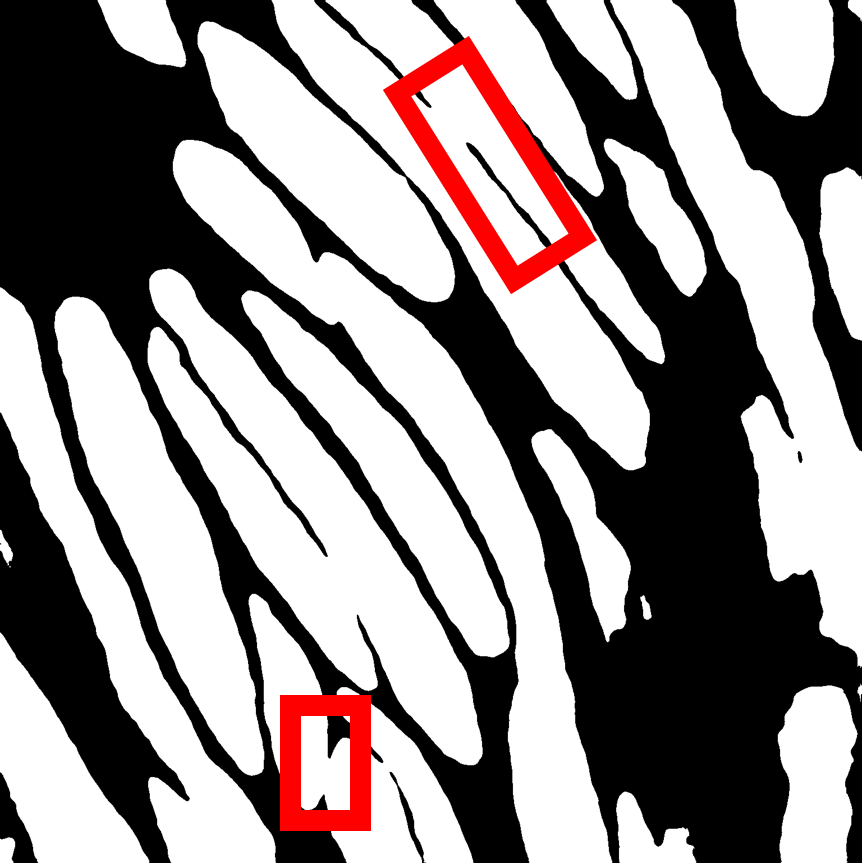} &
    \includegraphics[width=\linewidth]{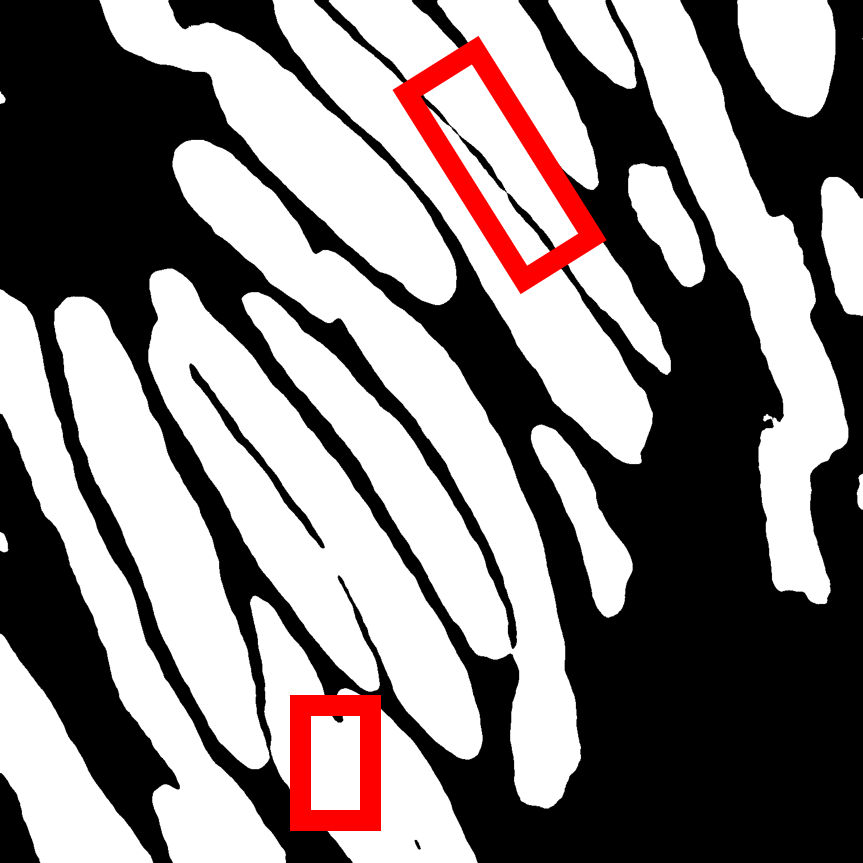} &
    \includegraphics[width=\linewidth]{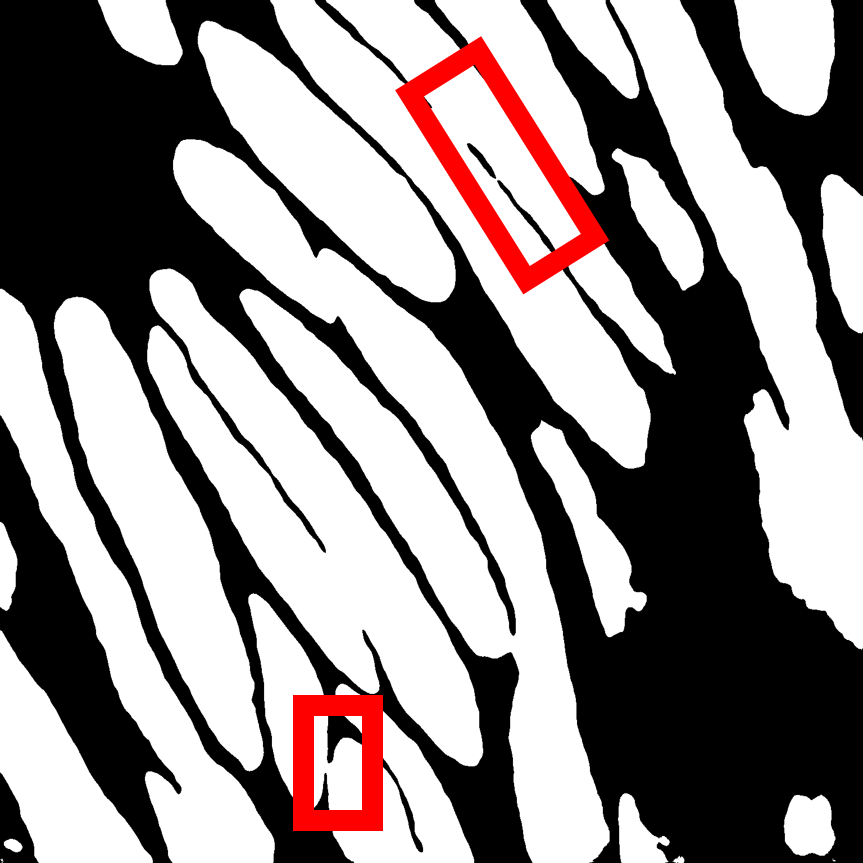}  &
    \includegraphics[width=\linewidth]{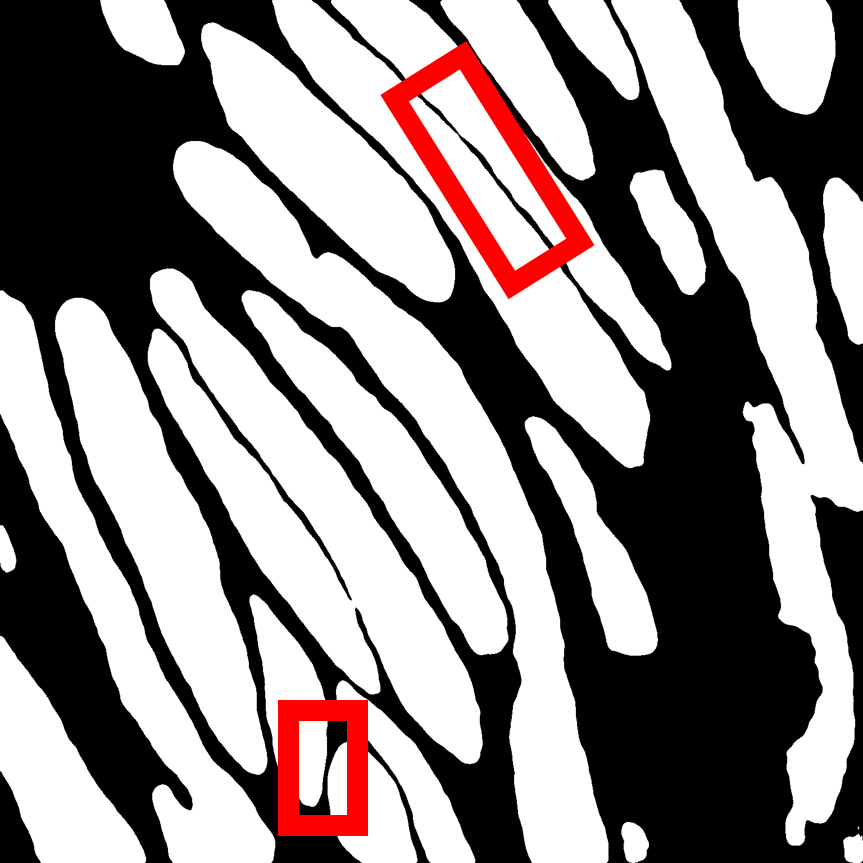} \\[10pt]
    % ------------ third row ------------
    \includegraphics[width=\linewidth]{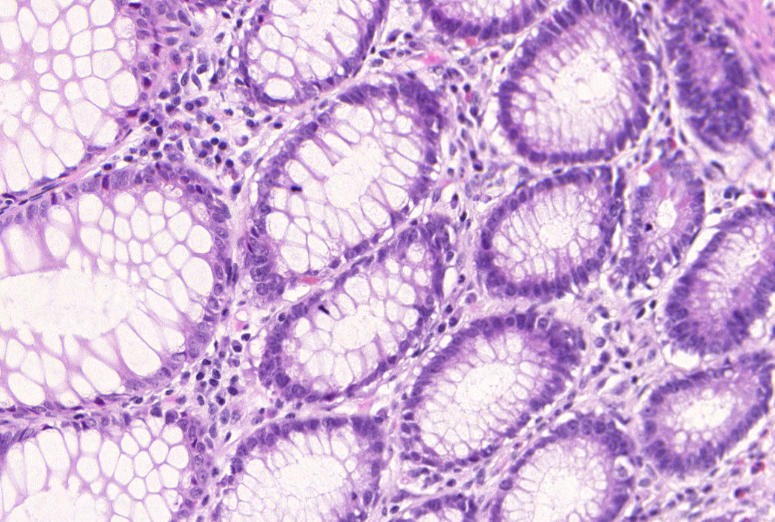}  &
    \includegraphics[width=\linewidth]{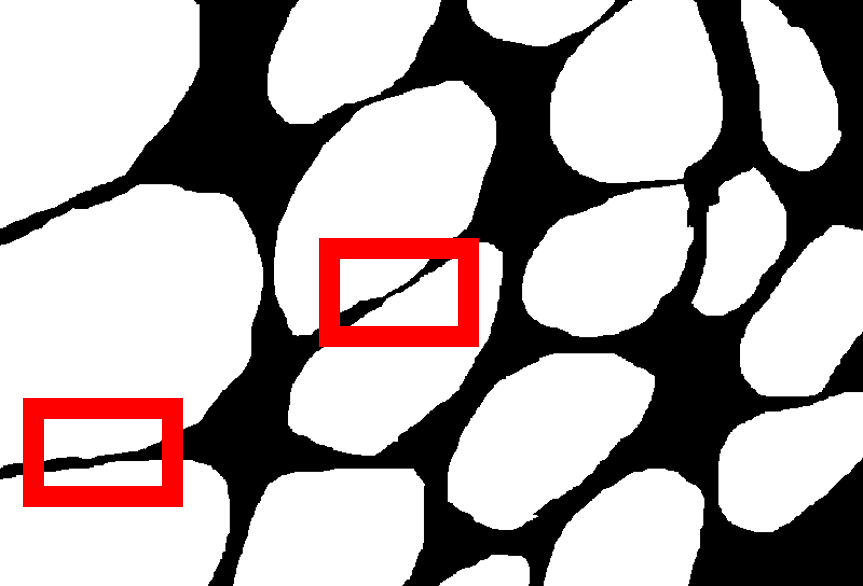}   &
    \includegraphics[width=\linewidth]{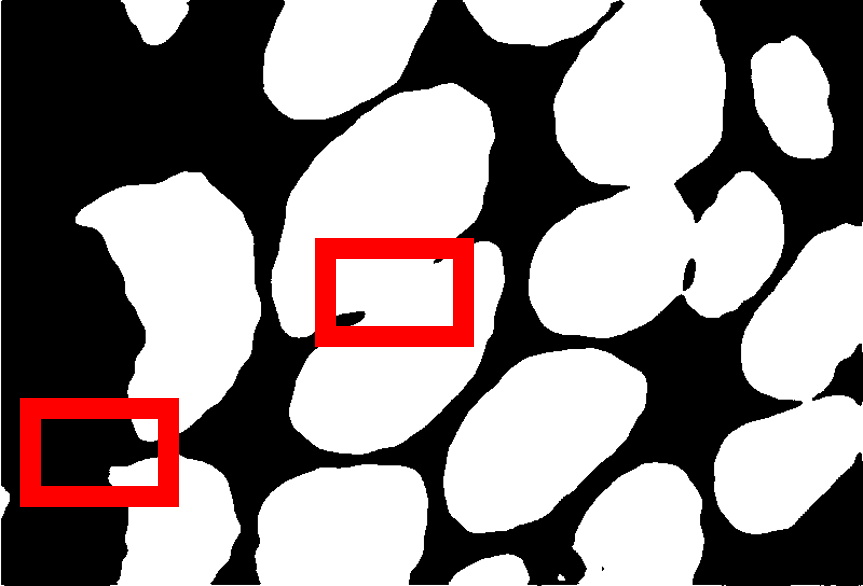}   &
    \includegraphics[width=\linewidth]{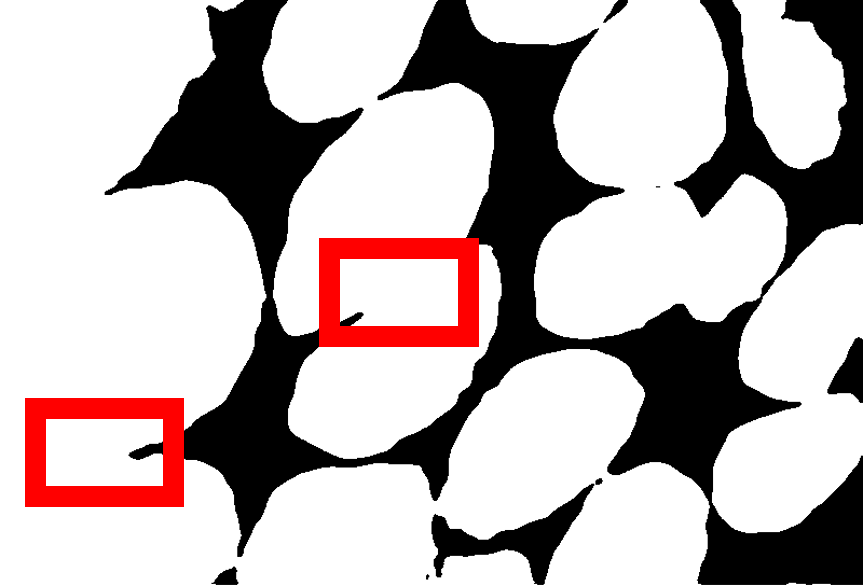}   &
    \includegraphics[width=\linewidth]{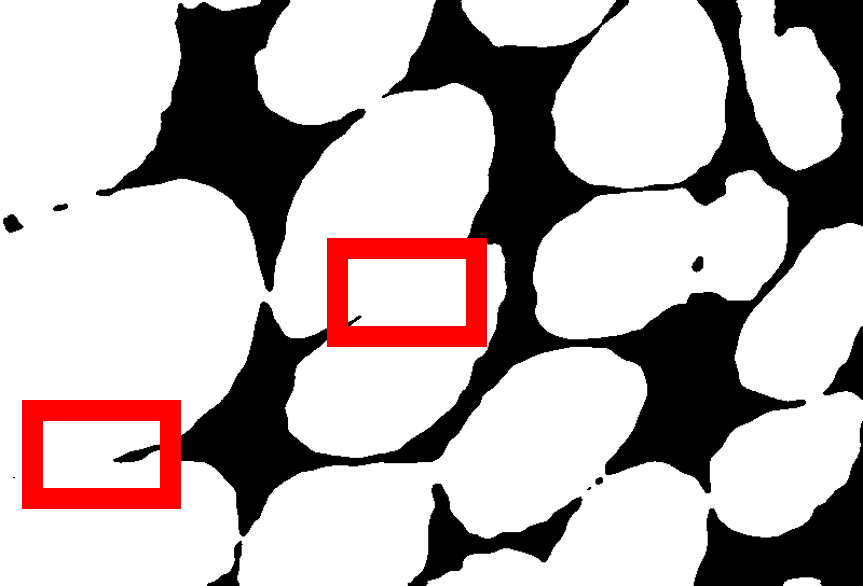} &
    \includegraphics[width=\linewidth]{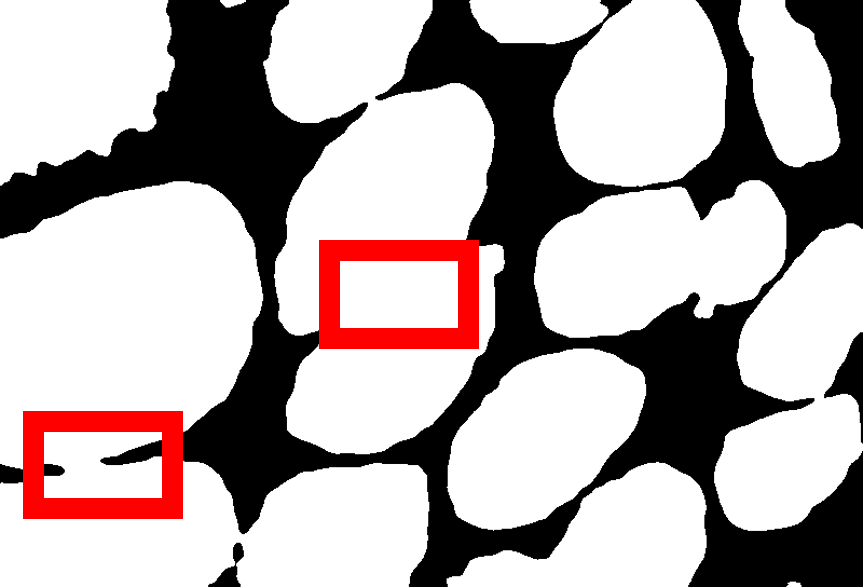}  &
    \includegraphics[width=\linewidth]{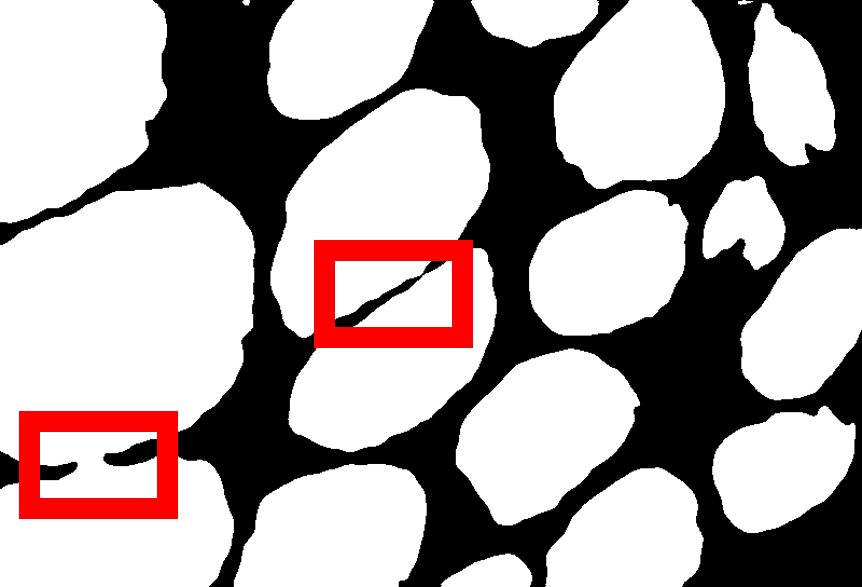} &
    \includegraphics[width=\linewidth]{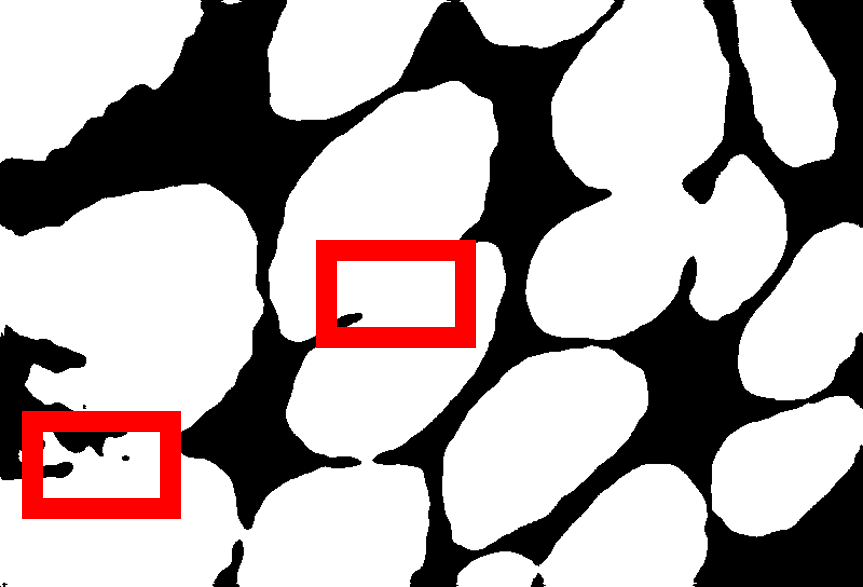} &
    \includegraphics[width=\linewidth]{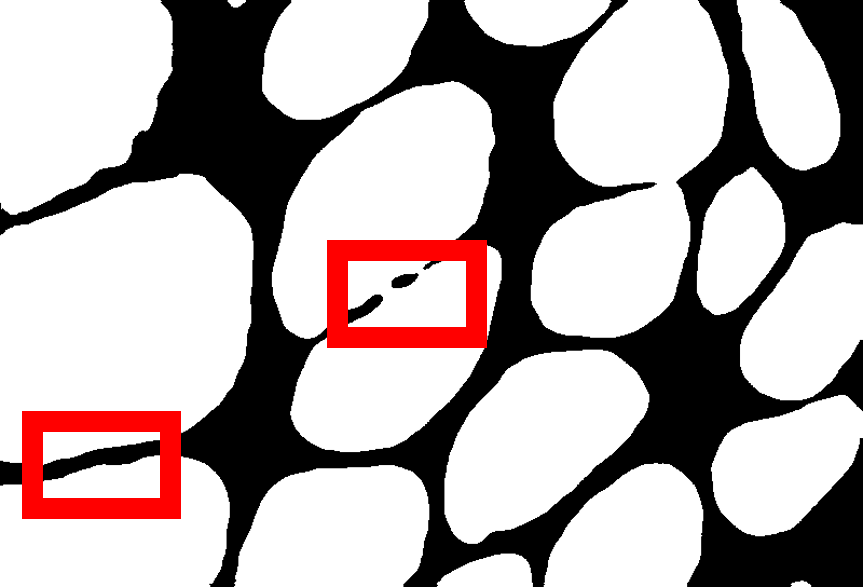}  &
    \includegraphics[width=\linewidth]{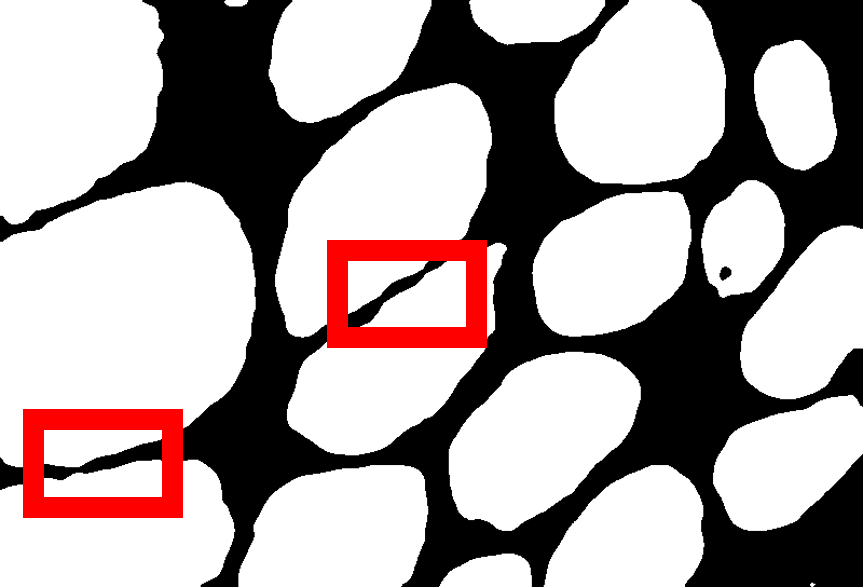} \\[10pt]
    % ------------ fourth row ----------
    \includegraphics[width=\linewidth]{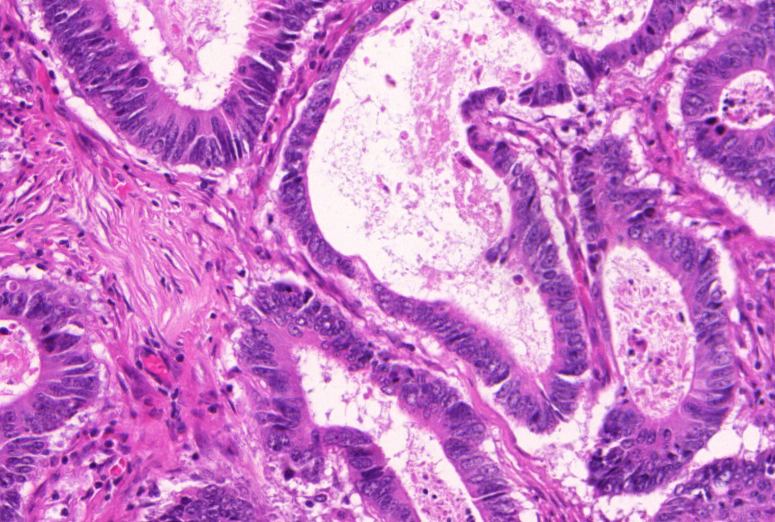}  &
    \includegraphics[width=\linewidth]{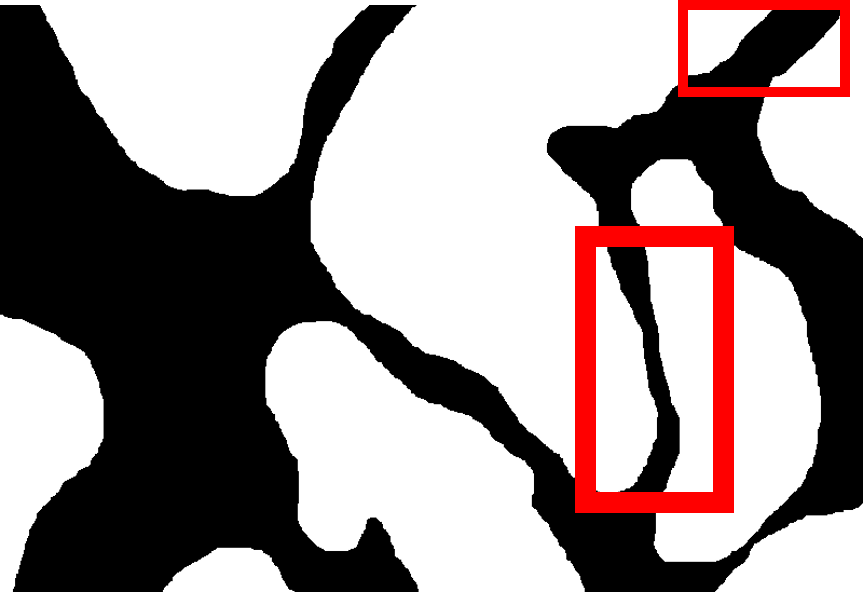}   &
    \includegraphics[width=\linewidth]{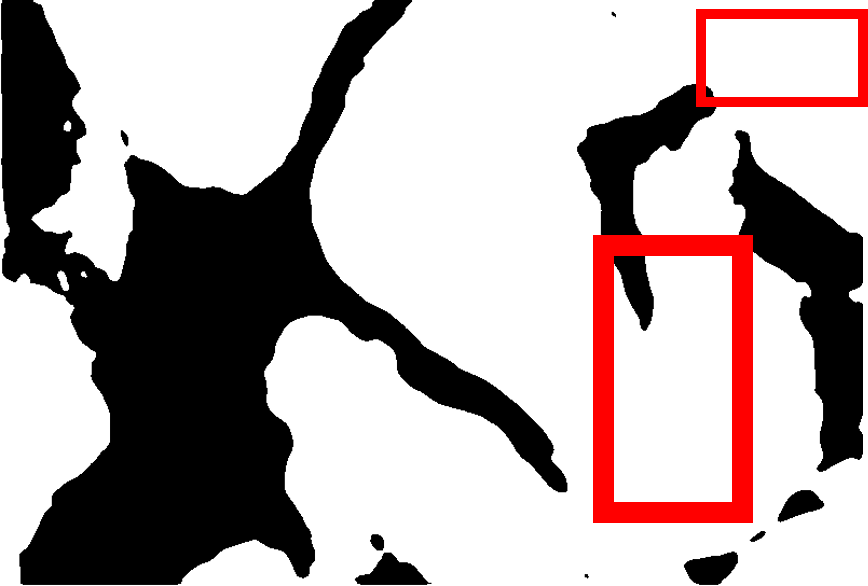}   &
    \includegraphics[width=\linewidth]{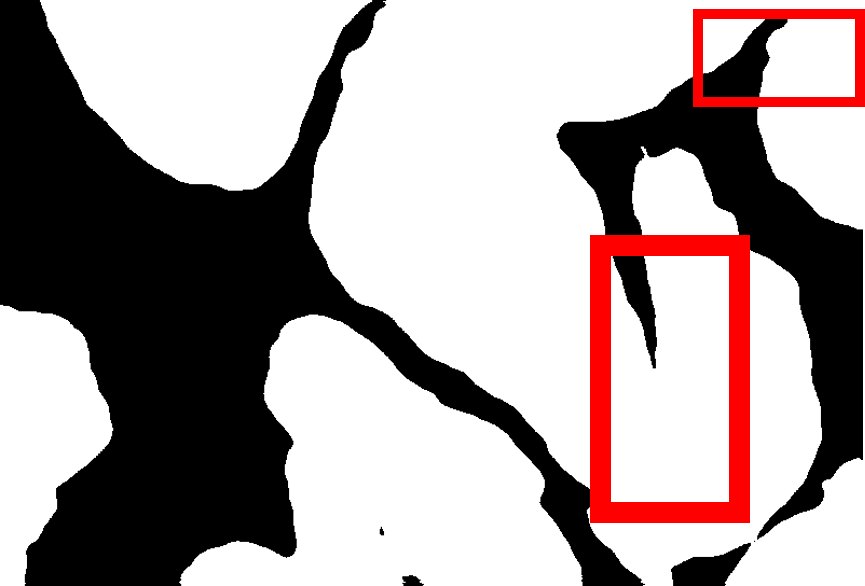}   &
    \includegraphics[width=\linewidth]{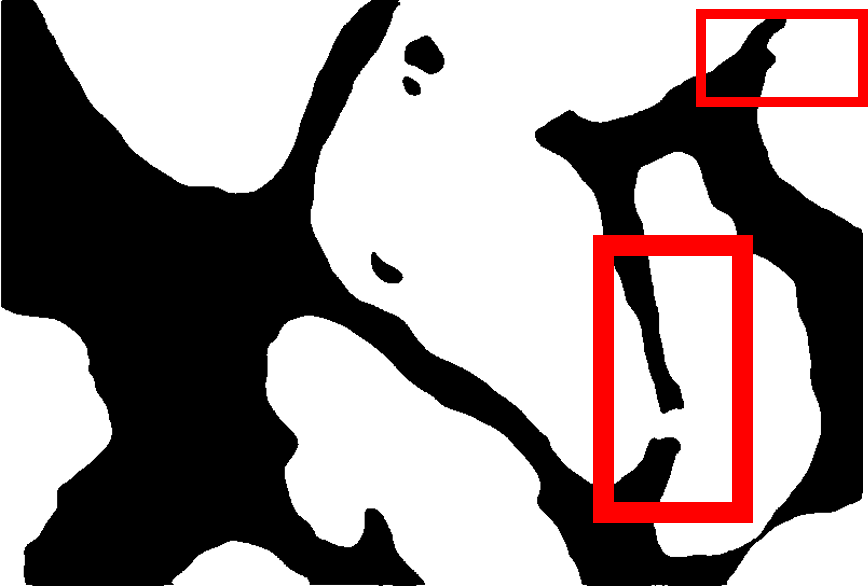} &
    \includegraphics[width=\linewidth]{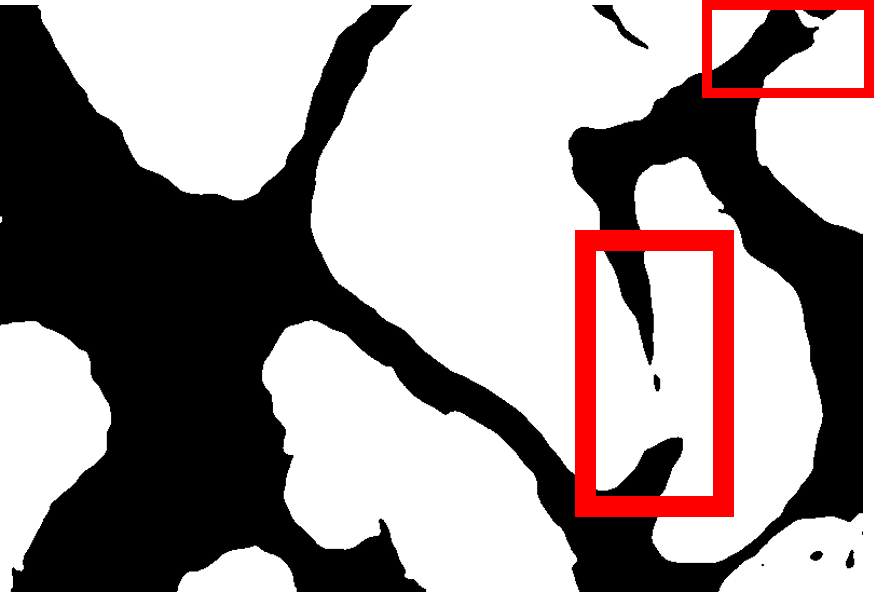}  &
    \includegraphics[width=\linewidth]{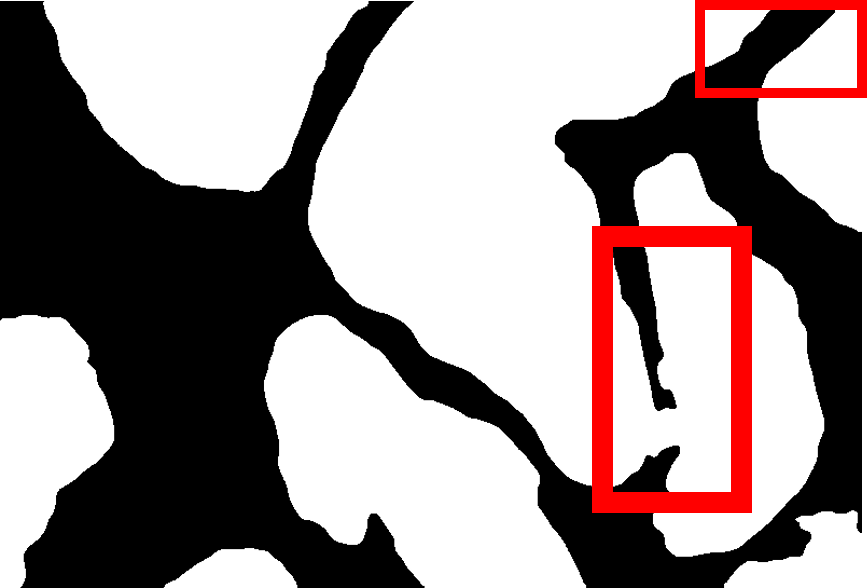} &
    \includegraphics[width=\linewidth]{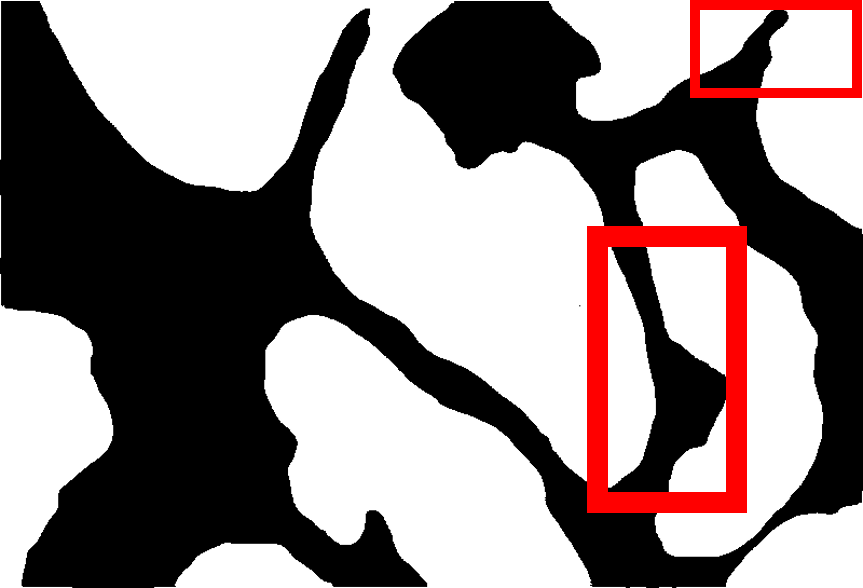} &
    \includegraphics[width=\linewidth]{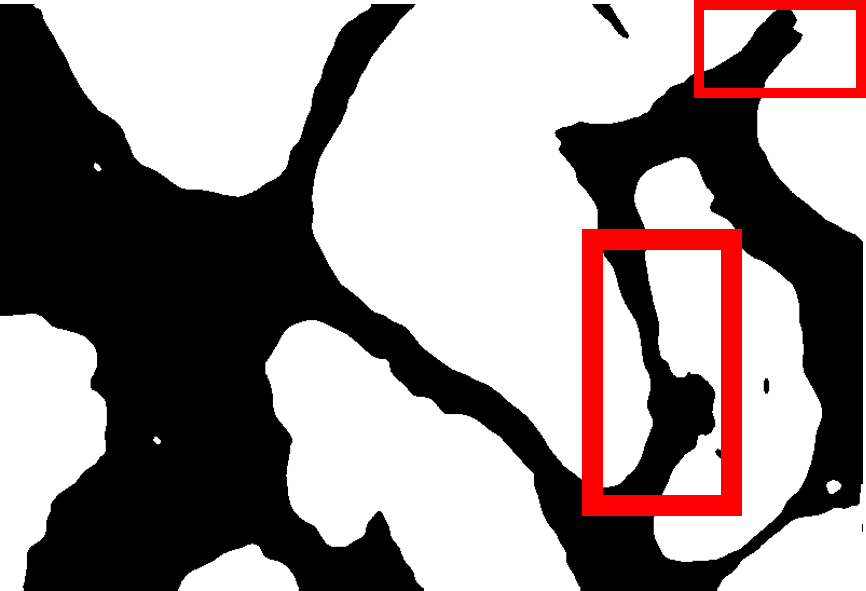}  &
    \includegraphics[width=\linewidth]{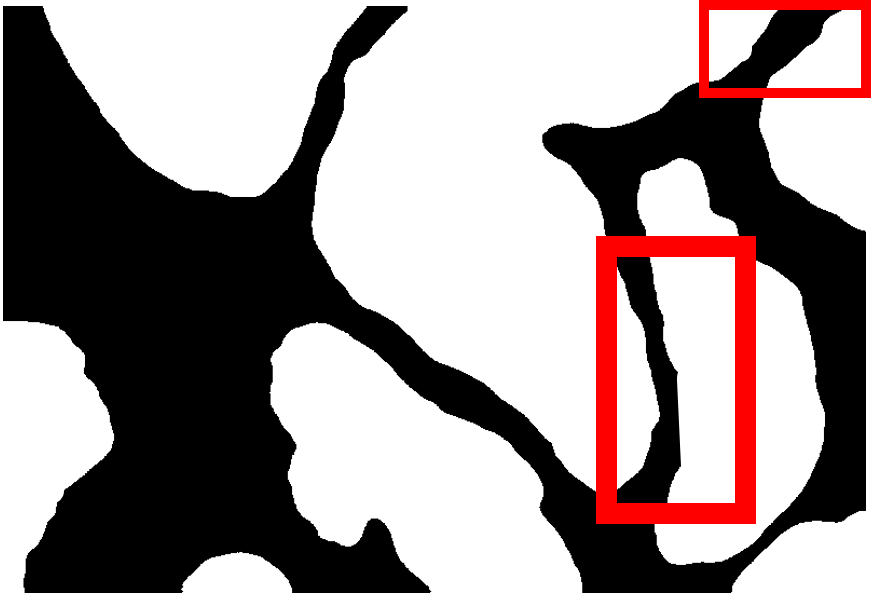} \\[16pt]
    % ---------- caption row -----------
    \scriptsize (a) Image &
    \scriptsize (b) GT &
    \scriptsize (c) MT &
    \scriptsize (d) EM &
    \scriptsize (e) UA-MT &
    \scriptsize (f) URPC &
    \scriptsize (g) XNet &
    \scriptsize (h) PMT &
    \scriptsize (i) TopoSemi &
    \scriptsize (j) Ours \\
  \end{tabular}
  \caption{Qualitative results for semi-supervised methods on $10\%$ and $20\%$ labeled data. Rows 1-2 correspond to CRAG dataset, rows 3-4 correspond to the GlaS dataset.
  From left to right: (a) raw image, (b) ground-truth mask, (c) to (i) present the $7$ baselines. (j) indicates the results of our method. The regions prone to topological errors are highlighted in \textcolor{red}{red} boxes.}
  \label{fig:qualitative_results}
  %\vspace{-.2in}
\end{figure*}

\myparagraph{Qualitative Results}.
We provide the qualitative results in~\Cref{fig:qualitative_results}. The qualitative comparison highlights that our proposed method consistently outperforms other semi-supervised methods in preserving accurate glandular structures and topology across various histopathology samples. The comparative methods exhibit notable topological errors, including fragmentation, merging, and boundary leakage, as indicated by the \textcolor{red}{red} boxes. In contrast, our method effectively mitigates these errors, demonstrating superior robustness in maintaining topological integrity and accurate boundary delineation, thereby underscoring its effectiveness for precise medical image analysis tasks.

\subsection{Ablation Study}
\label{subsec:ablation_study}
To comprehensively explore the robustness and efficacy of our proposed strategy, hyperparameter-selection, and experimental settings, we conduct the ablation experiments on the CRAG dataset using 20\% labeled data.

\myparagraph{Ablation Study on Matching Algorithm.}
To validate the effectiveness of our proposed matching algorithm, we compare it against established alternatives, including Wasserstein Matching~\cite{hu2019topology} and Betti-Matching~\cite{stucki2023topologically}. As shown in~\Cref{tab:matching_comparison}, our algorithm consistently achieves superior performance in both pixel- and topology-wise metrics. Specifically, Wasserstein Matching, relying exclusively on persistence values without spatial information, exhibits the worst results. Although Betti-Matching incorporates spatial context, it still performs suboptimally compared to our method.

\myparagraph{Ablation Study on IoU and Spatial Proximity (SP).}
To validate the effectiveness of the individual items of our matching cost, we conduct an ablation study on IoU and spatial proximity. The results in~\Cref{tab:matching_components} quantitatively substantiate the complementary roles of the IoU and the spatial proximity factor in our Hungarian assignment cost. Removing either the proximity or the overlap item could degrade the performance. The overlap itself cannot fully distinguish spatially adjacent structures. These results demonstrate that both items are necessary to achieve topologically accurate matching.

% Keep these two tables in one row; requires \usepackage{caption} for \captionof
\begin{table*}[t]
  \centering
  \tiny

  % -------- Left: Ablation study on matching algorithms --------
  \begin{minipage}[t]{.49\linewidth}
    \centering
    \captionof{table}{Ablation study on matching algorithms.}
    \label{tab:matching_comparison}
    \setlength{\tabcolsep}{1.8pt}
    \begin{tabular}{lccccc}
      \toprule
      \multirow{2}{*}{Matching} & Pixel-wise &  & \multicolumn{3}{c}{Topology-wise} \\ \cmidrule(lr){2-2}\cmidrule(lr){4-6}
       & Dice\_obj $\uparrow$ &  & BE $\downarrow$ & BME $\downarrow$ & DIU $\downarrow$ \\ \midrule
      Wasser.~\cite{hu2019topology}   & 0.864 $\pm$ 0.007 &  & 0.423 $\pm$ 0.026 & 9.647 $\pm$ 0.846 & 58.592 $\pm$ 2.574 \\
      Betti~\cite{stucki2023topologically} & 0.889 $\pm$ 0.005 &  & 0.237 $\pm$ 0.021 & 8.216 $\pm$ 0.717 & 44.157 $\pm$ 2.146 \\
      Ours                                & \textbf{0.909 $\pm$ 0.005} &  & \textbf{0.188 $\pm$ 0.018} & \textbf{7.425 $\pm$ 0.570} & \textbf{40.250 $\pm$ 1.720} \\ \bottomrule
    \end{tabular}
  \end{minipage}
  \hfill
  % -------- Right: Effect of IoU & spatial-proximity (SP) --------
  \begin{minipage}[t]{.49\linewidth}
    \centering
    \captionof{table}{Effect of IoU \& spatial-proximity (SP).}
    \label{tab:matching_components}
    \setlength{\tabcolsep}{1.8pt}
    \begin{tabular}{lcclccc}
      \toprule
      \multirow{2}{*}{IoU} & \multirow{2}{*}{SP} & Pixel-wise &  & \multicolumn{3}{c}{Topology-wise} \\ \cmidrule(lr){3-3}\cmidrule(lr){5-7}
       &  & Dice\_obj $\uparrow$ &  & BE $\downarrow$ & BME $\downarrow$ & DIU $\downarrow$ \\ \midrule
      \cmark & \xmark & 0.890 $\pm$ 0.005 &  & 0.233 $\pm$ 0.020 & 8.300 $\pm$ 0.650 & 43.750 $\pm$ 2.100 \\
      \xmark & \cmark & 0.882 $\pm$ 0.006 &  & 0.247 $\pm$ 0.022 & 9.600 $\pm$ 0.680 & 46.200 $\pm$ 2.250 \\
      \cmark & \cmark & \textbf{0.909 $\pm$ 0.005} &  & \textbf{0.188 $\pm$ 0.018} & \textbf{7.425 $\pm$ 0.570} & \textbf{40.250 $\pm$ 1.720} \\ \bottomrule
    \end{tabular}
  \end{minipage}

\end{table*}

\myparagraph{Sensitivity Analysis on $B_{\text{intra}}$ and $B_{\text{temp}}$.}
We further analyzed the sensitivity of our method to the number of MC dropout samples $B_{\text{intra}}$ and temporal training snapshots $B_{\text{temp}}$.~\Cref{tab:num_passes} shows that employing too few facets yields unreliable estimation of topological consistency, resulting in suboptimal segmentation performance. Conversely, increasing the number of facets beyond an optimal point introduces redundant information and additional variability, degrading model performance. Therefore, $4$ is the optimal number that strikes a practical balance, ensuring the best performance while remaining computationally efficient.

% Keep these two tables in one row; requires \usepackage{caption} for \captionof
\begin{table*}[ht]
  \centering
  \tiny

  % ---------- Left: Influence of B_intra and B_temp ----------
  \begin{minipage}[t]{0.49\linewidth}
    \centering
    \captionof{table}{Influence of $B_{\text{intra}}$ and $B_{\text{temp}}$.}
    \label{tab:num_passes}
    \setlength{\tabcolsep}{1pt}
    \begin{tabular}{lcccccc}
      \toprule
      \multirow{2}{*}{$B_{\text{intra}}$} &
      \multirow{2}{*}{$B_{\text{temp}}$} &
      \multicolumn{1}{c}{Pixel-wise} & &
      \multicolumn{3}{c}{Topology-wise} \\ \cmidrule(lr){3-3}\cmidrule(lr){5-7}
       & & Dice\_obj $\uparrow$ && BE $\downarrow$ & BME $\downarrow$ & DIU $\downarrow$ \\ \midrule
      2 & 2 & 0.878 $\pm$ 0.010 &  & 0.255 $\pm$ 0.025 & 9.350 $\pm$ 0.620 & 48.600 $\pm$ 2.300 \\
      3 & 3 & 0.892 $\pm$ 0.007 &  & 0.214 $\pm$ 0.020 & 8.105 $\pm$ 0.600 & 44.105 $\pm$ 2.050 \\ 
      5 & 5 & 0.872 $\pm$ 0.011 &  & 0.275 $\pm$ 0.023 & 10.050 $\pm$ 0.630 & 54.250 $\pm$ 2.253 \\ 
      4 & 4 & \textbf{0.909 $\pm$ 0.005} &  & \textbf{0.188 $\pm$ 0.018} & \textbf{7.425 $\pm$ 0.570} & \textbf{40.250 $\pm$ 1.720} \\ \bottomrule
    \end{tabular}
  \end{minipage}
  \hfill
  % ---------- Right: Efficacy of L_intra and L_temp ----------
  \begin{minipage}[t]{0.49\linewidth}
    \centering
    \captionof{table}{Efficacy of $\mathcal{L}_{\text{intra}}$ and $\mathcal{L}_{\text{temp}}$.}
    \label{tab:consistency_losses}
    \setlength{\tabcolsep}{1pt}
    \begin{tabular}{lcccccc}
      \toprule
      \multirow{2}{*}{$\mathcal{L}_{\text{intra}}$} &
      \multirow{2}{*}{$\mathcal{L}_{\text{temp}}$} &
      \multicolumn{1}{c}{Pixel-wise} & &
      \multicolumn{3}{c}{Topology-wise} \\ \cmidrule(lr){3-3}\cmidrule(lr){5-7}
       &  & Dice\_obj $\uparrow$ && BE $\downarrow$ & BME $\downarrow$ & DIU $\downarrow$ \\ \midrule
      \xmark & \xmark & 0.862 $\pm$ 0.011 &  & 0.460 $\pm$ 0.022 & 11.680 $\pm$ 0.610 & 59.930 $\pm$ 2.150 \\ 
      \cmark & \xmark & 0.898 $\pm$ 0.006 &  & 0.215 $\pm$ 0.020 & 7.920 $\pm$ 0.590 & 44.750 $\pm$ 1.970 \\ 
      \xmark & \cmark & 0.882 $\pm$ 0.008 &  & 0.238 $\pm$ 0.031 & 8.540 $\pm$ 0.450 & 45.310 $\pm$ 2.040 \\
      \cmark & \cmark & \textbf{0.909 $\pm$ 0.005} &  & \textbf{0.188 $\pm$ 0.018} & \textbf{7.425 $\pm$ 0.570} & \textbf{40.250 $\pm$ 1.720} \\ \bottomrule
    \end{tabular}
  \end{minipage}

\end{table*}

\myparagraph{Ablation Study on Loss Components.}
To evaluate the contributions of individual loss terms in our dual-level topological consistency framework, we conduct experiments selectively enabling or disabling the $\mathcal{L}_{\text{intra}}$ and $\mathcal{L}_{\text{temp}}$. As presented in~\Cref{tab:consistency_losses}, each loss individually improves the pixel- and topology-wise performance compared to the baseline without these constraints. Combining both losses achieves the strongest overall performance, confirming that $\mathcal{L}_{\text{intra}}$ and $\mathcal{L}_{\text{temp}}$ complement each other by addressing different sources of topological inaccuracies—stochastic noise within single facets and structural inconsistencies across training iterations.

\myparagraph{Ablation Study on 1-D Topological Features.}
We mainly focus on 0-D topological features due to the following factors: For the primary application in our study (gland and nuclei segmentation), the most critical topological errors involve incorrect splitting or merging of individual structures, which are well-captured by 0-D persistent homology. 
For the validation on 1-dimensional structures, we conducted additional experiments on the Roads dataset~\cite{mnih2013machine}. The results are shown in~\Cref{tab:road_dataset}. It verifies that our method could learn good topological representations from unlabeled data on 1-dimensional topological features.

\begin{table*}[t]
\centering
\tiny
\setlength{\tabcolsep}{7pt}

\begin{minipage}[t]{.47\textwidth}
\centering
\captionof{table}{The results on the Roads dataset.}
\label{tab:road_dataset}
\setlength{\tabcolsep}{1.3pt}
\begin{tabular}{llccc}
\toprule
Labeled Ratio & Method & BE\,$\downarrow$ & BME\,$\downarrow$ & DIU\,$\downarrow$ \\
\midrule
\multirow{2}{*}{10\%} & TopoSemiSeg & 8.324 $\pm$ 0.729 & 9.681 $\pm$ 0.647 & 10.952 $\pm$ 0.671 \\
& Ours & \textbf{7.892 $\pm$ 0.634} & \textbf{8.147 $\pm$ 0.521} & \textbf{9.376 $\pm$ 0.583} \\
\midrule
\multirow{2}{*}{20\%} & TopoSemiSeg & 7.467 $\pm$ 0.582 & 8.213 $\pm$ 0.514 & 9.387 $\pm$ 0.538 \\
& Ours & \textbf{6.983 $\pm$ 0.507} & \textbf{7.024 $\pm$ 0.436} & \textbf{8.149 $\pm$ 0.492} \\
\bottomrule
\end{tabular}
\end{minipage}
\hfill
\begin{minipage}[t]{.51\textwidth}
\centering
\captionof{table}{Density-aware quantitative results.}
\label{tab:crowding_ablation}
\setlength{\tabcolsep}{1.8pt}
\renewcommand{\arraystretch}{1.12}
\begin{tabular}{lccc}
\toprule
Setting & Dice\_obj\,$\uparrow$ & BE\,$\downarrow$ & BME\,$\downarrow$ \\
\midrule
Sparse (Ours, $\leq 30$ cells)              & 0.804 $\pm$ 0.004 & 4.620 $\pm$ 0.140 & 163.132 $\pm$ 2.136 \\
Crowded (\cite{xu2024semi}, $\geq 100$ cells)     & 0.756 $\pm$ 0.009 & 6.890 $\pm$ 0.240 & 198.525 $\pm$ 3.125 \\
Crowded (Ours, $\geq 100$ cells)            & 0.774 $\pm$ 0.007 & 5.610 $\pm$ 0.198 & 186.313 $\pm$ 2.715 \\
Ours (whole test image)                     & 0.790 $\pm$ 0.006 & 4.930 $\pm$ 0.156 & 179.225 $\pm$ 2.383 \\
\bottomrule
\end{tabular}
\end{minipage}

\end{table*}

\myparagraph{Crowding-Aware Ablation Study.}
To quantify the influence of nuclei density on model performance, we randomly cut the test images into patches of size $256\times256$. For every patch, we count nuclei in the ground-truth instance map. Patches with <= 30 nuclei are labeled Sparse; those with >=100 nuclei are labeled Crowded. We sampled $14$ samples for a fair comparison and show the results below in~\Cref{tab:crowding_ablation}. The experiments above verify that our approach is density-aware. It achieves state-of-the-art performance on typical tissue, excels in sparse fields, and maintains a clear advantage over the strongest baseline under extreme nuclear crowding.

\myparagraph{Ablation Study on the Alternatives of MC-dropout.}
We choose two alternative perturbation methods: Variational Inference (VI)~\cite{jordan1999introduction}, which generates multiple predictions by sampling from the learned variational posterior distribution, and Temperature Scaling~\cite{guo2017calibration}, which produces diverse predictions through multiple sampling from temperature-modulated probability distributions. The experiments are conducted on CRAG 20\% labeled data, and the results are shown in~\Cref{tab:diff_perturbation}.

\begin{table*}[ht]
\centering
\tiny
\setlength{\tabcolsep}{7pt}

\begin{minipage}[t]{.49\textwidth}
\centering
\captionof{table}{Ablation of perturbation methods.}
\label{tab:diff_perturbation}
\setlength{\tabcolsep}{5pt}
\begin{tabular}{lccc}
\toprule
Method & Dice\_obj\,$\uparrow$ & BE\,$\downarrow$ & BME\,$\downarrow$ \\
\midrule
Variational Inference & 0.895 $\pm$ 0.006 & 0.242 $\pm$ 0.022 & 9.125 $\pm$ 0.685 \\
Temperature Scaling   & 0.891 $\pm$ 0.007 & 0.258 $\pm$ 0.025 & 9.850 $\pm$ 0.795 \\
MC-Dropout            & \textbf{0.909 $\pm$ 0.005} & \textbf{0.188 $\pm$ 0.018} & \textbf{7.425 $\pm$ 0.570} \\
\bottomrule
\end{tabular}
\end{minipage}
\hfill
\begin{minipage}[t]{.49\textwidth}
\centering
\captionof{table}{Comparison with foundation models.}
\label{tab:ssl_comparison}
\setlength{\tabcolsep}{5pt}
\begin{tabular}{lccc}
\toprule
Method & Dice\_obj\,$\uparrow$ & BE\,$\downarrow$ & BME\,$\downarrow$ \\
\midrule
LoRA\textendash SAM     & 0.882 $\pm$ 0.006 & 0.440 $\pm$ 0.042 & 27.300 $\pm$ 2.937 \\
LoRA\textendash MedSAM  & 0.898 $\pm$ 0.005 & 0.268 $\pm$ 0.025 & 11.275 $\pm$ 1.899 \\
Ours                    & \textbf{0.909 $\pm$ 0.005} & \textbf{0.188 $\pm$ 0.018} & \textbf{7.425 $\pm$ 0.570} \\
\bottomrule
\end{tabular}
\end{minipage}

\end{table*}

\myparagraph{Comparison with Self-Supervised Methods Finetuned on Limited Labeled Data.}
To comprehensively evaluate the effectiveness of our method, we compare our method against some foundation models, like SAM~\cite{kirillov2023segment} and MedSAM~\cite{ma2024segment}. We use LoRA~\cite{hu2022lora} to finetune these two models using $20\%$ labeled data on the CRAG dataset and report the performance in~\Cref{tab:ssl_comparison}. The results show that even with powerful foundation models, like SAM or MedSAM, topological errors can still exist without explicit topological modeling.

\section{Conclusion}
\label{conclusion}
We present a semi-supervised segmentation framework that preserves significant topological structures in histopathology with limited annotations. Dual-level topological consistency across Monte Carlo dropout predictions and temporal training snapshots separates stable biological patterns from noise. For alignment, MATCH-Pair achieves spatially accurate matching between noisy persistence diagrams by combining spatial overlap, persistence, and proximity, and MATCH-Global scales to multiple facets. Experiments show consistent gains in robustness and substantial reductions in topological errors, enabling more reliable downstream analyses in digital pathology.

\myparagraph{Acknowledgement.} 
This work was partially supported by grants NSF CCF-2144901, NIH R01NS143143, and NIH R01CA297843.

{
\small
\bibliographystyle{plain}
\bibliography{arxiv_version}

@String(CVPR= {IEEE Conf. Comput. Vis. Pattern Recog.})

@String(ICCV= {Int. Conf. Comput. Vis.})

@String(ECCV= {Eur. Conf. Comput. Vis.})

@String(BMVC= {Brit. Mach. Vis. Conf.})

@String(ICLR = {Int. Conf. Learn. Represent.})

@String(AAAI = {AAAI})

@String(CVPR  = {CVPR})

@String(ICCV  = {ICCV})

@String(ECCV  = {ECCV})

@String(BMVC  =	{BMVC})

@String(ICLR  = {ICLR})

@inproceedings{xu2024semi,
  title={Semi-supervised segmentation of histopathology images with noise-aware topological consistency},
  author={Xu, Meilong and Hu, Xiaoling and Gupta, Saumya and Abousamra, Shahira and Chen, Chao},
  booktitle={ECCV},
  year={2024},
}

@inproceedings{hu2019topology,
  title={Topology-preserving deep image segmentation},
  author={Hu, Xiaoling and Li, Fuxin and Samaras, Dimitris and Chen, Chao},
  booktitle={NeurIPS},
  year={2019}
}

@inproceedings{gupta2023topology,
  title={Topology-aware uncertainty for image segmentation},
  author={Gupta, Saumya and Zhang, Yikai and Hu, Xiaoling and Prasanna, Prateek and Chen, Chao},
  booktitle={NeurIPS},
  year={2023}
}

@article{fleming2012colorectal,
  title={Colorectal carcinoma: Pathologic aspects},
  author={Fleming, Matthew and Ravula, Sreelakshmi and Tatishchev, Sergei F and Wang, Hanlin L},
  journal={Journal of gastrointestinal oncology},
  year={2012}
}

@article{montironi2005gleason,
  title={Gleason grading of prostate cancer in needle biopsies or radical prostatectomy specimens: contemporary approach, current clinical significance and sources of pathology discrepancies},
  author={Montironi, Rodolfo and Mazzuccheli, Roberta and Scarpelli, Marina and Lopez-Beltran, Antonio and Fellegara, Giovanni and Algaba, Ferran},
  journal={BJU international},
  year={2005},
  publisher={Wiley Online Library}
}

@article{karasaki2023evolutionary,
  title={Evolutionary characterization of lung adenocarcinoma morphology in TRACERx},
  author={Karasaki, Takahiro and Moore, David A and Veeriah, Selvaraju and Naceur-Lombardelli, Cristina and Toncheva, Antonia and Magno, Neil and Ward, Sophia and Bakir, Maise Al and Watkins, Thomas BK and Grigoriadis, Kristiana and others},
  journal={Nature medicine},
  year={2023},
  publisher={Nature Publishing Group US New York}
}

@article{sirinukunwattana2017gland,
  title={Gland segmentation in colon histology images: The glas challenge contest},
  author={Sirinukunwattana, Korsuk and Pluim, Josien PW and Chen, Hao and Qi, Xiaojuan and Heng, Pheng-Ann and Guo, Yun Bo and Wang, Li Yang and Matuszewski, Bogdan J and Bruni, Elia and Sanchez, Urko and others},
  journal={MedIA},
  year={2017},
}

@inproceedings{ronneberger2015u,
  title={U-net: Convolutional networks for biomedical image segmentation},
  author={Ronneberger, Olaf and Fischer, Philipp and Brox, Thomas},
  booktitle={MICCAI},
  year={2015},
}

@inproceedings{zhou2018unet++,
  title={Unet++: A nested u-net architecture for medical image segmentation},
  author={Zhou, Zongwei and Rahman Siddiquee, Md Mahfuzur and Tajbakhsh, Nima and Liang, Jianming},
  booktitle={Deep learning in medical image analysis and multimodal learning for clinical decision support: 4th international workshop, DLMIA 2018, and 8th international workshop, ML-CDS 2018, held in conjunction with MICCAI 2018},
  year={2018},
}

@article{kumar2019multi,
  title={A multi-organ nucleus segmentation challenge},
  author={Kumar, Neeraj and Verma, Ruchika and Anand, Deepak and Zhou, Yanning and Onder, Omer Fahri and Tsougenis, Efstratios and Chen, Hao and Heng, Pheng-Ann and Li, Jiahui and Hu, Zhiqiang and others},
  journal={TMI},
  year={2019},
}

@inproceedings{grandvalet2004semi,
  title={Semi-supervised learning by entropy minimization},
  author={Grandvalet, Yves and Bengio, Yoshua},
  booktitle={NeurIPS},
  year={2004}
}

@article{isensee2021nnu,
  title={nnU-Net: a self-configuring method for deep learning-based biomedical image segmentation},
  author={Isensee, Fabian and Jaeger, Paul F and Kohl, Simon AA and Petersen, Jens and Maier-Hein, Klaus H},
  journal={Nature methods},
  year={2021},
  publisher={Nature Publishing Group}
}

@inproceedings{lux2024topograph,
  title={Topograph: An efficient Graph-Based Framework for Strictly Topology Preserving Image Segmentation},
  author={Lux, Laurin and Berger, Alexander H and Weers, Alexander and Stucki, Nico and Rueckert, Daniel and Bauer, Ulrich and Paetzold, Johannes C},
  booktitle={ICLR},
  year={2025}
}

@article{ma2024segment,
  title={Segment anything in medical images},
  author={Ma, Jun and He, Yuting and Li, Feifei and Han, Lin and You, Chenyu and Wang, Bo},
  journal={Nature Communications},
  year={2024},
  publisher={Nature Publishing Group UK London}
}

@article{graham2019hover,
  title={Hover-net: Simultaneous segmentation and classification of nuclei in multi-tissue histology images},
  author={Graham, Simon and Vu, Quoc Dang and Raza, Shan E Ahmed and Azam, Ayesha and Tsang, Yee Wah and Kwak, Jin Tae and Rajpoot, Nasir},
  journal={MedIA},
  year={2019},
}

@inproceedings{he2023toposeg,
  title={Toposeg: Topology-aware nuclear instance segmentation},
  author={He, Hongliang and Wang, Jun and Wei, Pengxu and Xu, Fan and Ji, Xiangyang and Liu, Chang and Chen, Jie},
  booktitle={ICCV},
  year={2023}
}

@article{horst2024cellvit,
  title={Cellvit: Vision transformers for precise cell segmentation and classification},
  author={H{\"o}rst, Fabian and Rempe, Moritz and Heine, Lukas and Seibold, Constantin and Keyl, Julius and Baldini, Giulia and Ugurel, Selma and Siveke, Jens and Gr{\"u}nwald, Barbara and Egger, Jan and others},
  journal={MedIA},
  year={2024},
}

@inproceedings{zhang2022boostmis,
  title={Boostmis: Boosting medical image semi-supervised learning with adaptive pseudo labeling and informative active annotation},
  author={Zhang, Wenqiao and Zhu, Lei and Hallinan, James and Zhang, Shengyu and Makmur, Andrew and Cai, Qingpeng and Ooi, Beng Chin},
  booktitle={CVPR},
  year={2022}
}

@article{zhang2022discriminative,
  title={Discriminative error prediction network for semi-supervised colon gland segmentation},
  author={Zhang, Zhenxi and Tian, Chunna and Bai, Harrison X and Jiao, Zhicheng and Tian, Xilan},
  journal={MedIA},
  year={2022},
}

@inproceedings{li2023calibrating,
  title={Calibrating Uncertainty for Semi-Supervised Crowd Counting},
  author={Li, Chen and Hu, Xiaoling and Abousamra, Shahira and Chen, Chao},
  booktitle={ICCV},
  year={2023}
}

@article{li2020transformation,
  title={Transformation-consistent self-ensembling model for semi-supervised medical image segmentation},
  author={Li, Xiaomeng and Yu, Lequan and Chen, Hao and Fu, Chi-Wing and Xing, Lei and Heng, Pheng-Ann},
  journal={TNNLS},
  year={2020},
}

@inproceedings{sohn2020fixmatch,
  title={Fixmatch: Simplifying semi-supervised learning with consistency and confidence},
  author={Sohn, Kihyuk and Berthelot, David and Carlini, Nicholas and Zhang, Zizhao and Zhang, Han and Raffel, Colin A and Cubuk, Ekin Dogus and Kurakin, Alexey and Li, Chun-Liang},
  booktitle={NeurIPS},
  year={2020}
}

@inproceedings{yu2019uncertainty,
  title={Uncertainty-aware self-ensembling model for semi-supervised 3D left atrium segmentation},
  author={Yu, Lequan and Wang, Shujun and Li, Xiaomeng and Fu, Chi-Wing and Heng, Pheng-Ann},
  booktitle={MICCAI},
  year={2019},
}

@inproceedings{konwer2025enhancing,
  title={Enhancing SAM with Efficient Prompting and Preference Optimization for Semi-supervised Medical Image Segmentation},
  author={Konwer, Aishik and Yang, Zhijian and Bas, Erhan and Xiao, Cao and Prasanna, Prateek and Bhatia, Parminder and Kass-Hout, Taha},
  booktitle={CVPR},
  year={2025}
}

@inproceedings{basak2023pseudo,
  title={Pseudo-Label Guided Contrastive Learning for Semi-Supervised Medical Image Segmentation},
  author={Basak, Hritam and Yin, Zhaozheng},
  booktitle={CVPR},
  year={2023}
}

@inproceedings{zhou2023xnet,
  title={XNet: Wavelet-Based Low and High Frequency Fusion Networks for Fully-and Semi-Supervised Semantic Segmentation of Biomedical Images},
  author={Zhou, Yanfeng and Huang, Jiaxing and Wang, Chenlong and Song, Le and Yang, Ge},
  booktitle={ICCV},
  year={2023}
}

@inproceedings{luo2021semi,
  title={Semi-supervised medical image segmentation through dual-task consistency},
  author={Luo, Xiangde and Chen, Jieneng and Song, Tao and Wang, Guotai},
  booktitle={AAAI},
  year={2021}
}

@article{luo2022semi,
  title={Semi-supervised medical image segmentation via uncertainty rectified pyramid consistency},
  author={Luo, Xiangde and Wang, Guotai and Liao, Wenjun and Chen, Jieneng and Song, Tao and Chen, Yinan and Zhang, Shichuan and Metaxas, Dimitris N and Zhang, Shaoting},
  journal={MedIA},
  year={2022},
}

@article{you2022simcvd,
  title={Simcvd: Simple contrastive voxel-wise representation distillation for semi-supervised medical image segmentation},
  author={You, Chenyu and Zhou, Yuan and Zhao, Ruihan and Staib, Lawrence and Duncan, James S},
  journal={TMI},
  year={2022},
}

@article{you2024mine,
  title={Mine your own anatomy: Revisiting medical image segmentation with extremely limited labels},
  author={You, Chenyu and Dai, Weicheng and Liu, Fenglin and Min, Yifei and Dvornek, Nicha C and Li, Xiaoxiao and Clifton, David A and Staib, Lawrence and Duncan, James S},
  journal={TPAMI},
  year={2024},
  publisher={IEEE}
}

@inproceedings{you2023rethinking,
  title={Rethinking semi-supervised medical image segmentation: A variance-reduction perspective},
  author={You, Chenyu and Dai, Weicheng and Min, Yifei and Liu, Fenglin and Clifton, David and Zhou, S Kevin and Staib, Lawrence and Duncan, James},
  booktitle={NeurIPS},
  year={2023}
}

@inproceedings{wu2022cross,
  title={Cross-patch dense contrastive learning for semi-supervised segmentation of cellular nuclei in histopathologic images},
  author={Wu, Huisi and Wang, Zhaoze and Song, Youyi and Yang, Lin and Qin, Jing},
  booktitle={CVPR},
  year={2022}
}

@book{edelsbrunner2010computational,
  title={Computational topology: an introduction},
  author={Edelsbrunner, Herbert and Harer, John},
  year={2010},
  publisher={American Mathematical Soc.}
}

@article{graham2019mild,
  title={MILD-Net: Minimal information loss dilated network for gland instance segmentation in colon histology images},
  author={Graham, Simon and Chen, Hao and Gamper, Jevgenij and Dou, Qi and Heng, Pheng-Ann and Snead, David and Tsang, Yee Wah and Rajpoot, Nasir},
  journal={MedIA},
  year={2019},
}

@inproceedings{tarvainen2017mean,
  title={Mean teachers are better role models: Weight-averaged consistency targets improve semi-supervised deep learning results},
  author={Tarvainen, Antti and Valpola, Harri},
  booktitle={NeurIPS},
  year={2017}
}

@article{cohen2010lipschitz,
  title={Lipschitz functions have L p-stable persistence},
  author={Cohen-Steiner, David and Edelsbrunner, Herbert and Harer, John and Mileyko, Yuriy},
  journal={Foundations of Computational Mathematics},
  year={2010},
}

@inproceedings{vu2019advent,
  title={Advent: Adversarial entropy minimization for domain adaptation in semantic segmentation},
  author={Vu, Tuan-Hung and Jain, Himalaya and Bucher, Maxime and Cord, Matthieu and P{\'e}rez, Patrick},
  booktitle={CVPR},
  year={2019}
}

@inproceedings{ouali2020semi,
  title={Semi-supervised semantic segmentation with cross-consistency training},
  author={Ouali, Yassine and Hudelot, C{\'e}line and Tami, Myriam},
  booktitle={CVPR},
  year={2020}
}

@inproceedings{gal2016dropout,
  title={Dropout as a bayesian approximation: Representing model uncertainty in deep learning},
  author={Gal, Yarin and Ghahramani, Zoubin},
  booktitle={ICML},
  year={2016},
}

@inproceedings{stucki2023topologically,
  title={Topologically faithful image segmentation via induced matching of persistence barcodes},
  author={Stucki, Nico and Paetzold, Johannes C and Shit, Suprosanna and Menze, Bjoern and Bauer, Ulrich},
  booktitle={ICML},
  year={2023},
}

@inproceedings{xie2019deep,
  title={Deep segmentation-emendation model for gland instance segmentation},
  author={Xie, Yutong and Lu, Hao and Zhang, Jianpeng and Shen, Chunhua and Xia, Yong},
  booktitle={MICCAI},
  year={2019},
}

@article{kuhn1955hungarian,
  title={The Hungarian method for the assignment problem},
  author={Kuhn, Harold W},
  journal={Naval research logistics quarterly},
  year={1955},
  publisher={Wiley Online Library}
}

@article{wen2024topology,
  title={Topology-Preserving Image Segmentation with Spatial-Aware Persistent Feature Matching},
  author={Wen, Bo and Zhang, Haochen and Bartsch, Dirk-Uwe G and Freeman, William R and Nguyen, Truong Q and An, Cheolhong},
  journal={arXiv preprint arXiv:2412.02076},
  year={2024}
}

@inproceedings{gao2024pmt,
  title={PMT: Progressive Mean Teacher via Exploring Temporal Consistency for Semi-Supervised Medical Image Segmentation},
  author={Gao, Ning and Zhou, Sanping and Wang, Le and Zheng, Nanning},
  booktitle={ECCV},
  year={2024},
}

@article{wang2022semi,
  title={Semi-supervised medical image segmentation via a tripled-uncertainty guided mean teacher model with contrastive learning},
  author={Wang, Kaiping and Zhan, Bo and Zu, Chen and Wu, Xi and Zhou, Jiliu and Zhou, Luping and Wang, Yan},
  journal={MedIA},
  year={2022},
  publisher={Elsevier}
}

@inproceedings{berthelot2019mixmatch,
  title={Mixmatch: A holistic approach to semi-supervised learning},
  author={Berthelot, David and Carlini, Nicholas and Goodfellow, Ian and Papernot, Nicolas and Oliver, Avital and Raffel, Colin A},
  booktitle={NeurIPS},
  year={2019}
}

@article{xie2024entropy,
  title={Entropy-guided contrastive learning for semi-supervised medical image segmentation},
  author={Xie, Junsong and Wu, Qian and Zhu, Renju},
  journal={IET Image Processing},
  year={2024},
  publisher={Wiley Online Library}
}

@article{hung2018adversarial,
  title={Adversarial learning for semi-supervised semantic segmentation},
  author={Hung, Wei-Chih and Tsai, Yi-Hsuan and Liou, Yan-Ting and Lin, Yen-Yu and Yang, Ming-Hsuan},
  journal={arXiv preprint arXiv:1802.07934},
  year={2018}
}

@article{lei2022semi,
  title={Semi-supervised medical image segmentation using adversarial consistency learning and dynamic convolution network},
  author={Lei, Tao and Zhang, Dong and Du, Xiaogang and Wang, Xuan and Wan, Yong and Nandi, Asoke K},
  journal={TMI},
  year={2022},
  publisher={IEEE}
}

@article{nair2020exploring,
  title={Exploring uncertainty measures in deep networks for multiple sclerosis lesion detection and segmentation},
  author={Nair, Tanya and Precup, Doina and Arnold, Douglas L and Arbel, Tal},
  journal={MedIA},
  year={2020},
  publisher={Elsevier}
}

@article{xu2023ambiguity,
  title={Ambiguity-selective consistency regularization for mean-teacher semi-supervised medical image segmentation},
  author={Xu, Zhe and Wang, Yixin and Lu, Donghuan and Luo, Xiangde and Yan, Jiangpeng and Zheng, Yefeng and Tong, Raymond Kai-yu},
  journal={MedIA},
  year={2023},
  publisher={Elsevier}
}

@inproceedings{yao2022enhancing,
  title={Enhancing pseudo label quality for semi-supervised domain-generalized medical image segmentation},
  author={Yao, Huifeng and Hu, Xiaowei and Li, Xiaomeng},
  booktitle={AAAI},
  year={2022}
}

@inproceedings{seibold2022reference,
  title={Reference-guided pseudo-label generation for medical semantic segmentation},
  author={Seibold, Constantin Marc and Rei{\ss}, Simon and Kleesiek, Jens and Stiefelhagen, Rainer},
  booktitle={AAAI},
  year={2022}
}

@inproceedings{berger2024topologically,
  title={Topologically faithful multi-class segmentation in medical images},
  author={Berger, Alexander H and Lux, Laurin and Stucki, Nico and B{\"u}rgin, Vincent and Shit, Suprosanna and Banaszak, Anna and Rueckert, Daniel and Bauer, Ulrich and Paetzold, Johannes C},
  booktitle={MICCAI},
  year={2024},
}

@inproceedings{zhang2023topology,
  title={Topology-Preserving Hard Pixel Mining for Tubular Structure Segmentation.},
  author={Zhang, Guoqing and Dong, Caixia and Li, Yang},
  booktitle={BMVC},
  year={2023}
}

@inproceedings{gupta2022learning,
  title={Learning topological interactions for multi-class medical image segmentation},
  author={Gupta, Saumya and Hu, Xiaoling and Kaan, James and Jin, Michael and Mpoy, Mutshipay and Chung, Katherine and Singh, Gagandeep and Saltz, Mary and Kurc, Tahsin and Saltz, Joel and others},
  booktitle={ECCV},
  year={2022},
}

@inproceedings{hu2020topology,
  title={Topology-Aware Segmentation Using Discrete Morse Theory},
  author={Hu, Xiaoling and Wang, Yusu and Fuxin, Li and Samaras, Dimitris and Chen, Chao},
  booktitle={ICLR},
  year={2021}
}

@inproceedings{shit2021cldice,
  title={clDice-a novel topology-preserving loss function for tubular structure segmentation},
  author={Shit, Suprosanna and Paetzold, Johannes C and Sekuboyina, Anjany and Ezhov, Ivan and Unger, Alexander and Zhylka, Andrey and Pluim, Josien PW and Bauer, Ulrich and Menze, Bjoern H},
  booktitle={CVPR},
  year={2021}
}

@inproceedings{hu2022structure,
  title={Structure-Aware Image Segmentation with Homotopy Warping},
  author={Hu, Xiaoling},
  booktitle={NeurIPS},
  year={2022}
}

@inproceedings{wang2022ta,
  title={Ta-net: Topology-aware network for gland segmentation},
  author={Wang, Haotian and Xian, Min and Vakanski, Aleksandar},
  booktitle={WACV},
  year={2022}
}

@inproceedings{hu2023learning,
  title={Learning Probabilistic Topological Representations Using Discrete Morse Theory},
  author={Hu, Xiaoling and Samaras, Dimitris and Chen, Chao},
  booktitle={ICLR},
  year={2023}
}

@article{clough2020topological,
  title={A topological loss function for deep-learning based image segmentation using persistent homology},
  author={Clough, James R and Byrne, Nicholas and Oksuz, Ilkay and Zimmer, Veronika A and Schnabel, Julia A and King, Andrew P},
  journal={TPAMI},
  year={2020},
}

@inproceedings{yang2021topological,
  title={A topological-attention convlstm network and its application to em images},
  author={Yang, Jiaqi and Hu, Xiaoling and Chen, Chao and Tsai, Chialing},
  booktitle={MICCAI},
  year={2021},
}

@inproceedings{wang2020topogan,
  title={Topogan: A topology-aware generative adversarial network},
  author={Wang, Fan and Liu, Huidong and Samaras, Dimitris and Chen, Chao},
  booktitle={ECCV},
  year={2020},
}

@inproceedings{xu2024topocellgen,
  title={TopoCellGen: Generating Histopathology Cell Topology with a Diffusion Model},
  author={Xu, Meilong and Gupta, Saumya and Hu, Xiaoling and Li, Chen and Abousamra, Shahira and Samaras, Dimitris and Prasanna, Prateek and Chen, Chao},
  booktitle={CVPR},
  year={2025}
}

@inproceedings{zhou2024xnet,
  title={XNet v2: Fewer Limitations, Better Results and Greater Universality},
  author={Zhou, Yanfeng and Li, Lingrui and Wang, Zichen and Liu, Guole and Liu, Ziwen and Yang, Ge},
  booktitle={BIBM},
  year={2024},
}

@inproceedings{nguyen2025semi,
  title={Semi-Supervised Histopathology Image Segmentation with Feature Diversified Collaborative Learning},
  author={Nguyen, Thanh-Huy and Vu, Nguyen Lan Vi and Nguyen, Hoang-Thien and Dinh, Quang-Vinh and Li, Xingjian and Xu, Min},
  booktitle={AAAI Bridge Program on AI for Medicine and Healthcare},
  year={2025},
}

@article{pearson1895vii,
  title={VII. Note on regression and inheritance in the case of two parents},
  author={Pearson, Karl},
  journal={proceedings of the royal society of London},
  year={1895},
  publisher={The Royal Society London}
}

@inproceedings{li2023confidence,
  title={Confidence estimation using unlabeled data},
  author={Li, Chen and Hu, Xiaoling and Chen, Chao},
  booktitle={ICLR},
  year={2023}
}

@article{laine2016temporal,
  title={Temporal ensembling for semi-supervised learning},
  author={Laine, Samuli and Aila, Timo},
  journal={arXiv preprint arXiv:1610.02242},
  year={2016}
}

@article{shin2024revisiting,
  title={Revisiting and maximizing temporal knowledge in semi-supervised semantic segmentation},
  author={Shin, Wooseok and Park, Hyun Joon and Kim, Jin Sob and Han, Sung Won},
  journal={arXiv preprint arXiv:2405.20610},
  year={2024}
}

@inproceedings{smith1979tint,
  title={Tint fill},
  author={Smith, Alvy Ray},
  booktitle={Proceedings of the 6th annual conference on Computer graphics and interactive techniques},
  year={1979}
}

@article{paszke2019pytorch,
  title={Pytorch: An imperative style, high-performance deep learning library},
  author={Paszke, A},
  journal={arXiv preprint arXiv:1912.01703},
  year={2019}
}

@article{kingma2014adam,
  title={Adam: A method for stochastic optimization},
  author={Kingma, Diederik P},
  journal={arXiv preprint arXiv:1412.6980},
  year={2014}
}

@inproceedings{chen2018encoder,
  title={Encoder-decoder with atrous separable convolution for semantic image segmentation},
  author={Chen, Liang-Chieh and Zhu, Yukun and Papandreou, George and Schroff, Florian and Adam, Hartwig},
  booktitle={ECCV},
  year={2018}
}

@book{mnih2013machine,
  title={Machine learning for aerial image labeling},
  author={Mnih, Volodymyr},
  year={2013},
  publisher={University of Toronto (Canada)}
}

@article{jordan1999introduction,
  title={An introduction to variational methods for graphical models},
  author={Jordan, Michael I and Ghahramani, Zoubin and Jaakkola, Tommi S and Saul, Lawrence K},
  journal={Machine learning},
  year={1999},
  publisher={Springer}
}

@inproceedings{guo2017calibration,
  title={On calibration of modern neural networks},
  author={Guo, Chuan and Pleiss, Geoff and Sun, Yu and Weinberger, Kilian Q},
  booktitle={ICML},
  year={2017},
}

@inproceedings{kirillov2023segment,
  title={Segment anything},
  author={Kirillov, Alexander and Mintun, Eric and Ravi, Nikhila and Mao, Hanzi and Rolland, Chloe and Gustafson, Laura and Xiao, Tete and Whitehead, Spencer and Berg, Alexander C and Lo, Wan-Yen and others},
  booktitle={ICCV},
  year={2023}
}

@inproceedings{hu2022lora,
  title={Lora: Low-rank adaptation of large language models.},
  author={Hu, Edward J and Shen, Yelong and Wallis, Phillip and Allen-Zhu, Zeyuan and Li, Yuanzhi and Wang, Shean and Wang, Lu and Chen, Weizhu and others},
  booktitle={ICLR},
  year={2022}
}

@article{verma2021monusac2020,
  title={MoNuSAC2020: A multi-organ nuclei segmentation and classification challenge},
  author={Verma, Ruchika and Kumar, Neeraj and Patil, Abhijeet and Kurian, Nikhil Cherian and Rane, Swapnil and Graham, Simon and Vu, Quoc Dang and Zwager, Mieke and Raza, Shan E Ahmed and Rajpoot, Nasir and others},
  journal={TMI},
  year={2021},
  publisher={IEEE}
}

@article{Wu2009OptimizingCCL,
  author  = {Kesheng Wu and Ekow J. Otoo and Kenji Suzuki},
  title   = {Optimizing Two-Pass Connected-Component Labeling Algorithms},
  journal = {Pattern Analysis and Applications},
  year    = {2009},
  doi     = {10.1007/s10044-008-0109-y}
}

@inproceedings{yang20213d,
  title={3d topology-preserving segmentation with compound multi-slice representation},
  author={Yang, Jiaqi and Hu, Xiaoling and Chen, Chao and Tsai, Chialing},
  booktitle={ISBI},
  year={2021},
}

@article{yang2024anomaly,
  title={Anomaly-guided weakly supervised lesion segmentation on retinal OCT images},
  author={Yang, Jiaqi and Mehta, Nitish and Demirci, Gozde and Hu, Xiaoling and Ramakrishnan, Meera S and Naguib, Mina and Chen, Chao and Tsai, Chia-Ling},
  journal={MedIA},
  year={2024},
}

@article{yang2024multimodal,
  title={A Multimodal Approach Combining Structural and Cross-domain Textual Guidance for Weakly Supervised OCT Segmentation},
  author={Yang, Jiaqi and Mehta, Nitish and Hu, Xiaoling and Chen, Chao and Tsai, Chia-Ling},
  journal={arXiv preprint arXiv:2411.12615},
  year={2024}
}
}

\clearpage
\newpage

\setcounter{section}{5}
\setcounter{table}{9}

\begin{center}
\LARGE \textbf{MATCH: \underline{M}ulti-faceted \underline{A}daptive \underline{T}opo-\underline{C}onsistency for
Semi-Supervised \underline{H}istopathology Segmentation}\\[0.5em]
\large ---Supplementary Material---
\end{center}
% \vspace{1em}

\section{Overview}
\label{sec:overview}
In the supplementary, we begin with a brief introduction to the persistent homology in~\Cref{sec:intro_ph}, followed by detailed introductions of the datasets in~\Cref{sec:datasets} and the evaluation metrics in~\Cref{sec:eval_metrics}. Then, we provide the implementation details in~\Cref{sec:impl_details}, followed by the references of our baselines in~\Cref{sec:baseline_ref}. We also provide additional ablation studies in~\Cref{sec:addi_ablation} to further illustrate the efficacy and robustness of our method and hyper-parameter selections. The limitations are provided in~\Cref{sec:limitations}, followed by an analysis on the broader impact in~\Cref{sec:broader_impact}.

\section{Brief Introduction to Persistent Homology}
\label{sec:intro_ph}
Persistent homology~\cite{cohen2010lipschitz,edelsbrunner2010computational}, a fundamental concept in topological data analysis (TDA), offers a robust framework for capturing and quantifying the topological features of data across multiple scales. In the context of image segmentation, particularly when dealing with likelihood maps that represent the probability of each pixel belonging to a specific class, persistent homology provides a means to analyze the underlying topological structures inherent in these probabilistic representations.

Given a likelihood map \( f: \Omega \rightarrow [0,1] \), where \( \Omega \subset \mathbb{R}^2 \) represents the image domain, we construct a filtration of super-level sets:
\[
\mathcal{F}_\alpha = \{ x \in \Omega \mid f(x) \geq \alpha \}, \quad \alpha \in [0,1].
\]

As \( \alpha \) decreases from 1 to 0, the super-level set \( \mathcal{F}_\alpha \) transitions from empty regions to encompass the entire domain \( \Omega \), revealing the sequential emergence, merging, and disappearance of connected components and loops. Persistent homology tracks these topological changes across the filtration, recording the corresponding birth and death thresholds of each feature in a persistence diagram.

A persistence diagram is a multiset of points \( \{(b_i, d_i)\} \) in the extended plane \( \mathbb{R}^2 \), where each point corresponds to a topological feature that appears (birth \( b_i \)) and disappears (death \( d_i \)) during the filtration process. Features that persist across a wide range of \( \alpha \) values (i.e., with large \( |d_i - b_i| \)) are considered topologically significant, while those with short lifespans are often attributed to noise.

\section{Dataset Details}
\label{sec:datasets}
\textbf{Colorectal Adenocarcinoma Gland (CRAG)}~\cite{graham2019mild} consists of $213$ hematoxylin and eosin (H\&E)-stained colorectal adenocarcinoma image tiles acquired at $20\times$ magnification, each with detailed annotations at the instance level. Most images are in approximately $1512\times1516$ pixels. Officially, the dataset is partitioned into $173$ training samples and $40$ testing samples. For our experiments, the training subset is further divided into $153$ images for model training and $20$ images for validation. For semi-supervised scenarios with 10\% and 20\% labeled data, we randomly select $16$ and $31$ labeled images, respectively, for training.

\textbf{Gland Segmentation in Colon Histology Images Challenge (GlaS)}~\cite{sirinukunwattana2017gland} comprises $165$ images sourced from $16$ H\&E-stained histological slides of colorectal adenocarcinoma at stages T$3$ or T$4$. The official split includes $85$ training images and $80$ testing images. In our experimental setup, the training set is divided into $68$ images for model training and $17$ for validation. We randomly select $7$ and $14$ labeled images to represent 10\% and 20\% of labeled training data scenarios, respectively.

\textbf{Multi-Organ Nuclei Segmentation (MoNuSeg)}~\cite{kumar2019multi} dataset contains $44$ H\&E-stained histology images of dimensions $1000\times1000$ pixels, encompassing nuclei annotations from seven distinct organs. Officially, it consists of $30$ training images with a total of $21,623$ annotated nuclei and $14$ images designated for testing. For our experiments, we reserve $20\%$ ($6$ images) of the training set for validation. In experiments involving 10\% and 20\% labeled data splits, we randomly select $3$ and $5$ labeled images, respectively, for training.

\section{Evaluation Metrics}
\label{sec:eval_metrics}
We evaluate the segmentation quality from both pixel- and topology-wise. 
\textbf{Object-level Dice coefficient (Dice\_{Obj})}~\cite{xie2019deep} is selected to measure pixel-wise performance, which measures instance-wise overlap between predicted and ground-truth masks and is thus well suited to the precise delineation of individual structures required in digital pathology.

To evaluate the topological accuracy, we select three topological evaluation metrics, \textbf{Betti Error}~\cite{hu2019topology}, \textbf{Betti Matching Error}~\cite{stucki2023topologically}, and \textbf{Discrepancy between Intersection and Union (DIU)}~\cite{lux2024topograph}. 
Betti Error (BE) mainly computes the mean absolute difference in 0-dimensional Betti numbers over $256 \times 256$ sliding-window patches. 
Betti Matching Error (BME), which extends BE by enforcing spatial correspondence when pairing topological features, thereby penalizing misplaced components even when counts are preserved.
Introduced in~\cite{lux2024topograph}, DIU quantifies how faithfully the topology of the common and combined foreground regions agrees.

\section{Implementation Details}
\label{sec:impl_details}
Our model is trained in two distinct stages. In the initial stage, we perform pretraining using only supervised loss and pixel-wise consistency loss. For all three datasets, the pretraining stage proceeds for $12,000$ iterations. The second stage involves fine-tuning the model by integrating our proposed dual-level topological consistency constraints, which last for an additional $1,000$ iterations. We use UNet~\cite{ronneberger2015u} as the backbone for both the student and teacher models.

All training is implemented using PyTorch~\cite{paszke2019pytorch} and optimized using the Adam optimizer~\cite{kingma2014adam}. Training hyperparameters are set as follows: the batch size is $16$ and the learning rate is $5 \times 10^{-4}$. Both labeled and unlabeled data undergo pre-processing through random cropping (with cropping size of $256 \times 256$), followed by data augmentation procedures including random rotation and flipping as weak augmentations, and color adjustments and morphological shifts for stronger augmentations.

In particular, we adopt a random cropping strategy for enforcing intra-topological consistency, while a fixed patch cropping strategy is used for temporal-topological consistency. \textbf{The inputs to the student model to estimate the intra- and temporal-topological consistency are all original patches, without any transformations.}
The EMA decay rate $\alpha$ is set to $0.999$.
Within the supervised loss, the weights assigned to the cross-entropy loss and Dice loss are equally set to $0.5$. The weight of the pixel-wise consistency loss is calculated by the Gaussian ramp-up function $\lambda_{\text{cons}}=k*e^{-5*(1-\frac{\tau}{T})^2}$, where $k=0.1$ and $T$ is the total number of iterations. 

Additionally, $\lambda_{\text{intra}}$ and $\lambda_{\text{temp}}$ are both set to $0.001$. This balanced configuration ensures effective integration of topological constraints while maintaining stable training dynamics. \textbf{Note that dual-level topological consistency is used to optimize the student model directly, and we use the student model to do the inference.} 
%All the experiments are conducted on an NVIDIA RTX A6000 GPU with 48 GB RAM.
The experiments are conducted on an NVIDIA RTX A6000 GPU (48 GB), using a 24-core Intel® Xeon® Gold 6248R CPU @ 3.00 GHz and 192 GB RAM. The training time of one iteration is $1020.04$ ms, and GPU memory consumption is $25.726$ GB using UNet with batch size $16$. The training time of TopoSemiSeg for one iteration is $610.80$ ms, and the GPU memory consumption is $15.235$ GB. For the non-PH baseline, like PMT~\cite{gao2024pmt}, the training time per iteration is $582.34$ ms,

\section{Baseline Reference}
\label{sec:baseline_ref}
We select $7$ classical and recent state-of-the-art methods as comparatives. The implementations of some of them are based on publicly available repositories. Here, we provide the source of our baselines for reference and greatly appreciate their efforts in building the open-source community:

MT~\cite{tarvainen2017mean}, EM~\cite{vu2019advent}, UA-MT~\cite{yu2019uncertainty} and URPC~\cite{luo2022semi} are based on the implementations from: \href{https://github.com/HiLab-git/SSL4MIS}{https://github.com/HiLab-git/SSL4MIS}.

XNet~\cite{zhou2023xnet} is based on the implementations from:

\href{https://github.com/guspan-tanadi/XNetfromYanfeng-Zhou}{https://github.com/guspan-tanadi/XNetfromYanfeng-Zhou}.

PMT~\cite{gao2024pmt} is based on the implementations from: 
\href{http://github.com/Axi404/PMT}{http://github.com/Axi404/PMT}.

TopoSemiSeg~\cite{xu2024semi} is based on the implementations from: 

\href{https://github.com/Melon-Xu/TopoSemiSeg}{https://github.com/Melon-Xu/TopoSemiSeg}.

\section{Additional Ablation Study}
\label{sec:addi_ablation}
Here, we provide additional ablation studies to illustrate the efficacy and robustness of our selected backbone and hyperparameters.

\myparagraph{Ablation Study on $\lambda_{\text{intra}}$ and $\lambda_{\text{temp}}$.}
The results shown in~\Cref{tab:loss_weights} demonstrate the impact of varing weights of intra- and temporal-topological consistency ($\lambda_{\text{intra}}$ and $\lambda_{\text{temp}}$). When two weights are both $0.001$, the performance is the best across both pixel-wise and topology-wise metrics. As these weights increase from $0.001$ to $0.01$, there's a clear degradation in performance, indicating that excessively large consistency constraints may introduce unnecessary regularization, thus impairing the segmentation quality. When both weights are reduced to $0.0005$, the dual-level consistency regularization becomes too weak to meaningfully optimize the student model, leading to diminished topological guidance and a corresponding drop in both pixel-wise and topology-wise performance.

\myparagraph{Ablation Study on EMA Decay $\alpha$.}
\Cref{tab:ema_decay} investigates the influence of the EMA decay parameter $\alpha$. $\alpha=0.999$ yields the best performance.
When decreasing $\alpha$ from $0.999$ to $0.996$, the results remain competitive but slightly deteriorate, highlighting that a higher EMA decay value effectively leverages historical model parameters for improved topological and segmentation robustness. In contrast, very high values (e.g. $\alpha=0.9999$) excessively rely on historical information, marginally weakening the adaptability and performance of the model.

% Keep these two tables in one row; requires \usepackage{caption} for \captionof
\begin{table*}[ht]
  \centering
  \tiny

  % ---------- Left: Influence of lambda_intra and lambda_temp ----------
  \begin{minipage}[t]{0.49\linewidth}
    \centering
    \captionof{table}{Influence of $\lambda_{\text{intra}}$ and $\lambda_{\text{temp}}$.}
    \label{tab:loss_weights}
    \setlength{\tabcolsep}{1.5pt}
    \begin{tabular}{lcccccc}
      \toprule
      \multirow{2}{*}{$\lambda_{\text{intra}}$} &
      \multirow{2}{*}{$\lambda_{\text{temp}}$} &
      \multicolumn{1}{c}{Pixel-wise} & &
      \multicolumn{3}{c}{Topology-wise} \\ \cmidrule(lr){3-3}\cmidrule(lr){5-7}
       & & Dice\_obj $\uparrow$ && BE $\downarrow$ & BME $\downarrow$ & DIU $\downarrow$ \\ \midrule
      0.01   & 0.01   & 0.865 $\pm$ 0.012 && 0.275 $\pm$ 0.023 & 10.650 $\pm$ 0.630 & 53.500 $\pm$ 2.300 \\
      0.005  & 0.005  & 0.892 $\pm$ 0.007 && 0.214 $\pm$ 0.020 & 8.105 $\pm$ 0.600 & 44.105 $\pm$ 2.050 \\ 
      \underline{0.001}  & \underline{0.001}  & \textbf{0.909 $\pm$ 0.005} && \textbf{0.188 $\pm$ 0.018} & \textbf{7.425 $\pm$ 0.570} & \textbf{40.250 $\pm$ 1.720} \\ 
      0.0005 & 0.0005 & 0.895 $\pm$ 0.007 && 0.235 $\pm$ 0.020 & 8.950 $\pm$ 0.580 & 44.225 $\pm$ 1.850 \\
      \bottomrule
    \end{tabular}
  \end{minipage}
  \hfill
  % ---------- Right: Impact of EMA decay alpha ----------
  \begin{minipage}[t]{0.49\linewidth}
    \centering
    \captionof{table}{Impact of the EMA decay $\alpha$.}
    \label{tab:ema_decay}
    \setlength{\tabcolsep}{.8pt}
    \begin{tabular}{lccccc}
      \toprule
      \multirow{2}{*}{$\alpha$} &
      \multicolumn{1}{c}{Pixel-wise} & &
      \multicolumn{3}{c}{Topology-wise} \\ \cmidrule(lr){2-2}\cmidrule(lr){4-6}
       & Dice\_obj $\uparrow$ && BE $\downarrow$ & BME $\downarrow$ & DIU $\downarrow$ \\ \midrule
      0.9999 & 0.890 $\pm$ 0.007 && 0.230 $\pm$ 0.022 & 9.500 $\pm$ 0.630 & 46.000 $\pm$ 2.100 \\ 
      \underline{0.999} & \textbf{0.909 $\pm$ 0.005} && \textbf{0.188 $\pm$ 0.018} & \textbf{7.425 $\pm$ 0.570} & \textbf{40.250 $\pm$ 1.720} \\ 
      0.996  & 0.902 $\pm$ 0.006 && 0.205 $\pm$ 0.020 & 8.250 $\pm$ 0.610 & 42.500 $\pm$ 1.900 \\ 
      0.99   & 0.882 $\pm$ 0.008 && 0.260 $\pm$ 0.025 & 11.000 $\pm$ 0.700 & 50.000 $\pm$ 2.300 \\ 
      \bottomrule
    \end{tabular}
  \end{minipage}

\end{table*}

\myparagraph{Ablation Study on Different Backbones.}
To further verify the robustness of our proposed method, we conduct ablation experiments on different backbones. The results are shown in~\Cref{tab:backbone_comparison}. Specifically, DeepLabV3+~\cite{chen2018encoder} and UNet++~\cite{zhou2018unet++} show modest but clear improvements in both pixel-wise and topology-wise metrics. The UNet~\cite{ronneberger2015u} backbone achieves the most substantial gains, particularly in topology-wise metrics. These results demonstrate that integrating our MATCH framework consistently improves performance across multiple backbones.

\begin{table}[ht]
  \centering
  \tiny
  \caption{Performance comparison of different backbones w or w/o our MATCH.}
  \label{tab:backbone_comparison}
  \setlength{\tabcolsep}{7pt}
  \begin{tabular}{lccccc}
    \toprule
    \multirow{2}{*}{Backbone} & \multicolumn{1}{c}{Pixel-Wise} & &
    \multicolumn{3}{c}{Topology-Wise} \\ \cmidrule(lr){2-2}\cmidrule(lr){4-6}
     & Dice\_Obj $\uparrow$ && BE $\downarrow$ & BME $\downarrow$ & VOI $\downarrow$ \\ \midrule
    DeepLabV3+~\cite{chen2018encoder}               & 0.889 $\pm$ 0.010 && 0.272 $\pm$ 0.023 & 11.782 $\pm$ 0.690 & 50.867 $\pm$ 2.221\\
    DeepLabV3+~\cite{chen2018encoder}+Ours          & 0.892 $\pm$ 0.008 && 0.245 $\pm$ 0.022 & 10.129 $\pm$ 0.638 & 47.412 $\pm$ 2.047 \\ \midrule
    UNet++~\cite{zhou2018unet++}                  & 0.886 $\pm$ 0.008 && 0.245 $\pm$ 0.023 &  9.210 $\pm$ 0.603 & 45.517 $\pm$ 2.041 \\
    UNet++~\cite{zhou2018unet++}+Ours             & 0.890 $\pm$ 0.006 && 0.238 $\pm$ 0.020 &  9.021 $\pm$ 0.580 & 45.073 $\pm$ 1.995 \\ \midrule
    UNet~\cite{ronneberger2015u}                    & 0.894 $\pm$ 0.006 && 0.232 $\pm$ 0.019 &  8.872 $\pm$ 0.579 & 44.281 $\pm$ 1.881\\
    UNet~\cite{ronneberger2015u}+Ours               & \textbf{0.909 $\pm$ 0.005} && \textbf{0.188 $\pm$ 0.018} & \textbf{7.425 $\pm$ 0.570} & \textbf{40.250 $\pm$ 1.720}\\
    \bottomrule
  \end{tabular}
\end{table}

\myparagraph{Ablation Study on Applying Dual-Level Topo-Consistency between Teacher and Student Models.}
We conduct an ablation study to assess the impact of enforcing dual-level topological consistency in a teacher-student framework. Specifically, the dual-level topological consistency is estimated from the teacher model's multiple predictions, and consistency constraints are applied between the student output and the most recent prediction from the teacher. We compare this teacher-student configuration with a student-only model, both trained under identical consistency constraints. The results in~\Cref{tab:teacher_student_ablation} reveal that the student-only model consistently achieves superior performance. The relatively poorer performance of the teacher-student configuration suggests that leveraging the teacher's predictions, potentially noisy or outdated, introduces additional uncertainty and adversely affects the student's ability to effectively capture stable topological structures.

\begin{table}[ht]
  \centering
  \scriptsize
  \caption{Ablation study on applying dual-level topo-consistency between teacher and student models.}
  \label{tab:teacher_student_ablation}
  \setlength{\tabcolsep}{6pt}  % column separation
  \begin{tabular}{lccccc}
    \toprule
    \multirow{2}{*}{Mode} & \multicolumn{1}{c}{Pixel-wise} &  & \multicolumn{3}{c}{Topology-wise} \\ 
    \cmidrule(lr){2-2}\cmidrule(lr){4-6}
     & Dice\_Obj $\uparrow$ &  & BE $\downarrow$ & BME $\downarrow$ & DIU $\downarrow$ \\ 
    \midrule
    Teacher--Student & 0.885 $\pm$ 0.007 &  & 0.217 $\pm$ 0.021 & 8.102 $\pm$ 0.620 & 42.520 $\pm$ 1.880\\ 
    Student Only     & \textbf{0.909 $\pm$ 0.005} && \textbf{0.188 $\pm$ 0.018} & \textbf{7.425 $\pm$ 0.570} & \textbf{40.250 $\pm$ 1.720}\\ 
    \bottomrule
  \end{tabular}
\end{table}

\myparagraph{Extension from Binary to Multi-Class Segmentation.}
To extend our method to the multi-class setting, we choose a multi-class nuclei segmentation dataset, MoNuSAC~\cite{verma2021monusac2020}, to conduct experiments. This dataset contains four cell types: Epithelial, Lymphocyte, Macrophage, and Neutrophil. We conducted experiments using 20\% labeled data and report the class-wise performance of TopoSemiSeg and our method in~\Cref{tab:multi_class}. As demonstrated in our class-specific results on MoNuSAC (Epithelial, Lymphocyte, Macrophage, and Neutrophil), our approach consistently outperforms TopoSemiSeg across all cell types, with particularly notable improvements in topological metrics (BE and BME) that are crucial for distinguishing overlapping structures.

\begin{table}[ht]
\centering
\scriptsize
\caption{The multi-class segmentation results on the MoNuSAC dataset.}
\label{tab:multi_class}
\setlength{\tabcolsep}{6pt}
\begin{tabular}{llccc}
\toprule
Class & Method & Dice\_obj\,$\uparrow$ & BE\,$\downarrow$ & BME\,$\downarrow$ \\
\midrule
\multirow{2}{*}{Epithelial}
& TopoSemiSeg & 0.778 $\pm$ 0.009 & 5.342 $\pm$ 0.187 & 195.158 $\pm$ 4.627 \\
& Ours & \textbf{0.781 $\pm$ 0.008} & \textbf{5.128 $\pm$ 0.189} & \textbf{186.847 $\pm$ 3.958} \\
\midrule
\multirow{2}{*}{Lymphocyte}
& TopoSemiSeg & 0.751 $\pm$ 0.013 & 6.089 $\pm$ 0.223 & 218.394 $\pm$ 5.841 \\
& Ours & \textbf{0.756 $\pm$ 0.012} & \textbf{5.794 $\pm$ 0.235} & \textbf{207.693 $\pm$ 4.672} \\
\midrule
\multirow{2}{*}{Macrophage}
& TopoSemiSeg & 0.765 $\pm$ 0.011 & 5.687 $\pm$ 0.201 & 206.732 $\pm$ 4.985 \\
& Ours & \textbf{0.769 $\pm$ 0.010} & \textbf{5.423 $\pm$ 0.208} & \textbf{195.381 $\pm$ 4.127} \\
\midrule
\multirow{2}{*}{Neutrophil}
& TopoSemiSeg & 0.738 $\pm$ 0.016 & 6.521 $\pm$ 0.267 & 234.576 $\pm$ 6.123 \\
& Ours & \textbf{0.742 $\pm$ 0.015} & \textbf{6.187 $\pm$ 0.281} & \textbf{221.459 $\pm$ 5.894} \\
\bottomrule
\end{tabular}
\end{table}

\myparagraph{Ablation Study on the Sensitivity of $\tau_{primary}$.}
We provide the ablation study on the sensitivity of $\tau_{primary}$ in~\Cref{tab:tau_ablation}. The results have shown that our method is robust to selecting $\tau_{primary}$. Moreover, the low threshold of 0.1 was chosen to be inclusive rather than restrictive: it allows more potential matches to be considered valid while letting the Hungarian algorithm determine optimal assignments based on our comprehensive similarity metric (combining spatial overlap, persistence weights, and proximity).

\begin{table*}[ht]
\centering
\scriptsize
\setlength{\tabcolsep}{.4pt}

% -------- Left: Effect of tau_primary --------
\begin{minipage}[t]{0.48\textwidth}
\centering
\captionof{table}{Effect of $\tau_{\text{primary}}$.}
\label{tab:tau_ablation}
\begin{tabular}{lcccc}
\toprule
$\tau_{\text{primary}}$ & \textbf{Dice\_obj} $\uparrow$ & \textbf{BE} $\downarrow$ & \textbf{BME} $\downarrow$ & \textbf{DIU} $\downarrow$ \\
\midrule
0.05           & 0.906 $\pm$ 0.006 & 0.195 $\pm$ 0.019 & 7.850 $\pm$ 0.620 & 41.750 $\pm$ 1.850 \\
0.1 (current)  & \textbf{0.909 $\pm$ 0.005} & \textbf{0.188 $\pm$ 0.018} & \textbf{7.425 $\pm$ 0.570} & \textbf{40.250 $\pm$ 1.720} \\
0.2            & 0.908 $\pm$ 0.005 & 0.191 $\pm$ 0.020 & 7.680 $\pm$ 0.590 & 41.100 $\pm$ 1.780 \\
0.3            & 0.905 $\pm$ 0.006 & 0.201 $\pm$ 0.021 & 8.150 $\pm$ 0.650 & 42.850 $\pm$ 1.920 \\
\bottomrule
\end{tabular}
\end{minipage}
\hfill
% -------- Right: Effect of dropout rates --------
\begin{minipage}[t]{0.48\textwidth}
\centering
\captionof{table}{Effect of dropout rates.}
\label{tab:dropout_rate}
\begin{tabular}{lccc}
\toprule
\textbf{Dropout Rate} & \textbf{Dice\_obj} $\uparrow$ & \textbf{BE} $\downarrow$ & \textbf{BME} $\downarrow$ \\
\midrule
10\%            & 0.898 $\pm$ 0.006 & 0.210 $\pm$ 0.020 & 8.200 $\pm$ 0.650 \\
20\% (current)  & \textbf{0.909 $\pm$ 0.005} & \textbf{0.188 $\pm$ 0.018} & \textbf{7.425 $\pm$ 0.570} \\
30\%            & 0.910 $\pm$ 0.005 & 0.185 $\pm$ 0.017 & 7.350 $\pm$ 0.560 \\
50\%            & 0.890 $\pm$ 0.007 & 0.220 $\pm$ 0.022 & 8.800 $\pm$ 0.720 \\
\bottomrule
\end{tabular}
\end{minipage}

\end{table*}

\myparagraph{The Impact of Different Dropout Rates.}
We also add complementary ablation studies on the dropout rate of the MC-dropout. Other settings are kept unchanged. We conduct the ablation experiments on CRAG 20\% labeled data and report the performance in~\Cref{tab:dropout_rate}. The results reveal an optimal dropout rate range of 20\%-30\% for our framework, where performance plateaus with minimal differences between these rates. Lower dropout rates provide insufficient perturbation diversity for reliable topological matching. In contrast, excessive dropout introduces detrimental noise that degrades both pixel- and topology-wise performance, confirming that moderate stochasticity is essential for effective topological consistency estimation.

\myparagraph{Ablation Study on $\mathcal{L}_{\text{cons}}$.}
We conducted the ablation study on the $\mathcal{L}_{\text{cons}}$ and the results are shown in~\Cref{tab:const_ablation}. Based on the results of the ablation study and the principles of semi-supervised learning, removing the pixel-wise consistency term in the training stages would result in significant performance degradation across all metrics.

\begin{table}[ht]
\centering
\caption{Ablation Study on $\mathcal{L}_{\text{cons}}$.}
\label{tab:const_ablation}
\scriptsize
\setlength{\tabcolsep}{10pt}
\begin{tabular}{lccc}
\toprule
\textbf{Method} & \textbf{Dice\_obj} $\uparrow$ & \textbf{BE} $\downarrow$ & \textbf{BME} $\downarrow$ \\
\midrule
w/o $\mathcal{L}_{\text{cons}}$ & 0.875 $\pm$ 0.008 & 0.285 $\pm$ 0.025 & 9.850 $\pm$ 0.680 \\
Ours & \textbf{0.909 $\pm$ 0.005} & \textbf{0.188 $\pm$ 0.018} & \textbf{7.425 $\pm$ 0.570} \\
\bottomrule
\end{tabular}
\end{table}

\myparagraph{Downstream Analysis: Cell counting}
To further analyze the impact of our method on downstream analysis, we conducted a cell counting study on the same MoNuSeg test cohort. We used the connected component analysis~\cite{Wu2009OptimizingCCL} to identify the cells and calculate the total cell count, the predicted total cell count, and the absolute counting error (mean $\pm$ std). The results are shown in~\Cref{tab:cell_counting}.
Note that the Total GT cell count and the predicted cell count are reported for the entire test cohort, while the absolute count error is reported per image (with a total of $14$ test images).

We observed that our method yields noticeably smaller counting errors than both baseline approaches (one topo method and one non-topo method). This confirms that although the pixel-wise segmentation performances are comparable, fixing the topological errors on a few pixels leads to more accurate biological readouts.

\begin{table}[ht]
\centering
\scriptsize
\setlength{\tabcolsep}{5pt}
\caption{Downstream Analysis on Cell Counting.}
\label{tab:cell_counting}
\begin{tabular}{lcccc}
\toprule
\textbf{Method} & \textbf{Total GT Cell Count} & \textbf{Predicted Cell Count} & \textbf{Absolute Counting Error (Mean $\pm$ Std)} & \textbf{Dice\_obj} \\
\midrule
PMT [8]          & 6024 & 8106 & 148.71 $\pm$ 99.41 & 0.778 $\pm$ 0.006 \\
TopoSemiSeg [57] & 6024 & 7877 & 132.36 $\pm$ 56.09 & 0.793 $\pm$ 0.004 \\
MATCH   & 6024 & \textbf{7511} & \textbf{106.21 $\pm$ 49.30} & 0.790 $\pm$ 0.006 \\
\bottomrule
\end{tabular}
\end{table}

\section{Limitations}
\label{sec:limitations}
% A potential limitation of our proposed MATCH framework arises from its dependency on stable feature extraction from persistence diagrams, which may become challenging when predictions exhibit extreme noise or minimal structural differences.
A potential limitation of our MATCH framework arises from its reliance on stable feature extraction from persistence diagrams, which can be challenged when predictions exhibit extreme noise or minimal structural differences. In addition, the framework introduces nontrivial computational overhead: computing persistence diagrams and performing MATCH-Global/MATCH-Pair alignments across Monte Carlo dropout samples and temporal snapshots require multiple forward passes and matching steps, resulting in longer training times and increased memory usage.

\section{Broader Impact}
\label{sec:broader_impact}
Our method significantly contributes to enhancing segmentation robustness by effectively leveraging unlabeled data, reducing reliance on extensive annotations, and ensuring topological accuracy crucial for clinical and biomedical analysis. This approach not only facilitates efficient utilization of limited labeled data but also provides insightful uncertainty estimates beneficial for downstream diagnostic applications. 

A negative broader impact could include inadvertent propagation of segmentation inaccuracies if poorly matched topological structures influence model learning, potentially affecting reliability in critical medical decisions.

\end{document}